\documentclass{article}

\PassOptionsToPackage{numbers,sort&compress}{natbib}

\usepackage[final]{neurips_2024}

\usepackage[utf8]{inputenc}
\usepackage[T1]{fontenc}
\usepackage[hidelinks]{hyperref}
\usepackage{booktabs}
\usepackage{amsfonts}
\usepackage{nicefrac}
\usepackage{microtype}
\usepackage[x11names]{xcolor}

\usepackage{amsmath}
\usepackage{amssymb}
\usepackage{amsthm}
\usepackage{bm}
\usepackage{bbm}
\usepackage{enumitem}
\usepackage{graphicx}
\usepackage{mathtools}
\usepackage{physics}
\usepackage{silence}
\usepackage{thmtools}
\usepackage{thm-restate}
\usepackage{xparse}
\usepackage{xurl}

\usepackage[capitalize,nameinlink]{cleveref}
\usepackage[breakable]{tcolorbox}

\hypersetup{
  colorlinks,
  citecolor=blue,
  linkcolor=Firebrick3, 
  urlcolor=Firebrick3
}

\title{Wide Two-Layer Networks can Learn from Adversarial Perturbations}

\author{Soichiro Kumano\\
The University of Tokyo\\
\texttt{kumano@cvm.t.u-tokyo.ac.jp}
\And
Hiroshi Kera\\
Chiba University, Zuse Institute Berlin\\
\texttt{kera@chiba-u.jp}
\And
Toshihiko Yamasaki\\
The University of Tokyo\\
\texttt{yamasaki@cvm.t.u-tokyo.ac.jp}
}

\crefname{assumption}{Assumption}{Assumptions}
\crefname{figure}{Fig.}{Figs.}
\crefname{ineq}{Ineq.}{Ineqs.}
\crefname{enumi}{}{}
\crefname{table}{Tab.}{Tabs.}

\creflabelformat{ineq}{#2{\upshape(#1)}#3}

\theoremstyle{plain}
\newtheorem{theorem}{Theorem}[section]

\newtheorem{lemma}[theorem]{Lemma}

\theoremstyle{definition}
\newtheorem{assumption}[theorem]{Assumption}

\newtheorem{setting}[theorem]{Setting}

\tcbset{
  breakable,
  boxsep=0mm,
  boxrule=0pt,
  colframe=white,
  left=0.5mm,
  right=0.5mm
}

\allowdisplaybreaks

\newcommand{\warningfilter}[1]{\texorpdfstring{#1}{TEXT}}

\RenewDocumentCommand{\ip}{s m m}{
  \IfBooleanTF{#1}
    {\left\langle #2, #3 \right\rangle}  
    {\langle #2, #3 \rangle}  
}
\renewcommand{\parallel}{\mathbin{\!/\mkern-5mu/\!}}

\newcommand{\one}{\bm{1}}

\renewcommand{\a}{\bm{a}}
\renewcommand{\b}{\bm{b}}

\renewcommand{\r}{\bm{r}}
\newcommand{\s}{\bm{s}}

\renewcommand{\v}{\bm{v}}
\newcommand{\w}{\bm{w}}
\newcommand{\x}{\bm{x}}
\newcommand{\y}{\bm{y}}
\newcommand{\z}{\bm{z}}

\newcommand{\hatf}{\hat{f}}
\newcommand{\hatg}{\hat{g}}

\newcommand{\I}{\bm{I}}

\newcommand{\V}{\bm{V}}
\newcommand{\W}{\bm{W}}

\newcommand{\balpha}{\bm{\alpha}}

\newcommand{\btheta}{\bm{\theta}}

\newcommand{\bxi}{\bm{\xi}}

\newcommand{\bSigma}{\bm{\Sigma}}

\newcommand{\bnabla}{\bm{\nabla}}

\newcommand{\bbE}{\mathbb{E}}
\newcommand{\bbN}{\mathbb{N}}
\newcommand{\bbP}{\mathbb{P}}
\newcommand{\bbR}{\mathbb{R}}

\newcommand{\calD}{\mathcal{D}}

\newcommand{\calN}{\mathcal{N}}
\newcommand{\calO}{\mathcal{O}}
\newcommand{\calS}{\mathcal{S}}
\newcommand{\calL}{\mathcal{L}}

\DeclareMathOperator{\erfc}{erfc}
\DeclareMathOperator{\sgn}{sgn}
\DeclareMathOperator{\SE}{SE}

\newcommand{\tildeOmega}{\tilde{\Omega}}
\newcommand{\tildeTheta}{\tilde{\Theta}}
\newcommand{\tildecalO}{\tilde{\calO}}

\newcommand{\dt}{\dd{t}}
\newcommand{\dx}{\dd{x}}
\newcommand{\dy}{\dd{y}}
\newcommand{\dz}{\dd{z}}

\newcommand{\adv}{\mathrm{adv}}
\newcommand{\gen}{\mathrm{gen}}

\newcommand{\barell}{\Bar{\ell}}
\newcommand{\Dadv}{\calD^\adv}
\newcommand{\xadv}{\x^\adv}
\newcommand{\yadv}{y^\adv}
\newcommand{\bthetaVa}{\btheta_{\V,\a}}
\newcommand{\bthetaWb}{\btheta_{\W,\b}}
\newcommand{\Cthr}{C_{\mathrm{thr}}}

\begin{document}

\maketitle

\begin{abstract}
Adversarial examples have raised several open questions, such as why they can deceive classifiers and transfer between different models. A prevailing hypothesis to explain these phenomena suggests that adversarial perturbations appear as random noise but contain class-specific features. This hypothesis is supported by the success of perturbation learning, where classifiers trained solely on adversarial examples and the corresponding \textit{incorrect labels} generalize well to correctly labeled test data. Although this hypothesis and perturbation learning are effective in explaining intriguing properties of adversarial examples, their solid theoretical foundation is limited. In this study, we theoretically explain the counterintuitive success of perturbation learning. We assume wide two-layer networks and the results hold for any data distribution. We prove that adversarial perturbations contain sufficient class-specific features for networks to generalize from them. Moreover, the predictions of classifiers trained on mislabeled adversarial examples coincide with those of classifiers trained on correctly labeled clean samples. The code is available at \url{https://github.com/s-kumano/perturbation-learning}.
\end{abstract}

\section{Introduction}
Adversarial examples~\cite{AE}, which are imperceptibly perturbed inputs designed to deceive machine learning models, have raised significant concerns about the robustness and reliability of these models. Despite their importance, the underlying mechanisms of adversarial examples are not yet fully understood. A prevailing hypothesis to explain the intriguing properties of adversarial examples is the ``feature hypothesis''~\cite{NRF}. This hypothesis posits that adversarial perturbations, while appearing as imperceptible noise to humans, contain class-specific features. The feature hypothesis provides a unified explanation for several puzzling phenomena associated with adversarial examples, such as their ability to deceive classifiers, transferability across models, and so on~(cf. \cref{sec:related-work-1}).

\begin{figure}
\centering
\includegraphics[width=\textwidth]{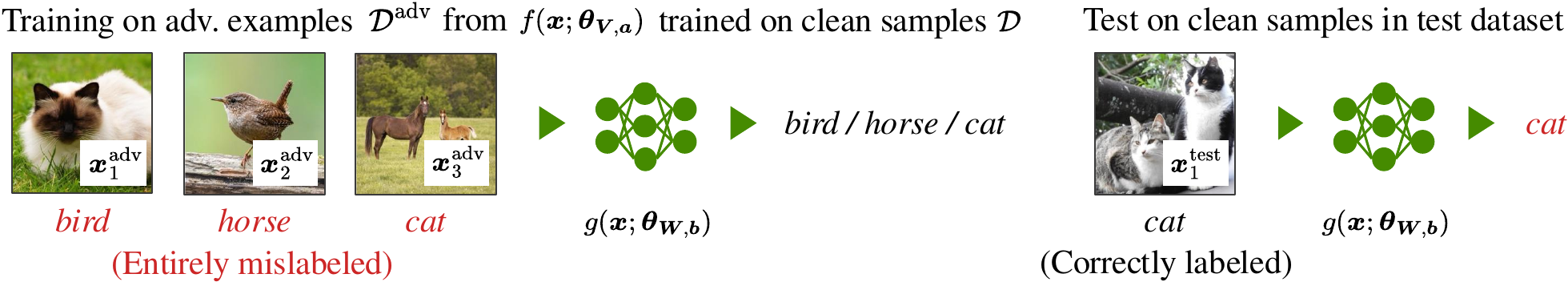}
\caption{Counterintuitive generalization of perturbation learning.\footref{footnote} A classifier $g$ is trained solely on mislabeled adversarial examples $\Dadv:=\{(\xadv_n,\yadv_n)\}^N_{n=1}$. These examples $\xadv_n$ are generated to mislead a classifier $f$, which is trained on correctly labeled clean samples $\calD:=\{(\x_n,y_n)\}^N_{n=1}$, into predicting $\yadv_n$~($\ne y_n$). Surprisingly, despite being trained only on mislabeled data, the classifier $g$ generalizes well to clean test samples. This counterintuitive result suggests that adversarial perturbations contain label-aligned class features, enabling the classifier $g$ to generalize from them.}
\label{fig:intro}
\end{figure}

Perturbation learning~\cite{NRF} provides empirical evidence supporting the feature hypothesis. In this learning, classifiers are trained \textit{solely} on adversarial examples that are \textit{mislabeled} in human perception,\footnote{\label{footnote}This is the critical difference between perturbation learning and adversarial training or training with noisy labels. Perturbation learning shows the learnability \textit{solely} from adversarial examples~(e.g., a cat adversarial image) that \textit{always} have \textit{incorrect} labels~(e.g., the bird label) to classify clean test images with the correct labels~(i.e., bird clean images to the bird class). Perturbation learning does not aim to learn robustly against adversarial examples or noisy labels. Refer to Appendix A in~\cite{EX} for further clarifications.} yet they demonstrate remarkable generalization to clean test data~(\cref{fig:intro}). For example, classifiers achieved 77\% accuracy on the correctly labeled clean test dataset of CIFAR-10~\cite{CIFAR10}, even though they were trained on entirely mislabeled adversarial examples~(e.g., a cat adversarial image labeled as a bird)~\cite{EX}. This surprising result suggests that adversarial perturbations encode class-relevant features that enable classifiers to learn meaningful representations. However, despite the empirical support, the theoretical foundations of the feature hypothesis and perturbation learning remain limited. While a recent study~\cite{EX} provided theoretical justifications, their results rely on stringent assumptions about data distribution, perturbation design, training procedure, and model architectures.

In this study, we theoretically address the understanding and justification of the feature hypothesis and perturbation learning. First, to support the feature hypothesis, we show that adversarial perturbations, while appearing as random noise, are parallel to the weighted sum of all training samples. This result suggests that a single perturbation derived from a classifier and input can potentially contain information about the entire training dataset. In particular, for some specific cases~(e.g., when training samples are mutually orthogonal), perturbations include all training data and labels without loss of information. We then reveal that class features within perturbations enable classifiers to generalize from them. Specifically, under three mild conditions, the predictions of a classifier trained on adversarial perturbations are consistent with those of a classifier trained on correctly labeled clean samples. These three conditions can be interpreted from geometric and quantitative perspectives. Finally, we demonstrate that under similar conditions, the prediction agreement is observed between a classifier trained on mislabeled adversarial examples and one trained on correctly labeled clean samples, justifying the empirical success of perturbation learning.

Our analysis assumes two-layer neural networks with sufficient width but does not impose any assumptions on data distribution, which is a substantial progress from prior work~\cite{EX} that considered mutually orthogonal training samples. In addition, our perturbation design, training procedure, activation functions, and bias availability are milder. In short, as shown in \cref{tab:comparison}, except for the wide width assumption, our analysis requires milder conditions than prior work. Our contributions can be summarized as follows:

\begin{itemize}
\item We provide a theoretical justification for the feature hypothesis and perturbation learning using wide two-layer neural networks, considering any data distribution and realistic problem settings. Except for the wide width, our assumptions are substantially milder than~\cite{EX}.

\item We demonstrate that adversarial perturbations are parallel to the weighted sum of training samples, suggesting that a single perturbation can potentially contain information about the entire training dataset. This result supports the feature hypothesis.

\item We prove that under three mild conditions, the predictions of a classifier trained on perturbations are consistent with those of a classifier trained on correctly labeled clean samples. Moreover, under similar conditions, the prediction agreement between a classifier trained on mislabeled adversarial samples and one trained on clean samples is observed, providing a theoretical justification for the empirical success of perturbation learning.
\end{itemize}

\section{Background and Related Work}

\subsection{Feature Hypothesis and Perturbation Learning}
\label{sec:related-work-1}
It has been hypothesized that adversarial perturbations contain class-specific features, although appearing as random noise~\cite{NRF}. This hypothesis, or feature hypothesis, offers a unified explanation for several open questions related to adversarial examples. For example, misclassification by classifiers and transferability across models~\cite{AE,FGSM} can be attributed to the response to features within perturbations. Furthermore, according to this hypothesis, adversarially robust models achieve robustness by discarding brittle yet predictive features and focusing on more stable and semantically meaningful features. This interpretation explains the phenomena observed with robust models, such as the trade-off between accuracy and robustness~\cite{su2018robustness,tsipras2019robustness,TRADES,raghunathan2019adversarial,raghunathan2020understanding,yang2020closer,mehrabi2021fundamental,dobriban2023provable}, perceptually-aligned gradients~\cite{etmann2019connection,zhang2019interpreting,santurkar2019image,kaur2019perceptually,engstrom2019adversarial,chalasani2020concise,RATIO,srinivas2023models,tsipras2019robustness,aggarwal2020benefits}, and enhanced transfer learning capabilities~\cite{aggarwal2020benefits,salman2020adversarially,deng2021adversarial,utrera2021adversarially}.

Perturbation learning\footref{footnote}~\cite{NRF} provides empirical support for the feature hypothesis. In perturbation learning, the dataset appears entirely mislabeled to human perception. However, the hypothesis suggests that adversarial perturbations in the dataset include label-aligned class features. Indeed, it has been observed that classifiers trained through perturbation learning can extract generalizable features from these perturbations and achieve high test accuracy~(e.g., 92\% for MNIST~\cite{MNIST}, 54\% for Fashion-MNIST~\cite{FMNIST}, and 77\% for CIFAR-10~\cite{CIFAR10}), empirically justifying the feature hypothesis~\cite{NRF,EX}.

While the feature hypothesis and perturbation learning are empirically effective in understanding adversarial examples, their theoretical foundations are very limited. Only one recent study~\cite{EX} theoretically demonstrated that perturbations contain class features and that classifiers can generalize from them. However, their results relied on stringent conditions~(e.g., mutually orthogonal training samples), which might not fully explain the success of perturbation learning in diverse settings.

In this study, for wide two-layer networks, we obtain results equivalent to those in~\cite{EX} under more relaxed conditions~(cf. \cref{sec:comparison}). We provide the first theoretical justification for the feature hypothesis and perturbation learning under any data distribution and in a mild training setting.

\subsection{Theoretical Framework: Lazy Training}
Theoretical analysis of neural networks is generally challenging due to the non-convex nature of the loss surface. To address this, recent studies have focused on the lazy training regime, where the parameters of neural networks hardly change during training~\cite{li2018learning,cao2019generalization,chizat2019lazy,geiger2020disentangling,zou2020gradient,cao2020generalization,ji2019polylogarithmic,woodworth2020kernel,montanari2022interpolation}. In this regime, neural networks behave almost linearly around their initialization, simplifying the learning dynamics. One of the key observation in lazy training is that, in wide two-layer neural networks, most derivatives of hidden outputs through (Leaky-) ReLU activation remain constant during training~\cite{li2018learning}, which forms the basis of our theoretical framework~(cf. \cref{sec:sketch}). This observation has been extended to show that the neural tangent kernel remains invariant during training~\cite{NTK,du2019gradient1,allen2019convergence,arora2019fine,du2019gradient2,lee2019wide}.

In contrast, the feature learning regime, where parameters move significantly away from their initialization, has been explored in various studies~\cite{chizat2019lazy,geiger2020disentangling,woodworth2020kernel}. Prior work on justifying perturbation learning~\cite{EX} employs the feature learning regime, building on related findings in this area~\cite{lyu2020gradient,ji2020directional,frei2023implicit}. In our study, we adopt the lazy training regime and relax several conditions assumed in previous work~\cite{EX} by introducing a wide width assumption~(cf. \cref{tab:comparison}). This adjustment is enabled by differences in the theoretical tools used.

\begin{table}
\centering
\caption{Comparison with existing work~\cite{EX}. With a wide network assumption, we improve the existing results from the perspective of data distribution, perturbation design, training time, loss function, and network architecture. Note that the non-bias and leaky-ReLU assumptions of~\cite{EX} are critical for deriving their results. A detailed comparison can be found in \cref{sec:comparison}.}
\begin{tabular}{lll}
  \toprule
  & \cite{EX} & Ours \\ \midrule
  Training samples & Mutually orthogonal
  & \textbf{Any} \\
  Perturbation type & Oracle-based
  & \textbf{Standard gradient-based}  \\
  Perturbation budget & Unrealistically tight & \textbf{Any} \\
  Training time & Infinite & \textbf{Any} \\
  Loss function & Exponential or logistic & \textbf{Differentiable, non-decreasing} \\
  Network bias & Not available & \textbf{Available} \\
  Activation & Leaky-ReLU & \textbf{ReLU and Leaky-ReLU} \\
  Network width & \textbf{Any} & Sufficiently wide (but finite) \\
  Theoretical framework & Feature learning & Lazy training \\ \midrule
  Common & \multicolumn{2}{l}{Binary classification, two-layer network, gradient flow} \\
  \bottomrule
\end{tabular}
\label{tab:comparison}
\end{table}

\section{Theoretical Results}

\textbf{Notation.}
For $n \in \bbN$, let $[n] := \{1,\ldots,n\}$. For $\z_1, \z_2 \in \bbR^d$, we denote the Euclidean norm by $\|\z_1\|$ and the inner product by $\ip{\z_1}{\z_2}$. Vectors $\z_1$ and $\z_2$ are called parallel and are denoted by $\z_1 \parallel \z_2$ if there exists $C \in \bbR$ such that $\z_1 = C \z_2$. Let $\calN(\mu,\sigma^2)$ be the Gaussian distribution with mean $\mu \in \bbR$ and variance $\sigma^2 \geq 0$ and $U(\calS)$ be the uniform distribution on a set $\calS \subset \bbR$. We use $\Omega(\,\cdot\,)$, $\Theta(\,\cdot\,)$, and $\calO(\,\cdot\,)$ only to hide constant factors, and $\tildeOmega(\,\cdot\,)$, $\tildeTheta(\,\cdot\,)$, and $\tildecalO(\,\cdot\,)$ to hide polylogarithmic factors.

\subsection{Problem Setup}
\label{sec:problem-setup}
In this study, we consider the dynamics of perturbation learning in binary classification problem with a two-layer neural network trained by gradient flow. First, we formally define the perturbation learning framework. The outline of perturbation learning is as follows: (i)~train a classifier on correctly labeled clean samples, (ii)~create adversarial samples based on the trained classifier, and (iii)~train another classifier on the mislabeled adversarial samples.

\textbf{Network trained on correctly labeled clean samples.}
We consider a two-layer neural network $f: \bbR^d \to \bbR$. Let $\V := (\v_1, \ldots, \v_m)^\top \in \bbR^{m \times d}$ and $\a := (a_1, \ldots, a_m)^\top \in \bbR^m$ be the hidden weight and bias, respectively. We also describe $\V := (V_{ij})_{1 \leq i \leq m, 1 \leq j \leq d}$. Let $\balpha := (\alpha_1, \ldots, \alpha_m)^\top  \in \bbR^m$ be the readout weight. While $\V$ and $\a$ are trainable, $\balpha$ is fixed during training. Denote the trainable parameters by $\bthetaVa := \{\V, \a\}$. We initialize $V_{ij} \sim \calN(0,1/d)$, $a_i \sim \calN(0,1)$, and $\alpha_i \sim \calN(0,1/m)$ for each $i \in [m]$ and $j \in [d]$. The activation function is either ReLU or Leaky-ReLU $\phi(x) := \max(\gamma x, x)$ for $\gamma \in [0,1)$. Finally, the network is given by $f(\x;\bthetaVa) := \sum^m_{i=1} \alpha_i \phi(\ip{\v_i}{\x} + a_i)$.

\textbf{Network trained on mislabeled adversarial samples.}
Similarly to $f$, we define a network trained on mislabeled adversarial samples as $g(\x;\bthetaWb) := \sum^m_{i=1} \beta_i \phi(\ip{\w_i}{\x} + b_i)$. Note that the initializations of $f$ and $g$ are independent.

\textbf{Loss function.}
We consider a differentiable, non-decreasing loss function $\ell: \bbR \to \bbR$, satisfying $\ell'(z) \geq 0$ for any $z \in \bbR$. Examples of such loss functions include the identity loss $\ell(z) := z$, exponential loss $\ell(z) := \exp(z)$, and logistic loss $\ell(z) := \ln(1 + \exp(z))$.

\textbf{Training.}
We here describe the training process of the network $f$ on correctly labeled clean samples. The training of $g$ is similarly defined. Let $\calD := \{(\x_n,y_n)\}^N_{n=1} \subset \bbR^d \times \{\pm 1\}$ be a correctly labeled training dataset. The loss over $\calD$ is defined as $\calL(\bthetaVa;\calD) := (1/N) \sum^N_{n=1} \ell(-y_nf(\x_n;\bthetaVa))$. The network parameters are updated by gradient flow $\dv*{\btheta_{\V,\a}(t)}{t} := - \pdv*{\calL(\btheta_{\V,\a}(t);\calD)}{\btheta_{\V,\a}}$, where $t \geq 0$ is the training time. We consider $T_f > 0$ training steps, producing $f(\,\cdot\,;\bthetaVa(T_f))$. For notational simplicity, we write $f(\,\cdot\,;t) := f(\,\cdot\,;\bthetaVa(t))$.

Note that we do not consider whether $f(\,\cdot\,;T_f)$ perfectly classify $\calD$. We discuss whether the classifier $g$, trained on adversarial examples crafted via $f(\,\cdot\,;T_f)$, can mimic the predictions of $f(\,\cdot\,;T_f)$.

\textbf{Adversarial perturbations.}
We consider a single-step gradient-based perturbation, which is a common perturbation design~\cite{FGSM}. An adversarial example $\xadv_n \in \bbR^d$ and its corresponding adversarial perturbation $\r_n \in \bbR^d$ are defined as follows:
\begin{align}
  \label{eq:AP}
  \xadv_n &:= \x_n + \r_n, &
  \r_n &:= -\epsilon
  \frac{ \bnabla_{\x_n} \ell(-\yadv_nf(\x_n;T_f)) }
  { \| \bnabla_{\x_n} \ell(-\yadv_nf(\x_n;T_f)) \| },
\end{align}
where $\epsilon > 0$ is the perturbation constraint and $\yadv_n \in \{\pm1\}$ is the target label. The adversarial perturbation $\r_n$ on $\x_n$ is designed to increase $\yadv_n f(\xadv_n;T_f)$ under the constraint $\|\r_n\| \leq \epsilon$.

\textbf{Mislabeled dataset.}
We consider two configurations of a dataset $\Dadv$ for training $g$. First, we follow the original perturbation learning approach, where classifiers are trained on adversarial perturbations superposed on natural images, i.e., $\Dadv := \{(\xadv_n, \yadv_n)\}^N_{n=1}$. This setting helps to understand the prior perturbation learning process. Second, we directly consider learning from perturbations rather than adversarial examples, i.e., $\Dadv := \{(\r_n, \yadv_n)\}^N_{n=1}$. This setting directly addresses the question of whether classifiers can generalize from class features in perturbations.

\textbf{Summary.}
The problem setting is summarized as follows:

\begin{tcolorbox}
\begin{setting}[Perturbation learning]
\label{setting}
Independently initialize $V_{ij} \sim \calN(0,1/d)$, $W_{ij} \sim \calN(0,1/d)$, $a_i \sim \calN(0,1)$, $b_i \sim \calN(0,1)$, $\alpha_i \sim \calN(0,1/m)$, and $\beta_i \sim \calN(0,1/m)$ for each $i \in [m]$ and $j \in [d]$. Train a two-layer neural network $f$ parameterized by $\bthetaVa$ with~($\gamma$-scaled Leaky-) ReLU on a dataset $\calD := \{(\x_n,y_n)\}^N_{n=1}$ using gradient flow with a loss $\calL(\btheta_{\V,\a};\calD)$ for training time $T_f > 0$. Create a dataset $\Dadv$ by one of the following procedures with $\{\yadv_n\}^N_{n=1} \in \{\pm1\}^N$:
\begin{align}
  & \text{Scenario (a)} 
  & \Dadv &:= \{(\r_n, \yadv_n)\}^N_{n=1}, \\
  & \text{Scenario (b)}
  & \Dadv &:= \{(\xadv_n, \yadv_n)\}^N_{n=1}.
\end{align}
Train a two-layer neural network $g$ parameterized by $\btheta_{\W,\b}$ on the dataset $\Dadv$ using gradient flow with a loss $\calL(\btheta_{\W,\b};\Dadv)$ for training time $T_g > 0$.
\end{setting}
\end{tcolorbox}

Our interests are (i)~the relationship between perceptually-noise-like adversarial perturbations $\{\r_n\}^N_{n=1}$ and clean training samples $\{(\x_n, y_n)\}^N_{n=1}$~(cf. \cref{th:AP}), and (ii)~whether the classifier $g(\,\cdot\,;T_g)$ trained on the adversarial perturbations or samples $\Dadv$ can mimic the predictions of the classifier $f(\,\cdot\,;T_f)$ trained on the clean samples $\calD$~(cf. \cref{th:PL-a,th:PL-b}).

\subsection{Main Results}
For $\z_1, \z_2 \in \bbR^d$, we use $\Phi(\z_1, \z_2) \in (\gamma (1+\gamma) / 2, (1+\gamma) / 2]$ defined as~(cf. \cref{le:expected-derivatives-2}):
\begin{align}
  \label{eq:Phi}
  \Phi(\z_1, \z_2)
  := \bbE_{ \v\sim\calN(0,\I/d), a\sim\calN(0,1) }
     [ \phi'(\ip{\v}{\z_1}+a) \phi'(\ip{\v}{\z_2}+a) ],
\end{align}
where $\phi'(x) := \dv*{\phi(x)}{x}$. First, we introduce an assumption on network width. 

\begin{tcolorbox}
\begin{assumption}[Wide network]
\label{asm:width}
Network width $m$ satisfies
\begin{align}
  m > \tildecalO\qty(
    d^2 \qty{
      \frac{1}{N} \sum^N_{n=1} \qty(
        \int^{T_f}_0 \ell'( -y_n f(\x_n;t) ) \dt
        + \int^{T_g}_0 \ell'( -\yadv_n f(\x'_n;t) ) \dt
      )
    }^2
  ),
\end{align}
where $\x'_n := \r_n$ for Scenario (a) and $\x'_n := \xadv_n$ for Scenario (b) in \cref{setting}. In particular, $m > \tildecalO(d^2(T_f+T_g)^2)$ for $\ell(s) = s$.
\end{assumption}
\end{tcolorbox}

This assumption requires sufficiently large width $m$ that regularizes the variations in parameters and forms the basis of lazy training~(cf. \cref{sec:sketch}). The width is always required to grow with the speed of the squared input dimension $d^2$. The relationship between the width and two training times, $T_f$ and $T_g$, depends on the training set $\{(\x_n,y_n)\}^N_{n=1}$ and loss function $\ell$. For example, if the training set is easily separable and the loss has an exponential tail, the derivative of the loss function might decrease rapidly with training time $t$ and small $m$ is enough to satisfy the assumption. For the identity loss, $m$ is consistently required to satisfy $\tildeOmega(d^2(T_f+T_g)^2)$. Note that the required values of $T_f$ and $T_g$~(and the corresponding $m$) for a desirable loss value remain an open question in the community. Our experimental results show that $m \approx 100$ is sufficient to verify our theorems for high-dimensional Gaussian distributions. Under this assumption, we consider the direction of the adversarial perturbation.

\begin{figure}
\centering
\includegraphics[width=\textwidth]{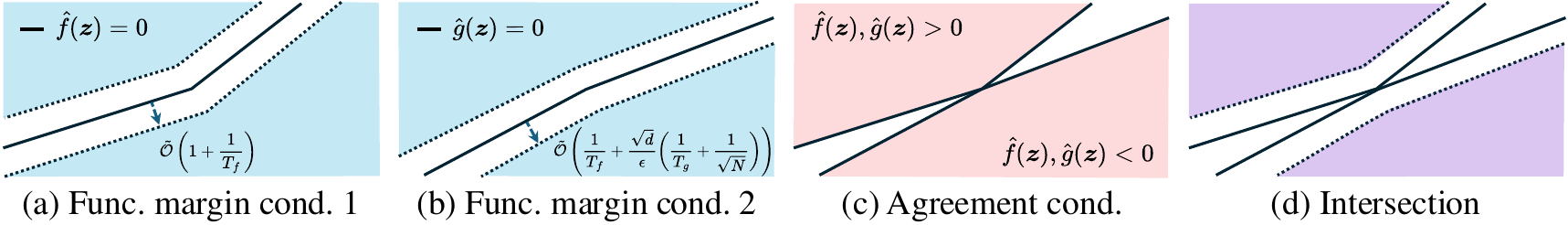}
\caption{The regions where \cref{ineq:cond-1,ineq:cond-2-a,eq:cond-3-a} hold~(colored areas) and their intersection.}
\label{fig:geometry}
\end{figure}

\begin{tcolorbox}
\begin{restatable}[Direction of adversarial perturbation]{theorem}{AP}
\label{th:AP}
Let $\delta = \Theta(1)$ be a small positive number. Under \cref{asm:width}, for any $n \in [N]$, with probability at least $1 - \delta$, the adversarial perturbation $\r_n$ is parallel to the weighted sum of training samples as follows:
\begin{align}
  \r_n \parallel \frac{1}{N} \sum^N_{k=1} y_k \Phi(\x_n, \x_k) \x_k
  \int^{T_f}_0 \ell'(-y_kf(\x_k;t)) \dt + \bxi_n,
\end{align}
where $\bxi_n$ satisfies $\|\bxi_n\| = \tildecalO(1)$. In particular, for $\ell(s) = s$,
\begin{align}
  \r_n \parallel \frac{T_f}{N} \sum^N_k y_k \Phi(\x_n, \x_k) \x_k + \bxi_n.
\end{align}
\end{restatable}
\end{tcolorbox}

Note that the confidence level $\delta$ only logarithmically affects the norm of the remainder term $\xi_n$. This theorem indicates that the direction of a single perturbation can be represented as the weighted sum of $y_k \x_k$ and remainder term $\xi_n$. Interestingly, this result suggests that a single perturbation derived from a classifier and sample can potentially contain information about the entire training dataset $\{(\x_n,y_n)\}^N_{n=1}$. Particularly, in some cases~(e.g., training samples are mutually orthogonal), $y_k \x_k$ are not cancelled out by each other, and thus the single perturbation $\r_n$ contains all training data and labels without loss of information.\footnote{Recall $\Phi(\x_n, \x_k) > 0$.} These results theoretically support the feature hypothesis. Consider the case with the identity loss. While the norm of the first term is $\calO(T_f\sqrt{d})$, the norm of the remainder is constrained to $\tildecalO(1)$, suggesting that larger training time $T_f$ and input dimension $d$ strengthen the alignment between the perturbation and weighted sum.

Then, we consider the learning solely from these perturbations. The following theorem is a special case of \cref{th:PL-a-general}, which addresses a broader loss class and any sampling of $\yadv_n \in \{\pm1\}$. 

\begin{tcolorbox}
\begin{restatable}[Perturbation learning, Scenario (a), special case of \cref{th:PL-a-general}]{theorem}{PLa}
\label{th:PL-a}
Consider Scenario~(a) in \cref{setting}. Assume $\ell(s) = s$ and $\yadv_n \sim U(\{\pm1\})$ for every $n \in [N]$. Let $\delta = \Theta(1)$ be a small positive number and
\begin{align}
  \label{eq:hat-f-and-g-a}
  &
  \hatf(\z)
  := \frac{1}{N} \sum^N_{n=1} y_n \Phi(\x_n, \z) \ip{\x_n}{\z}, &
  \hatg_a(\z)
  := \frac{1}{N^2} \sum^N_{n=1} \Phi(\r_n,\z) \sum^N_{k=1} 
  y_k \Phi(\x_n, \x_k) \ip{\x_k}{\z}.
\end{align}
Under \cref{asm:width}, for any $\z \in \bbR^d$, if
\begin{align}
  \label[ineq]{ineq:cond-1}
  & \text{\emph{(Functional margin condition 1)}} &
  |\hatf(\z)| &> \tildecalO\qty( 1 + \frac{1}{T_f} ), \\
  \label[ineq]{ineq:cond-2-a}
  & \text{\emph{(Functional margin condition 2)}} &
  |\hatg_a(\z)| &> \tildecalO\qty( \frac{1}{T_f}
    + \frac{\sqrt{d}}{\epsilon}
      \qty( \frac{1}{T_g} + \frac{1}{\sqrt{N}} ) ), \\
  \label{eq:cond-3-a}
  & \text{\emph{(Agreement condition)}} &
  \sgn(\hatf(\z)) &= \sgn(\hatg_a(\z)),
\end{align}
then, with probability at least $1 - \delta$, $\sgn(f(\z;T_f)) = \sgn(g(\z;T_g))$ holds. 
\end{restatable}
\end{tcolorbox}

Note that the confidence level $\delta$ only logarithmically affects the right terms of \cref{ineq:cond-1,ineq:cond-2-a}, which is why these terms appear independent of $\delta$. This theorem states that the predictions of a classifier $g$ trained solely on adversarial perturbations $\{(\r_n,\yadv_n)\}^N_{n=1}$ coincide with those of a classifier $f$ trained on standard training samples $\{(\x_n,y_n)\}^N_{n=1}$ if the three conditions hold. The two functions, $\hatf$ and $\hatg$, which govern these conditions, can be viewed as key components that significantly influence the predictions of $f$ and $g$~(cf. \cref{sec:sketch}). The conditions can be interpreted as follows.

\textbf{Geometrical perspective.}
The functional margin conditions, \cref{ineq:cond-1,ineq:cond-2-a}, require the functional margins of $\hatf$ and $\hatg$ to exceed certain thresholds. In the input space $\z \in \bbR^d$, these conditions correspond to regions far from the decision boundaries $\hatf(\z) = 0$ and $\hatg(\z) = 0$~(\cref{fig:geometry}(a) and (b)). In a $d$-dimensional space, $L_2$-distance scales with $\sqrt{d}$, making the right terms of \cref{ineq:cond-1,ineq:cond-2-a} relatively small when the perturbation size $\epsilon = \Theta(\sqrt{d})$; hence, a larger $d$ facilitates the satisfaction of these conditions. Furthermore, \cref{eq:cond-3-a} necessitates the agreement of the signs of these two decision boundaries~(\cref{fig:geometry}(c)). Consequently, the region where all conditions hold, i.e., where the prediction match occurs, can be characterized by the two piecewise linear functions~(\cref{fig:geometry}(d)).

\textbf{Quantitative perspective~(functional margin conditions).}
A large perturbation size $\epsilon$ facilitates the satisfaction of \cref{ineq:cond-2-a}. In high-dimensional spaces, achieving the required margin conditions demands perturbations of at least $\Omega(\sqrt{d})$ in magnitude, aligning with empirical scaling laws for $L_2$ perturbations. However, increasing $\epsilon$ alone is insufficient for the satisfaction because the right term of \cref{ineq:cond-2-a} has an $\epsilon$-irrelevant term $\tildecalO(1/T_f)$. Assume $\epsilon = \Theta(\sqrt{d})$. The absolute values of $\hatf$ and $\hatg$ grow with $\Theta(d)$ due to $\ip{\x}{\z}$, while the right terms of \cref{ineq:cond-1,ineq:cond-2-a} are independent of $d$. Thus, a larger $d$ consistently facilitates the satisfaction. The training times $T_f$ and $T_g$ can reduce the right terms, but these terms contain time-independent terms $\tildecalO(1)$ and $\tildecalO(1/\sqrt{N})$, indicating that longer training times do not necessarily guarantee the satisfaction. A large sample size $N$ also helps to satisfy \cref{ineq:cond-2-a}, but similarly, it is not sufficient. In summary, while a larger input dimension $d$ consistently support the success of perturbation learning, a larger perturbation size $\epsilon$, sample size $N$, and training times $T_f$, $T_g$ provide partial, but not definitive, benefits.

\textbf{Quantitative perspective~(agreement condition).}\footnote{One can analyze the agreement condition by using the relationship between $\Phi$ and the arc-cosine kernel~\cite{arccosine}.}
It is difficult to interpret the agreement condition, \cref{eq:cond-3-a}, in its current form. We consider the following sufficient condition~(cf. \cref{le:sufficient-agree-cond}):
\begin{align}
  \label[ineq]{ineq:cond-3-another}
  \frac{ | \sum^N_{n=1} y_n \ip{\x_n}{\z} | }
       { \max_{ \x \in \{\x_1,\ldots,\x_N,\z\} }
         \sum^N_{n=1} \lambda(\x_n, \x) |\ip{\x_n}{\z}| }
  > \frac{1 - \gamma}{1 + \gamma}
  \qquad \Rightarrow \qquad \text{\cref{eq:cond-3-a}}, \\
  \lambda(\z_1, \z_2) := 1 -
  \sqrt{ \frac{e}{2\pi} \frac{ \|\z_1\|^2 \|\z_2\|^2 - \ip{\z_1}{\z_2}^2
  + d \|\z_1 - \z_2\|^2 }{ \|\z_1\|^2 \|\z_2\|^2 + \ip{\z_1}{\z_2}^2
  + d \|\z_1 + \z_2\|^2 + 2d^2 } } = \Theta(1).
\end{align}
Note that the left term can exceed one as $\lambda(\z_1, \z_2)$ lies in $(0.34, 1]$. It is clear that a large negative slope of Leaky-ReLU $\gamma$ facilitates the satisfaction. The magnitude of the left term depends on the consistency of the correlation~(inner product) between $\z$ and $y_n \x_n$ for every $n$. For example, when $\z$ consistently exhibits a positive or negative correlation with $y_n \x_n$, the left term exceeds one, and the condition is satisfied. In contrast, if $\z$ positively correlates with half of the samples and negatively with the other half, the left term may output a small value, and the condition is not satisfied. In summary, the agreement condition depends on the consistency of the correlation between $\z$ and $y_n \x_n$.

Finally, we justify the success of perturbation learning in Scenario (b).

\begin{tcolorbox}
\begin{restatable}[Perturbation learning, Scenario (b), special case of \cref{th:PL-b-general}]{theorem}{PLb}
\label{th:PL-b}
Consider Scenario~(b) in \cref{setting}. Assume $\ell(s) = s$ and $\yadv_n \sim U(\{\pm1\})$ for every $n \in [N]$. Let $\delta = \Theta(1)$ be a small positive number and
\begin{align}
  \hatg_b(\z)
  := \frac{1}{N^2} \sum^N_{n=1} \Phi(\xadv_n,\z) \sum^N_{k=1} 
  y_k \Phi(\x_n, \x_k) \ip{\x_k}{\z}.
\end{align}
Under \cref{asm:width}, for any $\z \in \bbR^d$, if the functional margin condition 1~(\cref{ineq:cond-1}),
\begin{align}
  \label[ineq]{ineq:cond-2-b}
  & \text{\emph{(Func. margin cond. 2)}} &
  |\hatg_b(\z)| &> \tildecalO\qty( \frac{1}{T_f}
    + \frac{\sqrt{d}}{\epsilon}
      \qty( \frac{1}{T_g}
            + \frac{ \sqrt{ \sum^N_n (\ip{\x_n}{\z}+1)^2 } }{N} ) ), \\
  \label{eq:cond-3-b}
  & \text{\emph{(Agreement condition)}} &
  \sgn(\hatf(\z)) &= \sgn(\hatg_b(\z)),
\end{align}
then, with probability at least $1 - \delta$, $\sgn(f(\z;T_f)) = \sgn(g(\z;T_g))$ holds. 
\end{restatable}
\end{tcolorbox}

This is a special case of \cref{th:PL-b-general}, which addresses differentiable, non-decreasing losses and any sampling of $\yadv_n \in \{\pm1\}$. Similarly to \cref{th:PL-a}, this theorem states that the predictions of a classifier $g$ trained on randomly labeled adversarial examples $\{(\xadv_n,\yadv_n)\}^N_{n=1}$ coincide with those of a classifier $f$ trained on standard training samples $\{(\x_n,y_n)\}^N_{n=1}$ if the three conditions hold. Functional margin condition 1 is consistent with that in \cref{th:PL-a}, i.e., \cref{ineq:cond-1}. The definition of $\hatg_b(\z)$ is slightly different from $\hatg_a(\z)$ in \cref{th:PL-a}, with $\r_n$ replaced by $\xadv_n$. Due to this change in the definition of $\hatg_b(\z)$, functional margin condition 2, \cref{ineq:cond-2-b}, and the agreement condition, \cref{eq:cond-3-b}, slightly differ from those in \cref{th:PL-a}.

Assume $\epsilon = \Theta(\sqrt{d})$. Similarly to \cref{ineq:cond-2-a}, the left term of \cref{ineq:cond-2-b} grows with $\calO(d)$ due to the inner product. In contrast to \cref{ineq:cond-2-a}, the right term of \cref{ineq:cond-2-b} includes a term that grows with $\sqrt{ \sum^N_n (\ip{\x_n}{\z}+1)^2 } / N = \calO(d/\sqrt{N})$. This suggests that Scenario~(b) necessitates a larger sample size $N$ than Scenario~(a) to mitigate the effect of $d$.

Furthermore, \cref{eq:cond-3-a,eq:cond-3-b} hold simultaneously~(i.e., $\sgn(\hatf(\z)) = \sgn(\hatg_a(\z)) = \sgn(\hatg_b(\z))$) if $\sum^N_{k=1} y_k \Phi(\x_n,\x_k) \ip{\x_k}{\z}$ outputs the same sign for any $n$. This indicates that there might exist regions where the prediction match is observed regardless of the scenarios, and these regions are partially determined by the $n$ linear boundaries. Note that \cref{ineq:cond-3-another} also serves as a sufficient condition for \cref{eq:cond-3-b}, and the quantitative analysis for \cref{eq:cond-3-a} can be applied to \cref{eq:cond-3-b} as well.

\subsection{Sketch of Proof}
\label{sec:sketch}
In this section, for simplicity, we provide a sketch of the proof for \cref{th:AP,th:PL-a} with infinite network width $m \to \infty$, networks without biases, and identity loss. A proof for the general case can be found in \cref{sec:main-proof}.

\textbf{Lazy training.}
First, we introduce the concept of lazy training, where network parameters and outputs of hidden neurons change negligibly during training when the network width is sufficiently large~\cite{li2018learning,chizat2019lazy}. Since a readout weight $\alpha_i$ is sampled from $\calN(0,1/m)$, from gradient flow, for any $\z \in \bbR^d$,
\begin{align}
  \abs{ \ip*{ \int^{T_f}_0 \dv{\v_i(t)}{t} \dt }{\z} }
  = \abs{ \frac{1}{N} \sum^N_{n=1} y_n \alpha_i
          \ip{\x_n}{\z} \int^{T_f}_0 \phi'(\ip{\v_i(t)}{\x_n}) \dt }
  = \tildecalO\qty(\frac{dT_f}{\sqrt{m}}).
\end{align}
Therefore, as $m \to \infty$, the inner product between the time variation of a hidden parameter and an input approaches zero. This suggests that the sign of the output of a hidden neuron do not change:
\begin{align}
  \label{eq:tmp-bvvz}
  & \sgn(\ip{\v_i(T_f)}{\z})
  = \sgn\qty(\ip{\v_i(0)}{\z} + \ip*{ \int^{T_f}_0 \dv{\v_i(t)}{t} \dt }{\z})
  \xrightarrow{m\to\infty} \sgn(\ip{\v_i(0)}{\z}).
\end{align}
Therefore, $\phi'(\ip{\v_i(T_f)}{\z}) = \phi'(\ip{\v_i(0)}{\z})$. Recall $\phi(z) := \max(\gamma z, z)$.

\textbf{\cref{th:AP}.}
From the perturbation definition \cref{eq:AP}, the perturbation $\r_n$ is parallel to $\bnabla_{\x_n} f(\x_n;\V(T_f))$. Using $\phi'(\ip{\v_i(T_f)}{\z}) = \phi'(\ip{\v_i(0)}{\z})$,
\begin{align}
  \label{eq:tmp-qmnn}
  \r_n \parallel
  \bnabla_{\x_n} f(\x_n;\V(T_f))
  =& \sum^m_{i=1} \alpha_i \phi'(\ip{\v_i(0)}{\x_n})
     \qty(\v_i(0) + \int^{T_f}_0 \dv{\v_i(t)}{t} \dt).
\end{align}
The first term is constrained to $\tildecalO(1)$. Let $\Phi(\x_n, \x_k) := \bbE_{\v\sim\calN(0,1/d)} [\phi'(\ip{\v}{\x_n}) \phi'(\ip{\v}{\x_k})]$. Using $\sum^m_{i=1} \alpha^2_i \phi'(\ip{\v_i(0)}{\x_n}) \phi'(\ip{\v_i(0)}{\x_k}) \to \Phi(\x_n, \x_k)$ as $m \to \infty$, the second term becomes
\begin{align}
  \sum^m_{i=1} \alpha_i \phi'(\ip{\v_i(0)}{\x_n})
  \int^{T_f}_0 \dv{\v_i(t)}{t} \dt
  = \frac{T_f}{N} \sum^N_{k=1} y_k \Phi(\x_n, \x_k) \x_k.
\end{align}

\textbf{\cref{th:PL-a}.}
Similarly to the above, we can represent the adversarial perturbation $\r_n$ as follows:
\begin{align}
  \label{eq:tmp-aaqa}
  \r_n := \epsilon \yadv_n
  \frac{ \bnabla_{\x_n} f(\x_n;\V(T_f)) }{ \|\bnabla_{\x_n} f(\x_n;\V(T_f))\| }
  \approx
  \Omega\qty(\frac{\epsilon}{N\sqrt{d}}) \yadv_n
  \sum^N_{k=1} y_k \Phi(\x_n, \x_k) \ip{\x_k}{\z}.
\end{align}
Assuming $\sum^m_{i=1} \alpha_i \phi'(\ip{\v_i(0)}{\z}) \ip{\v_i(0)}{\z} = \tildecalO(1) \approx 0$ for simplicity, the network prediction $f(\z;\V(T_f))$ trained on $\{(\x_n,y_n)\}^N_{n=1}$ can be represented as follows:
\begin{align}
  f(\z;\V(T_f))
  =& \sum^m_{i=1} \alpha_i \phi'(\ip{\v_i(0)}{\z}) \ip{\v_i(T_f)}{\z}
  \approx
  \frac{T_f}{N} \sum^N_{n=1} y_n \Phi(\x_n, \z) \ip{\x_n}{\z},
\end{align}
and $\sgn(f(\z;\V(T_f))) = \sgn( \sum^N_{n=1} y_n \Phi(\x_n, \z) \ip{\x_n}{\z} )$. In addition, $g(\z;\W(T_f))$ trained on $\{(\r_n,\yadv_n)\}^N_{n=1}$ can be represented as follows:
\begin{align}
  g(\z;\W(T_g))
  \approx
  \Omega\qty(\frac{\epsilon T_g}{N^2\sqrt{d}})
  \sum^N_{n=1} \Phi(\r_n, \z) \sum^N_{k=1} y_k \Phi(\x_n,\x_k) \ip{\x_k}{\z},
\end{align}
and $\sgn(g(\z;\W(T_g))) = \sgn( \sum^N_{n=1} \Phi(\r_n, \z) \sum^N_{k=1} y_k \Phi(\x_n,\x_k) \ip{\x_k}{\z} )$. Thus, if the agreement condition \cref{eq:cond-3-a} holds, then $\sgn(f(\z;\V(T_f))) = \sgn(g(\z;\W(T_g)))$.

\textbf{Formal proof.}
In the above sketch of proof, we have introduced several approximations. Rigorous evaluations are provided in \cref{sec:main-proof}. For example, in the sketch, we assumed $m \to \infty$, ensuring that the signs of all hidden layer outputs remain unchanged. In contrast, the formal proof derives a bound on the width $m$ that ensures that the number of hidden neurons with flipped signs is at most $\calO(\sqrt{m})$, which makes the discussion~(e.g., about \cref{eq:tmp-bvvz,eq:tmp-qmnn}) more complicated. Moreover, in \cref{eq:tmp-aaqa}, we neglected the first term of \cref{eq:tmp-qmnn}, but the formal proof carefully considers the impact on the subsequent steps. The functional margin conditions arise from the evaluation of these remainder terms.

\subsection{Comparison with Prior Work and Limitations}
\label{sec:comparison}
In this section, we compare our results with~\cite{EX} and discuss the limitations of our work. In summary, our results justify the feature hypothesis and perturbation learning under substantially milder conditions than~\cite{EX}, except for network width~(cf. \cref{tab:comparison}). The assumption of wide two-layer networks is our main limitation.

\textbf{Goals, results, and tools.}
The goals of our work and~\cite{EX} are the same: justifying the feature hypothesis and perturbation learning. The conclusions drawn are also equivalent. However, our assumptions are much milder than theirs. This is due to the differences in the analytical approaches. While they leverage research on feature learning~\cite{lyu2020gradient,ji2020directional,frei2023implicit}, we utilize the concept of lazy training~\cite{li2018learning,chizat2019lazy}, which enables us to substantially relax the conditions.

\textbf{Data distribution.}
Prior work imposes a strong assumption that training samples with/without adversarial perturbations are mutually orthogonal, i.e., $\ip{\x_n}{\x_k} \approx 0$ and $\ip{\xadv_n}{\xadv_k} \approx 0$ for any $n \ne k$. This condition is stringent and is hard to hold for real-world datasets. Moreover, it may not even hold for data sampled from a zero-mean Gaussian distribution in some common situations~(e.g., the sample size is sufficiently larger than the dimension). We do not impose any assumptions on the data distribution. This is the first result that theoretically supports the feature hypothesis and the success of perturbation learning on realistic data distribution.

\textbf{Perturbation design.}
Prior work defined the perturbation form using the decision boundary of a classifier. However, this is not only uncommon but also theoretically computable only in limited problem settings. Additionally, they constrained perturbation size to $\epsilon = \Theta(\sqrt{d/N})$, which becomes unrealistically small for a large sample size and is far from the practical constraint $\epsilon = \Theta(\sqrt{d})$. We employ a single-step gradient-based method~\cite{FGSM}, which is commonly used in practice, and the perturbation constraint can be set arbitrarily.

\textbf{Training time, loss function, network bias, and activation.}
First, it should be noted that these constraints are critical for deriving the results in~\cite{EX}. This is because their theoretical framework~\cite{lyu2020gradient,ji2020directional,frei2023implicit} substantially requires the above conditions. We consider arbitrary training time, a wide class of loss functions, and (Leaky-) ReLU networks with bias availability. In contrast, they considered infinite training time, loss functions with exponential tails, homogeneous neural networks~(thus requiring no bias), and Leaky-ReLU networks~(the theorem becomes harder to hold as the negative slope of Leaky-ReLU approaches zero), which are essential for deriving their results~(cf. the proofs of Theorem 4.4 in \cite{lyu2020gradient} and the proof of Theorem 3.2 in \cite{frei2023implicit}).

\textbf{Limitations.}
Compared to \cite{EX}, our main limitation is the requirement for sufficient, though finite, network width. Moreover, our analysis is confined to two-layer networks, a common constraint in previous work. In practice, perturbation learning often employs deeper, and not necessarily wider, networks, which limits the direct applicability of our theoretical insights to more complex architectures. This assumption of a shallow network introduces another limitation. While deep neural networks typically capture high-level features from images and adversarial attacks are considered to exploit them, our framework focuses solely the low-level features~(i.e., $\x_n$ itself) in adversarial perturbations and their extraction through perturbation learning, as shown in \crefrange{th:AP}{th:PL-b}. Relaxing the shallow network constraint may allow us to capture a broader set of features present in adversarial perturbations. Despite these limitations, our work is the first to rigorously support the feature hypothesis and validate perturbation learning under realistic data distributions, perturbation designs, and training settings, marking a substantial advancement in the theoretical understanding of adversarial examples.

\section{Experiments}
\label{sec:experiments}
\begin{figure}
\centering
\includegraphics[width=\textwidth]{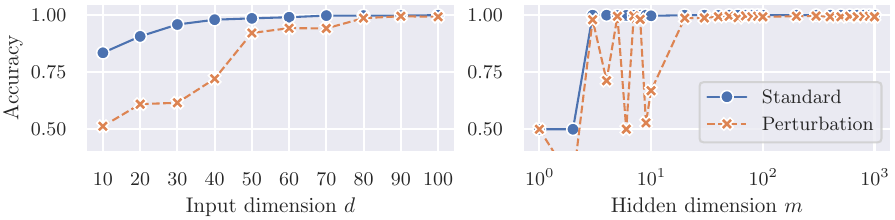}
\caption{Accuracy on the mean-shifted Gaussian dataset in Scenario~(a). The blue lines represent accuracy of the classifier $f$ on $\calD$, i.e., training accuracy. The orange lines represent accuracy of the classifier $g$ on $\calD$.}
\label{fig:shifted-gauss-a-mini}
\end{figure}

A comprehensive set of experiments conducted to validate our theorems can be found in \cref{sec:experiments-ap}. In this section, we briefly present two results that confirm \cref{th:PL-a}. As a training dataset $\calD := \{(\x_n, y_n)\}^N_{n=1}$, we employed a synthetic training dataset to easily change the input dimension, which effectively helps perturbation learning in both scenarios, as predicted by our theorems. Note that the perturbation learning on real-world datasets can be found in the literature~\cite{NRF,EX}. We generated synthetic data and labels from the mean-shifted Gaussian distribution as follows: $\{\x_n\}^N_{n=1}$ are independently sampled from $\calN(0.3 \times y_n \times \one, \I)$, and $y_n$ is set to one if $n \in [N/2]$ and minus one otherwise. The experimental settings are as follows: $d = 100$, $N = 1,000$, $m = 100$, $\gamma = 0$, $\ell(s) := s$, $\epsilon = 0.01$, and the number of training steps is set to 1,000 for both $f$ and $g$. The experimental results for perturbation learning under Scenario~(a) are shown in \cref{fig:shifted-gauss-a-mini}. A high input dimension facilitates the alignment between $f$ and $g$. Our theoretical results assume a wide network width, and \cref{fig:shifted-gauss-a-mini} indicates that a sufficiently large width consistently stabilize the alignment.

\section{Conclusion}
We provided a theoretical justification for perturbation learning and the feature hypothesis. We demonstrated that adversarial perturbations contain class-specific features sufficient for networks to generalize from. Moreover, we revealed that the predictions of a classifier trained solely on these perturbations or mislabeled adversarial examples coincide with those of a classifier trained on correctly labeled training samples under three mild conditions. Except for wide two-layer networks, our assumption is substantially milder than prior work~\cite{EX}.

\begin{ack}
S. Kumano was supported by JSPS KAKENHI Grant Number JP23KJ0789 and by JST, ACT-X Grant Number JPMJAX23C7, JAPAN. H. Kera was supported by JSPS KAKENHI Grant Number JP22K17962. T. Yamasaki was supported by JSPS KAKENHI Grant Number JP22H03640, JST ASPIRE Program Grant Number JPMJAP2303, and Institute for AI and Beyond of The University of Tokyo.
\end{ack}

\bibliography{
  bib/attack,
  bib/gradient-alignment,
  bib/implicit-bias,
  bib/NTK,
  bib/others,
  bib/perturbation-learning,
  bib/trade-off,
  bib/transfer-learning
}
\bibliographystyle{abbrv}

\appendix

\clearpage

\renewcommand{\theequation}{A\arabic{equation}}
\renewcommand{\thefigure}{A\arabic{figure}}
\renewcommand{\thetable}{A\arabic{table}}

\section{Comparison with Prior Work}
\label{sec:comparison-ap}
In this section, we compare our findings with those of prior work~\cite{EX} and highlight new insights beyond technical contributions.

\subsection{Feature Hypothesis~(\warningfilter{\cref{th:AP}})}
Our result offers a new insight into the alignment between perturbations and training samples through the residual term $\xi_n$. The direction of a perturbation vector comprises two components: a weighted sum of the training samples~(the main term) and a residual term. Our result suggests that as the input dimension increases, the residual term becomes smaller than the main term, thereby enhancing the alignment. In other words, perturbations more robustly encode class-specific features. This insight is unattainable in prior research due to the absence of a residual term in their limited problem setting.

Additionally, our finding suggests that extended training time reinforces the directional alignment between perturbations and training samples. This concept is supported by practical intuition and experience but not addressed in prior work.

Our result further introduces a coefficient, $\Phi(\x_n, \x_k)$, for each adversarial perturbation. The coefficient $\Phi(\x_n, \x_k)$ depends on the slope of the activation function and the similarity between $\x_n$ and $\x_k$~(cf. \cref{eq:Phi}). This suggests that training samples with higher similarity to each other exhibit a stronger influence within a perturbation. Although prior work includes similar coefficients, they cannot be explicitly computed.

\subsection{Perturbation Learning~(\warningfilter{\cref{th:PL-a,th:PL-b}})}
Our result establishes an explicit connection between the success of perturbation learning and training factors, such as training time $T$, perturbation size $\epsilon$, input dimension $d$, sample size $N$, and confidence level $\delta$. This result enhances our understanding of how these factors influence perturbation learning. For example, perturbation learning is more likely to succeed with a larger input dimension $d$ and sample size $N$. Notably, our findings indicate that while a larger $d$ consistently facilitates successful perturbation learning, a larger $N$ alone is insufficient. In contrast, existing research does not elucidate the roles of these variables, merely showing that success is achievable when both $d$ and $N$ approach infinity.

\section{Experimental Settings and Other Results}
\label{sec:experiments-ap}

\subsection{Datasets}
We utilized two synthetic datasets and two widely used datasets, MNIST~\cite{MNIST} and Fashion-MNIST~\cite{FMNIST}. The first synthetic dataset is derived from a zero-mean Gaussian distribution: $\{\x_n\}^N_{n=1}$ are independently sampled from $\calN(0, \I)$ and $\{y_n\}^N_{n=1}$ are independently sampled from $U(\{\pm1\})$. The second synthetic dataset is based on a mean-shifted Gaussian distribution: $\{\x_n\}^N_{n=1}$ are independently sampled from $\calN(0.3 \times y_n \times \one, \I)$ and $y_n$ is set to one for $n \in [N/2]$ and minus one otherwise. We used data only from classes 1 and 2 in MNIST~(i.e., digits 1 and 2) and those from classes 0 and 9 in Fashion-MNIST~(i.e., T-shirt and ankle boot). To measure the agreement ratio between network predictions from standard training and perturbation learning, for the real-world dataset cases, we used standard test datasets, and for the synthetic dataset cases, we used 1,000 samples independently and identically sampled according to the training data distribution.

\subsection{Settings}
In this section, for notational simplicity, we denote the number of training epochs in standard training by $T_f$ and in perturbation learning by $T_g$. Note that the original definitions of $T_f$ and $T_g$ are continuous training steps in gradient flow, i.e., gradient descent with an infinitely small learning rate~(cf. \cref{sec:problem-setup}), which is conceptually distinct from the discrete steps in gradient descent with finitely small learning rates in practice. In addition, we denote the learning rates in standard training and perturbation learning as $\eta_f$ and $\eta_g$, respectively.

\begin{figure}
\centering
\includegraphics[width=\textwidth]{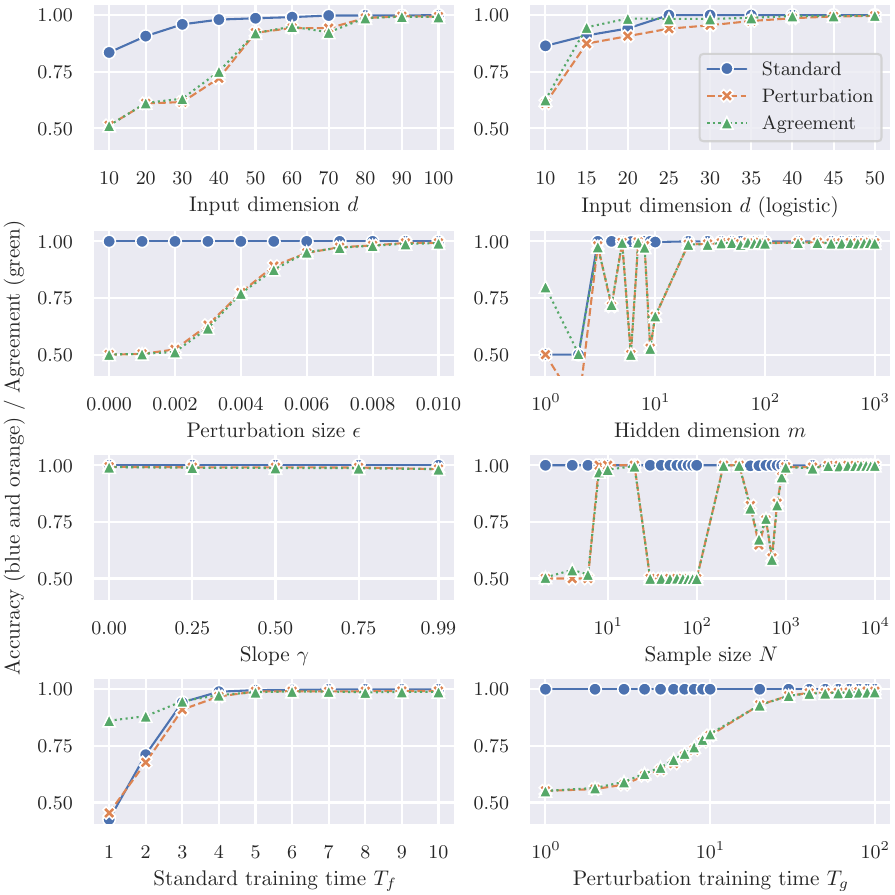}
\caption{Accuracy and agreement ratio on the mean-shifted Gaussian in Scenario~(a). The blue lines represent the accuracy of the classifier $f$ on $\calD := \{(\x_n,y_n)\}^N_{n=1}$, i.e., training accuracy. The orange lines represent the accuracy of the classifier $g$ on $\calD$. The green lines represent the prediction agreement between $f$ and $g$ on the test dataset.}
\label{fig:shifted-gauss-a}
\end{figure}

\begin{figure}
\centering
\includegraphics[width=\textwidth]{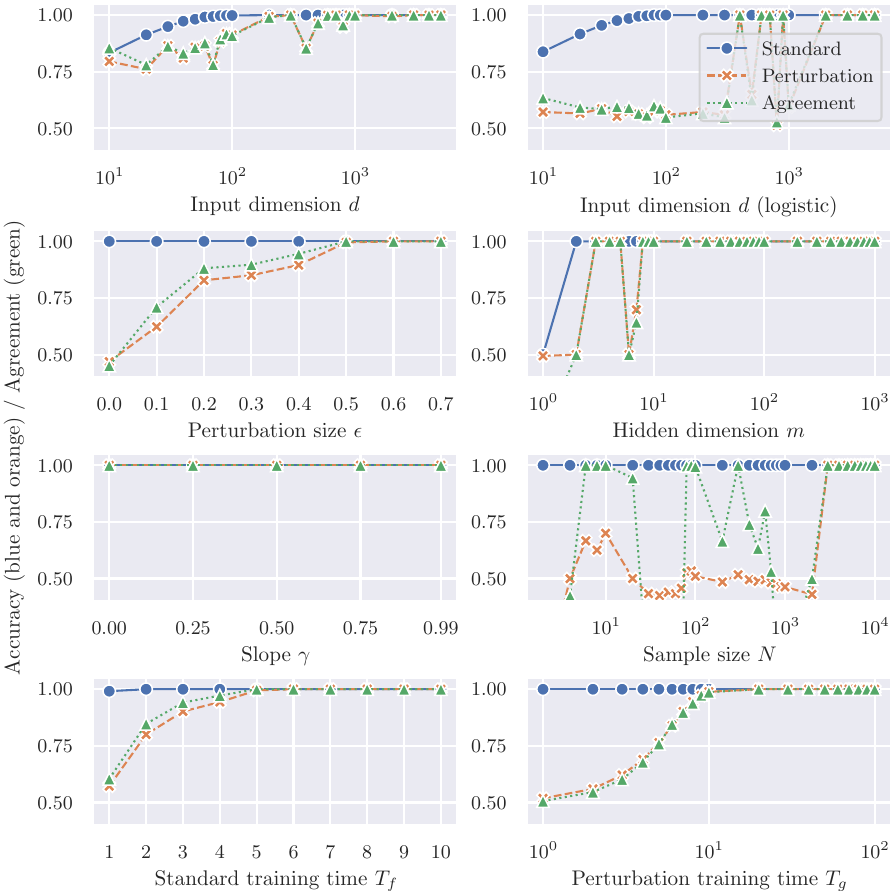}
\caption{Accuracy and agreement ratio on the mean-shifted Gaussian in Scenario~(b). The description is the same as \cref{fig:shifted-gauss-a}.}
\label{fig:shifted-gauss-b}
\end{figure}

\begin{figure}
\centering
\includegraphics[width=\textwidth]{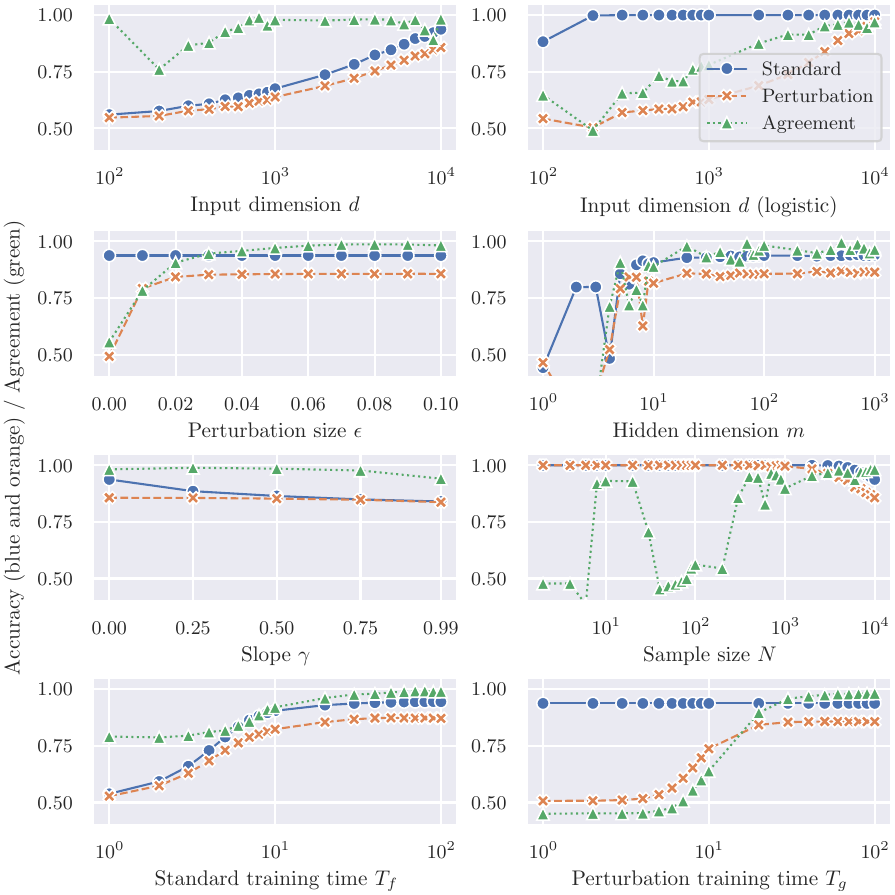}
\caption{Accuracy and agreement ratio on the zero-mean Gaussian in Scenario~(a). The description is the same as \cref{fig:shifted-gauss-a}.}
\label{fig:gauss-a}
\end{figure}

\begin{figure}
\centering
\includegraphics[width=\textwidth]{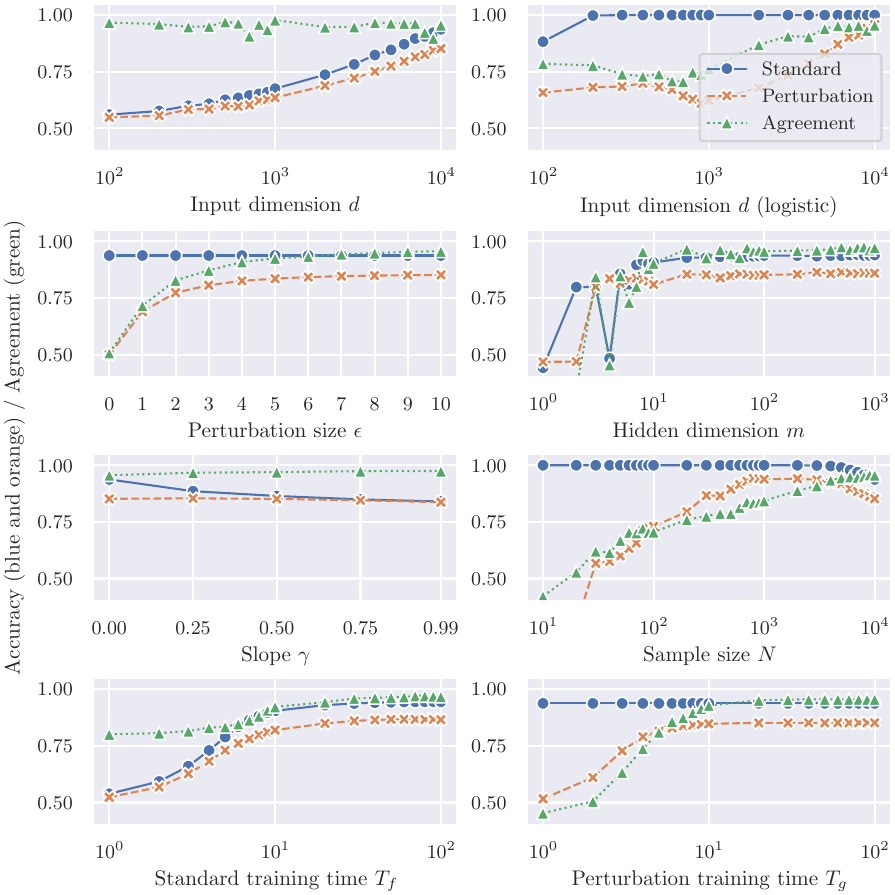}
\caption{Accuracy and agreement ratio on the zero-mean Gaussian in Scenario~(b). The description is the same as \cref{fig:shifted-gauss-a}.}
\label{fig:gauss-b}
\end{figure}

\begin{figure}
\centering
\includegraphics[width=\textwidth]{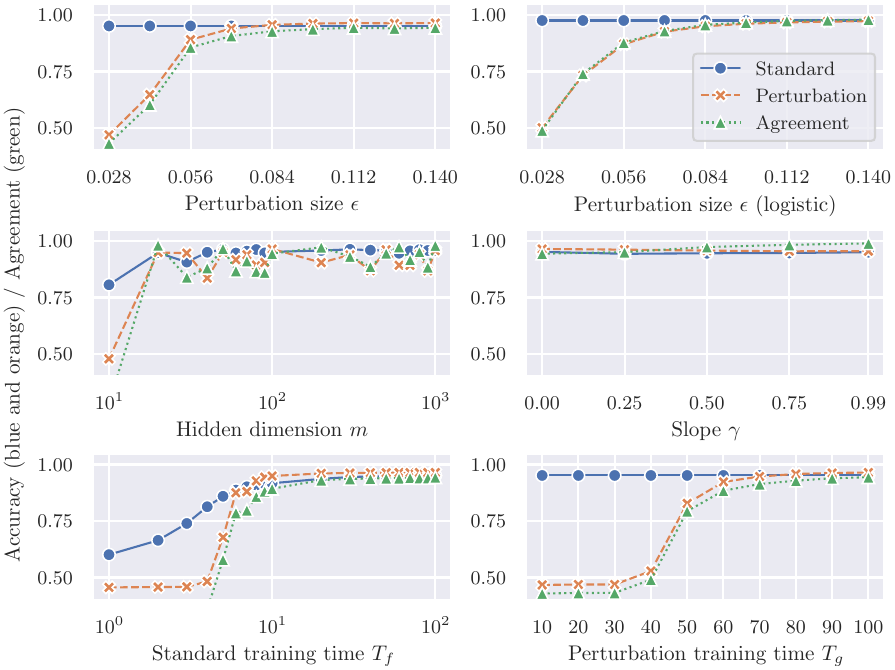}
\caption{Accuracy and agreement ratio on MNIST in Scenario~(a). The description is the same as \cref{fig:shifted-gauss-a}.}
\label{fig:MNIST-a}
\end{figure}

\begin{figure}
\centering
\includegraphics[width=\textwidth]{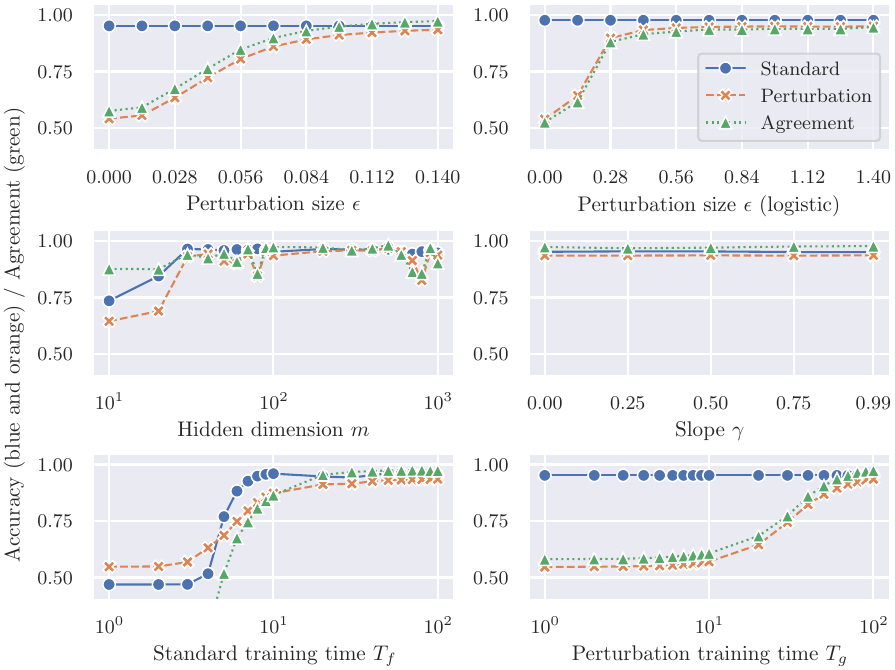}
\caption{Accuracy and agreement ratio on MNIST in Scenario~(b). The description is the same as \cref{fig:shifted-gauss-a}.}
\label{fig:MNIST-b}
\end{figure}

\begin{figure}
\centering
\includegraphics[width=\textwidth]{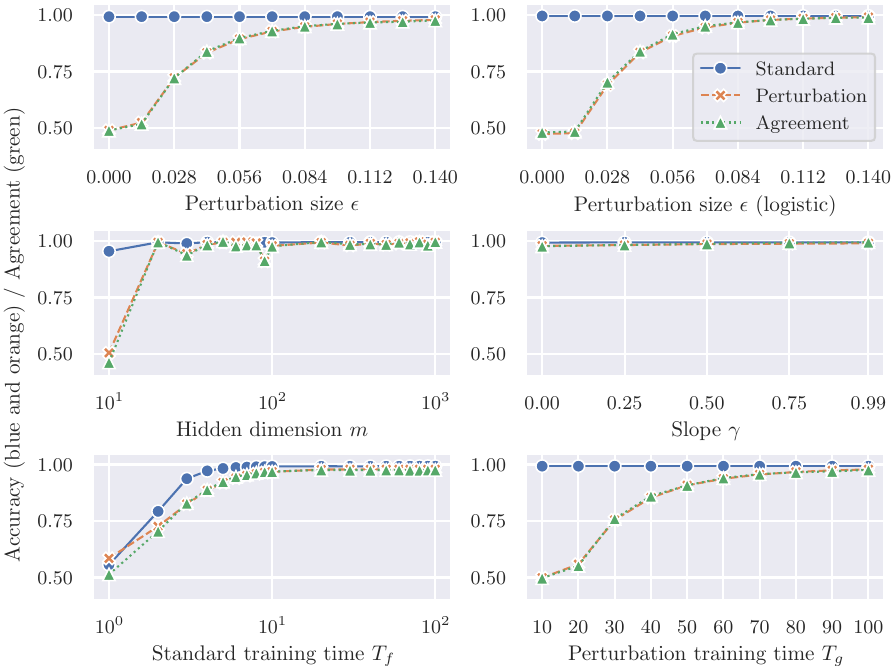}
\caption{Accuracy and agreement ratio on Fashion-MNIST in Scenario~(a). The description is the same as \cref{fig:shifted-gauss-a}.}
\label{fig:FMNIST-a}
\end{figure}

\begin{figure}
\centering
\includegraphics[width=\textwidth]{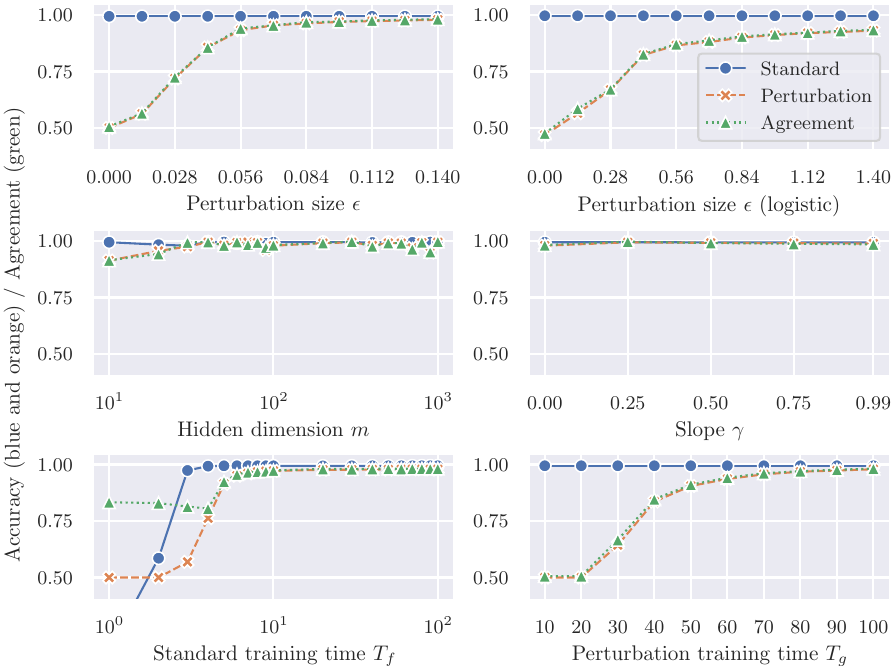}
\caption{Accuracy and agreement ratio on Fashion-MNIST in Scenario~(b). The description is the same as \cref{fig:shifted-gauss-a}.}
\label{fig:FMNIST-b}
\end{figure}

\begin{figure}
\centering
\includegraphics[width=\textwidth]{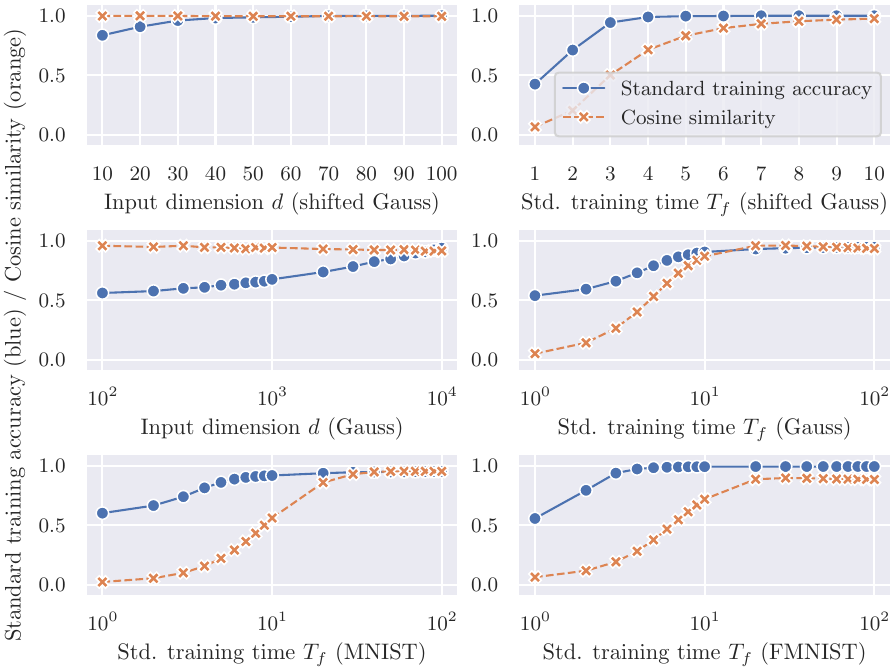}
\caption{Standard training accuracy and cosine similarity. The blue lines represent the accuracy of the classifier $f$ on $\calD := \{(\x_n,y_n)\}^N_{n=1}$, i.e., training accuracy. The orange lines represent the average cosine similarity between the experimentally calculated adversarial perturbations and the theoretically predicted ones.}
\label{fig:cossim}
\end{figure}

We used non-stochastic gradient descent~(i.e., each gradient calculation uses the entire dataset) with 0.9 momentum and the learning rate scheduler that multiplies a learning rate by 0.1 when a training loss has stopped improving during 10 epochs. For \crefrange{fig:shifted-gauss-a-mini}{fig:cossim}, we selected the best accuracy, agreement ratio, and cosine similarity from training with multiple initial learning rates.

In \crefrange{fig:shifted-gauss-a}{fig:FMNIST-b}, the blue lines represent the accuracy of the classifier from standard training on the training dataset $\{(\x_n, y_n)\}^N_{n=1}$. Namely, these mean the training accuracy of $f(\,\cdot\,;T_f)$. The orange lines represent the accuracy of the classifier from perturbation learning on the \textit{standard~(clean)} training dataset $\{(\x_n, y_n)\}^N_{n=1}$ rather than the perturbed dataset $\{(\xadv_n, \yadv_n)\}^N_{n=1}$. Namely, these mean the ratio that counterintuitive generalization occurs. Note that the orange lines stay around fifty percent~(chance accuracy) if adversarial perturbations are not included in $\{\xadv_n\}^N_{n=1}$~(cf. $\epsilon = 0$ in \crefrange{fig:shifted-gauss-a}{fig:FMNIST-b}). The green lines represent the agreement ratio between predictions $f(\,\cdot\,;T_f)$ and $g(\,\cdot\,;T_g)$ on a test dataset.

The cosine similarity in \cref{fig:cossim} is the average one between the experimentally calculated adversarial perturbation and the theoretically predicted one~(cf. \cref{eq:AP}) across all $n$. The blue lines are the same as those in \crefrange{fig:shifted-gauss-a}{fig:FMNIST-b}.

\begin{figure}
\centering
\includegraphics[width=\textwidth]{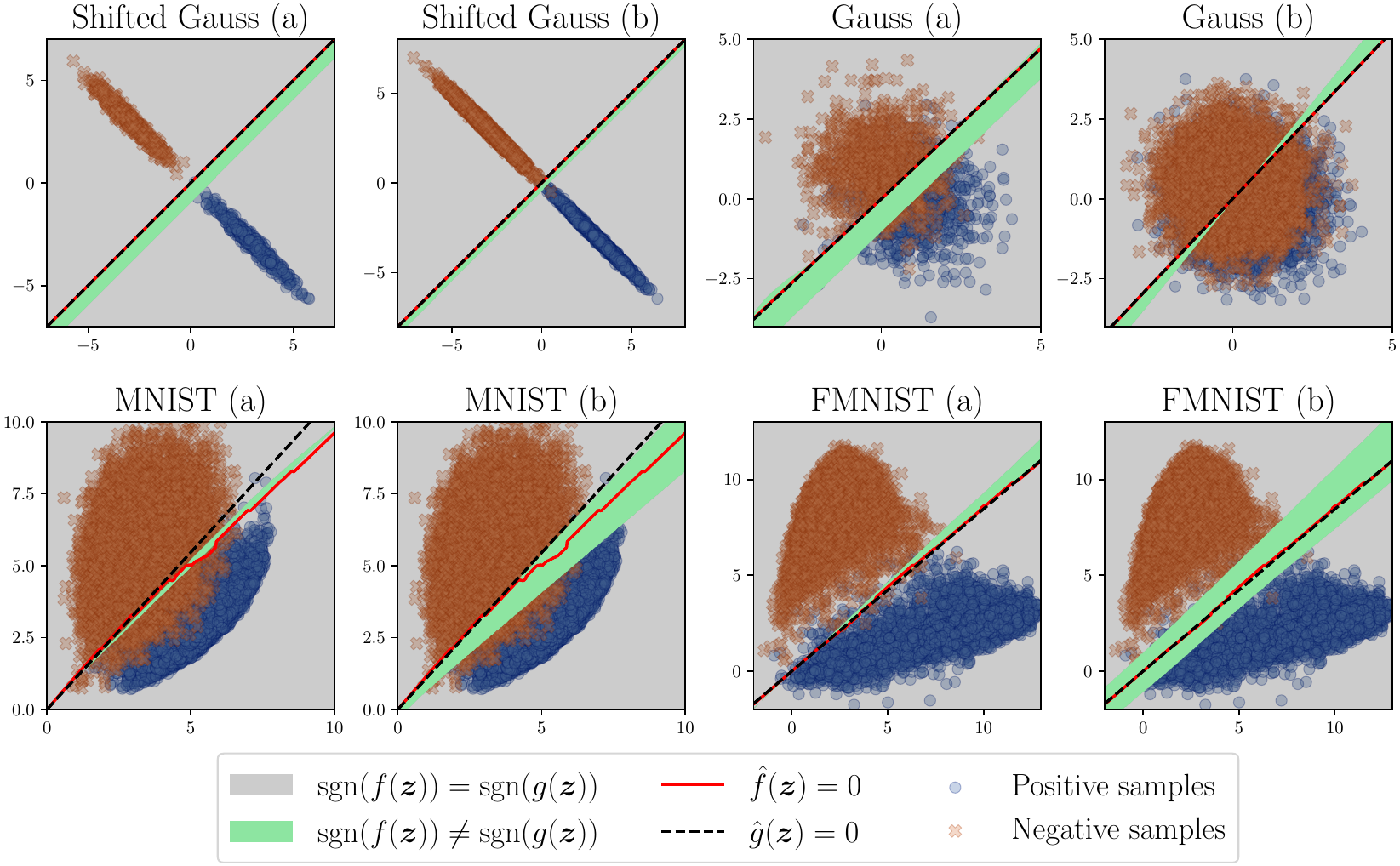}
\caption{Prediction matching between the classifiers from standard training and perturbation learning, $f$ and $g$. The two axes are the normalized average vectors of samples from the positive and negative classes, respectively. The blue circles and orange crosses correspond to the projections of positive and negative samples onto these axes. The gray and green areas indicate regions where two predictions are consistent and inconsistent, respectively. The red solid lines represent $\hatf(\z) = 0$. The black dashed lines represent $\hatg_a(\z) = 0$ in Scenario~(a) and $\hatg_b(\z) = 0$ in Scenario~(b).}
\label{fig:map}
\end{figure}

The two axes in \cref{fig:map} are the normalized average vectors of samples from the positive and negative classes, respectively. The blue circles and orange crosses correspond to the projections of positive and negative samples onto these axes. The gray and green areas indicate regions where two predictions are consistent and inconsistent, respectively. The red solid lines represent $\hatf(\z) = 0$. The black dashed lines represent $\hatg_a(\z) = 0$ in Scenario~(a) and $\hatg_b(\z) = 0$ in Scenario~(b).

\textbf{Mean-Shifted Gaussian and Scenario~(a).}
A common experimental setting for the mean-shifted Gaussian dataset and Scenario~(a) in \cref{fig:shifted-gauss-a,fig:cossim} is as follows: input dimension $d = 100$, hidden dimension $m = 100$, activation function slope $\gamma = 0$, number of training samples $N = 1,000$, loss function $\ell(s) = s$, training epochs in standard training $T_f = 1,000$ and in perturbation learning $T_g = 1,000$, perturbation size $\epsilon = \sqrt{d\times0.001^2} = 0.01$, and learning rates in standard training $\eta_f = 1$ or $0.1$ and in perturbation learning $\eta_g = 1$ or $0.1$. However, for the comparison of $T_f$~(i.e., the graph in the fourth row and the first column in \cref{fig:shifted-gauss-a}), we set $\eta_f$ only to 0.1. For the comparison of $d$~(i.e., the graph in the first row in \cref{fig:shifted-gauss-a}), we employed $\epsilon = \sqrt{d\times0.001^2}$. In \cref{fig:map}, we used $d = 100$, $m = 100$, $\gamma = 0$, $N = 1,000$, $\ell(s) = s$, $T_f = 100$, $T_g = 100$, $\epsilon = \sqrt{d\times0.001^2} = 0.01$, $\eta_f = 1$, and $\eta_g = 1$.

\textbf{Mean-Shifted Gaussian and Scenario~(b).}
A common experimental setting for the mean-shifted Gaussian dataset and Scenario~(b) in \cref{fig:shifted-gauss-b,fig:cossim} is as follows: $d = 5,000$, $m = 100$, $\gamma = 0$, $N = 10,000$, $\ell(s) = s$, $T_f = 1,000$, $T_g = 1,000$, $\epsilon = \sqrt{d\times0.01^2} = 0.7$, $\eta_f = 1$ or $0.1$, and $\eta_g = 1$ or $0.1$. However, for the comparison of $T_f$~(i.e., the graph in the fourth row and the first column in \cref{fig:shifted-gauss-b}), we set $\eta_f$ only to 0.01. In addition, for the comparison of $T_g$~(i.e., the graph in the fourth row and the second column in \cref{fig:shifted-gauss-b}), we set $\eta_g$ only to 0.01. Furthermore, we set $\eta_f = 10$, $5$, $1$, $0.1$, $0.01$ and $\eta_g = 10$, $5$, $1$, $0.1$, $0.01$ for the evaluation with the logistic loss~(i.e., the graph in the first row and the second column in \cref{fig:shifted-gauss-b}). For the comparison of $d$~(i.e., the graph in the first row in \cref{fig:shifted-gauss-b}), we employed $\epsilon = \sqrt{d\times0.01^2}$. In \cref{fig:map}, we used $d = 100$, $m = 100$, $\gamma = 0$, $N = 5,000$, $\ell(s) = s$, $T_f = 100$, $T_g = 100$, $\epsilon = \sqrt{d\times0.01^2} = 0.1$, $\eta_f = 1$, and $\eta_g = 1$.

\textbf{Zero-Mean Gaussian and Scenario~(a).}
A common experimental setting for the zero-mean Gaussian dataset and Scenario~(a) in \cref{fig:gauss-a,fig:cossim} is as follows: $d = 10,000$, $m = 100$, $\gamma = 0$, $N = 10,000$, $\ell(s) = s$, $T_f = 1,000$, $T_g = 1,000$, $\epsilon = \sqrt{d\times0.001^2} = 0.1$, $\eta_f = 1$ or $0.1$, and $\eta_g = 1$ or $0.1$. However, for the comparison of $T_f$~(i.e., the graph in the fourth row and the first column in \cref{fig:gauss-a}), we set $\eta_f$ only to 0.1. For the comparison of $d$~(i.e., the graph in the first row in \cref{fig:gauss-a}), we employed $\epsilon = \sqrt{d\times0.001^2}$. In \cref{fig:map}, we used $d = 1,000$, $m = 1,000$, $\gamma = 0$, $N = 2,000$, $\ell(s) = s$, $T_f = 1,000$, $T_g = 1,000$, $\epsilon = \sqrt{d\times0.001^2} = 0.031$, $\eta_f = 1$, and $\eta_g = 1$.

\textbf{Zero-Mean Gaussian and Scenario~(b).}
A common experimental setting for the zero-mean Gaussian dataset and Scenario~(b) in \cref{fig:gauss-b,fig:cossim} is as follows: $d = 10,000$, $m = 100$, $\gamma = 0$, $N = 10,000$, $\ell(s) = s$, $T_f = 1,000$, $T_g = 1,000$, $\epsilon = \sqrt{d\times0.1^2} = 10$, $\eta_f = 1$ or $0.1$, and $\eta_g = 1$ or $0.1$. However, for the comparison of $T_f$~(i.e., the graph in the fourth row and the first column in \cref{fig:gauss-b}), we set $\eta_f$ only to 0.1. In addition, for the comparison of $T_g$~(i.e., the graph in the fourth row and the second column in \cref{fig:gauss-b}), we set $\eta_g$ only to 0.1. For the comparison of $d$~(i.e., the graph in the first row in \cref{fig:gauss-b}), we employed $\epsilon = \sqrt{d\times0.1^2}$. In \cref{fig:map}, we used $d = 1,000$, $m = 1,000$, $\gamma = 0$, $N = 10,000$, $\ell(s) = s$, $T_f = 1,000$, $T_g = 1,000$, $\epsilon = \sqrt{d\times0.01^2} = 0.31$, $\eta_f = 0.1$, and $\eta_g = 0.1$.

\textbf{MNIST and Scenario~(a).}
A common experimental setting for MNIST and Scenario~(a) in \cref{fig:MNIST-a,fig:cossim} is as follows: $m = 100$, $\gamma = 0$, $\ell(s) = s$, $T_f = 100$, $T_g = 100$, $\epsilon = \sqrt{784\times0.01^2} / 2 = 0.14$, $\eta_f = 0.01$ or $0.001$, and $\eta_g = 0.01$ or $0.001$. In \cref{fig:map}, we used $m = 1,000$, $\gamma = 0$, $\ell(s) = s$, $T_f = 100$, $T_g = 100$, $\epsilon = \sqrt{784\times0.01^2} / 2 = 0.14$, $\eta_f = 0.01$, and $\eta_g = 0.01$.

\textbf{MNIST and Scenario~(b).}
A common experimental setting for MNIST and Scenario~(b) in \cref{fig:MNIST-b,fig:cossim} is as follows: $m = 100$, $\gamma = 0$, $\ell(s) = s$, $T_f = 100$, $T_g = 100$, $\epsilon = \sqrt{784\times0.01^2} / 2 = 0.14$, $\eta_f = 0.01$ or $0.001$, and $\eta_g = 0.01$ or $0.001$. In \cref{fig:map}, we used $m = 1,000$, $\gamma = 0$, $\ell(s) = s$, $T_f = 1,000$, $T_g = 1,000$, $\epsilon = \sqrt{784\times0.01^2} / 2 = 0.14$, $\eta_f = 0.01$, and $\eta_g = 0.01$.

\textbf{Fashion-MNIST and Scenario~(a).}
A common experimental setting for Fashion-MNIST and Scenario~(a) in \cref{fig:FMNIST-a,fig:cossim} is as follows: $m = 100$, $\gamma = 0$, $\ell(s) = s$, $T_f = 100$, $T_g = 100$, $\epsilon = \sqrt{784\times0.01^2} / 2 = 0.14$, $\eta_f = 0.01$ or $0.001$, and $\eta_g = 0.01$ or $0.001$. In \cref{fig:map}, we used $m = 1,000$, $\gamma = 0$, $\ell(s) = s$, $T_f = 100$, $T_g = 100$, $\epsilon = \sqrt{784\times0.01^2} / 2 = 0.14$, $\eta_f = 0.01$, and $\eta_g = 0.01$.

\textbf{Fashion-MNIST and Scenario~(b).}
A common experimental setting for Fashion-MNIST and Scenario~(b) in \cref{fig:FMNIST-b,fig:cossim} is as follows: $m = 100$, $\gamma = 0$, $\ell(s) = s$, $T_f = 100$, $T_g = 100$, $\epsilon = \sqrt{784\times0.01^2} / 2 = 0.14$, $\eta_f = 0.01$ or $0.001$, and $\eta_g = 0.01$ or $0.001$. In \cref{fig:map}, we used $m = 1,000$, $\gamma = 0$, $\ell(s) = s$, $T_f = 100$, $T_g = 100$, $\epsilon = \sqrt{784\times0.1^2} / 2 = 1.4$, $\eta_f = 0.001$, and $\eta_g = 0.001$.

\section{Lemmas}
In this section, we derive fundamental properties of random variables.

\begin{tcolorbox}
\begin{lemma}[Properties of Gaussian random variables]
\label{le:properties-Gaussian}
Let $\sigma^2 > 0$ be a positive constant. Let $X_1, \ldots, X_m \in \bbR$ be $m \in \bbN$ i.i.d.~Gaussian random variables that follow $\calN(0,\sigma^2)$.
\begin{enumerate}[label=(\alph*)]

  \item \label{item:properties-Gaussian-lkru}
  For $0 < \delta < 1$, with probability at least $1 - \delta$,
  \begin{align}
    \max_i |X_i| < \sqrt{ 2 \sigma^2 \ln(2m/\delta) }.
  \end{align}
  
  \item \label{item:properties-Gaussian-mmbv}
  Let $Y_1, \ldots, Y_m \in [\gamma^2, 1]$ be $m$ independent random variables with $0 \leq \gamma^2 < 1$. Suppose that $Y_1, \ldots, Y_m$ are independent of $X_1, \ldots, X_m$. For $0 < \delta < 1$, with probability at least $1 - \delta$,
  \begin{align}
    \abs{ \sum^m_i X^2_i Y_i - \sigma^2 \sum^m_i \bbE[Y_i] }
    < \max\qty(
        16 \sigma^2 \ln(\frac{2}{\delta}),
        \sqrt{ 128 m \sigma^4 \ln(\frac{2}{\delta}) }
    ).
  \end{align}

  \item \label{item:properties-Gaussian-pomn}
  For $\exp(-2(\eta\sqrt{m}+1)^2/m) < \delta < 1$, with probability at least $1 - \delta$, there are at most $\eta\sqrt{m} \in [m-1]$ instances such that
  \begin{align}
    |X_i| < \sqrt{
      -\frac{\pi\sigma^2}{2} \ln(
        1 - \qty( \frac{\eta\sqrt{m}+1}{m} - \sqrt{-\frac{\ln\delta}{2m}} )^2
      )
    }.
  \end{align}
  
\end{enumerate}
\end{lemma}
\end{tcolorbox}

\begin{proof}
Let $C > 0$ be a positive constant.

\cref*{item:properties-Gaussian-lkru}
From \cite{rigollet2023high},
\begin{align}
  \bbP[\max_i |X_i| \geq C] \leq 2 m \exp(-\frac{C^2}{2\sigma^2}).
\end{align}
Thus,
\begin{align}
  \bbP\qty[ \max_i |X_i| \geq \sqrt{2\sigma^2\ln(2m/\delta)} ] \leq \delta.
\end{align}

\cref*{item:properties-Gaussian-mmbv}
For any $i \in [m]$,
\begin{align}
  \bbE\qty[\exp(t (X_i^2 Y_i - \bbE[X_i^2] \bbE[Y_i]) )]
  =& \bbE\qty[ \sum^\infty_{n=0}
     \frac{ t^n (X_i^2 Y_i - \bbE[X_i^2] \bbE[Y_i])^n }{ n! } ] \\
  =& 1 + \sum^\infty_{n=2}
     \frac{ t^n \bbE[(X_i^2 Y_i - \bbE[X_i^2] \bbE[Y_i])^n] }{ n! } \\
  \leq &
  1 + \sum^\infty_{n=2}
  \frac{ t^n \bbE[(X_i^2 Y_i + \bbE[X_i^2] \bbE[Y_i])^n] }{ n! }.
\end{align}
By Jensen's inequality,
\begin{align}
  \sum^\infty_{n=2} \frac{ t^n \bbE[(X_i^2 Y_i + \bbE[X_i^2] \bbE[Y_i])^n] }{ n! }
  \leq & \sum^\infty_{n=2} \frac{
           2^{n-1} t^n (\bbE[X_i^{2n}] \bbE[Y_i^n] + \bbE[X_i^2]^n \bbE[Y_i]^n)
         }{ n! } \\
  \leq & \sum^\infty_{n=2} \frac{ 2^n t^n \bbE[X_i^{2n}] \bbE[Y_i^n] }{ n! }.
\end{align}
Since $\bbE[X_i^{2n}] = \sigma^{2n} (2n - 1)!! \leq 2^n \sigma^{2n} n!$ and $\bbE[Y_i^n] \leq 1$,
\begin{align}
  1 + \sum^\infty_{n=2} \frac{ 2^n t^n \bbE[X_i^{2n}] \bbE[Y_i^n] }{ n! }
  \leq 1 + \sum^\infty_{n=2} 4^n t^n \sigma^{2n}
  = 1 + \frac{ 16 t^2 \sigma^4 }{1 - 4 t \sigma^2}.
\end{align}
For $|t| \leq 1 / (8 \sigma^2)$,
\begin{align}
  1 + \frac{ 16 t^2 \sigma^4 }{1 - 4 t \sigma^2}
  \leq 1 + 32 t^2 \sigma^4
  \leq \exp(32 t^2 \sigma^4).
\end{align}
Thus, $X_i^2 Y_i$ follows $\SE(64\sigma^4, 8\sigma^2)$, where $SE(a, b)$ is a sub-exponential random variable with parameters $a, b > 0$. Note that a random variable $Z$ is called sub-exponential with parameters $a, b > 0$, $\SE(a, b)$, if its moment generating function satisfies
\begin{align}
  \forall |t| \leq \frac{1}{b}, \qquad
  \bbE[\exp(t(Z-\bbE[Z]))] \leq \exp(\frac{t^2a}{2}).
\end{align}
By \cite{john2023lecture}, $\sum^m_i X_i^2 Y_i$ follows $\SE(64m\sigma^4, 8\sigma^2)$. In addition, by \cite{john2023lecture}, $Z \sim \SE(a, b)$ satisfies
\begin{align}
  \bbP[|Z - \bbE[Z]| \geq C] \leq
  2 \exp( -\frac{1}{2} \min\qty(\frac{C^2}{a}, \frac{C}{b}) ).
\end{align}
Therefore, with probability at least $1 - \delta$,
\begin{align}
  |Z - \bbE[Z]|
  < \max\qty(2 b \ln(\frac{2}{\delta}), \sqrt{ 2 a \ln(\frac{2}{\delta}) }).
\end{align}

\cref*{item:properties-Gaussian-pomn}
Let $k \in [m-1]$ be a positive integer. Let $\mathrm{Bi}(m,p)$ be the Binomial distribution and $\mathrm{Be}(p)$ be the Bernoulli distribution with $p \in (0, (k+1)/m)$. By Hoeffding's inequality,
\begin{align}
  \sum^m_{i=k+1} \binom{m}{i} p^i (1-p)^{m-i}
  =& \bbP_{Y \sim \mathrm{Bi}(m,p)} [Y \geq k+1] \\
  =& \bbP_{Z_1,\ldots,Z_m \sim \mathrm{Be}(p)} \qty[\sum^m_{i=1} Z_i \geq k+1] \\
  \leq & \exp( -\frac{ 2 (k+1 - mp)^2 }{ m } ) \\
  =& \exp( -2m\qty( \frac{k+1}{m} - p )^2 ).
\end{align}
Now,
\begin{align}
  \notag
  & \bbP[\text{
    there are at least $k+1$ instances such that $|X_i| < C$
  }] \\
  =& \sum^m_{i=k+1} \binom{m}{i}
     \bbP[|X_i| < C]^i \bbP[|X_i| \geq C]^{m-i} \\
  \leq & \exp( -2m\qty( \frac{k+1}{m} - \bbP[|X_i| < C] )^2 ).
\end{align}
For $\delta > \exp( -\frac{2(k+1)^2}{m} )$ and $\bbP[|X_i| < C] \leq \frac{k+1}{m} - \sqrt{ - \frac{\ln \delta}{2m} }$,
\begin{align}
  \bbP[\text{
    there are at least $k+1$ instances
    such that $|X_i| < C$
  }] \leq \delta.
\end{align}
Since the areas of a square with sides of length $x$ and a quarter-circle with a radius of $2x/\sqrt{\pi}$ are the same, and because $s^2+t^2$ is always larger in the square than in the quarter-circle outside the common area, an upper bound of $\erf(x)$ can be computed as
\begin{align}
  \erf(x)^2
  =& \frac{4}{\pi} \int^x_0 \int^x_0 \exp(-(s^2+t^2)) \dd{s} \dd{t} \\
  \leq &
  \frac{4}{\pi} \int^{2x/\sqrt{\pi}}_0 \int^{\frac{\pi}{2}}_0
  r \exp(-r^2) \dd{\theta} \dd{r} \\
  =& 1 - \exp(-\frac{4}{\pi}x^2).
\end{align}
Thus,
\begin{align}
  \bbP[|X_i| < C] = \erf\qty( \frac{C}{\sqrt{2\sigma^2}} )
  \leq \sqrt{ 1 - \exp(-\frac{2C^2}{\pi\sigma^2}) }.
\end{align}
Therefore,
\begin{align}
  C \leq \sqrt{
    -\frac{\pi\sigma^2}{2} \ln(
      1 - \qty( \frac{k+1}{m} - \sqrt{-\frac{\ln\delta}{2m}} )^2
    )
  } \Longrightarrow
  \bbP[|X_i| < C] \leq \frac{k+1}{m} - \sqrt{ - \frac{\ln \delta}{2m} }.
\end{align}

\end{proof}

\begin{tcolorbox}
\begin{lemma}[Hoeffding's inequality]
\label{le:Hoeffding}
Let $s_1, \ldots, s_N \in \bbR$ be real numbers and $X_1, \ldots, X_N \in \{\pm1\}$ be i.i.d.~random variables. Suppose that $\bbE[X_n] = 0$ for every $n \in [N]$. Then, for $0 < \delta < 1$, with probability at least $1 - \delta$,
\begin{align}
  \abs{ \sum^N_n s_n X_n } < \sqrt{ 2 \ln(\frac{2}{\delta}) \sum^N_n s^2_n }.
\end{align}
\end{lemma}
\end{tcolorbox}

\begin{proof}
By Hoeffding's inequality, the claim is established.
\end{proof}

\begin{tcolorbox}
\begin{lemma}[Expectation of product of derivatives of activation functions, part 1]
\label{le:expected-derivatives-1}
Denote a symmetric positive definite matrix by
\begin{align}
  \bSigma :=
  \begin{bmatrix}
    a & b \\
    b & c
  \end{bmatrix},
\end{align}
where $a, c > 0$ and $ac - b^2 > 0$. Then,
\begin{align}
  \notag
  & \abs{ \int^\infty_{-\infty} \int^\infty_{-\infty}
  \frac{ \phi'(x_1) \phi'(x_2) }{ 2\pi \sqrt{|\bSigma|} }
  \exp(-\frac{1}{2} \x^\top \bSigma^{-1} \x) \dx - \frac{(1+\gamma)^2}{4} } \\
  \leq &
  \frac{(1 + \gamma)(1 - \gamma)}{4} \qty( 1 -
  \sqrt{ \frac{e}{2\pi} \frac{ac-b^2}{ac+b^2} } ).
\end{align}
\end{lemma}
\end{tcolorbox}

\begin{proof}
The inverse of $\bSigma$ is
\begin{align}
  \bSigma^{-1}
  =& \frac{1}{ac - b^2}
  \begin{bmatrix}
    c & -b \\
    -b & a
  \end{bmatrix}
  =: \frac{\tilde{\bSigma}^{-1}}{ac - b^2}.
\end{align}
Using $\y := \x / \sqrt{ac - b^2}$, the quadratic form can be represented as
\begin{align}
  \x^\top \bSigma^{-1} \x
  = \frac{ \x^\top \tilde{\bSigma}^{-1} \x }{ac - b^2}
  = \y^\top \tilde{\bSigma}^{-1} \y.
\end{align}
Because $|\partial\x/\partial\y| = ac - b^2$,
\begin{align}
  \notag
  & \int^\infty_{-\infty} \int^\infty_{-\infty}
  \frac{ \phi'(x_1) \phi'(x_2) }{ 2\pi \sqrt{|\bSigma|} }
  \exp(-\frac{1}{2} \x^\top \bSigma^{-1} \x) \dd{\x} \\
  =& \frac{\sqrt{ac-b^2}}{2\pi}
     \int^\infty_{-\infty} \int^\infty_{-\infty} \phi'(y_1) \phi'(y_2)
     \exp(-\frac{1}{2} \y^\top \tilde{\bSigma}^{-1} \y) \dd{\y}.
\end{align}
With $z := y_1 - by_2/c$,
\begin{align}
  \y^\top \tilde{\Sigma}^{-1} \y
  = c\qty(y_1 - \frac{by_2}{c})^2 + \qty(a - \frac{b^2}{c}) y^2_2
  = cz^2 + \qty(a - \frac{b^2}{c}) y^2_2.
\end{align}
Now,
\begin{align}
  \notag
  & \int^\infty_{-\infty} \int^\infty_{-\infty} \phi'(y_1) \phi'(y_2)
  \exp(-\frac{1}{2} \y^\top \tilde{\bSigma}^{-1} \y) \dd\y \\
  =& \int^\infty_{-\infty} \phi'(y_2) \qty( \int^\infty_{-\infty}
     \phi'(z + by_2/c) \exp(-\frac{c}{2}z^2) \dz )
     \exp( -\frac{ac-b^2}{2c} y^2_2 ) \dy_2.
\end{align}
The integral along $z$ can be computed as
\begin{align}
  \notag
  & \int^\infty_{-\infty} \phi'(z + by_2/c) \exp(-\frac{c}{2}z^2) \dz \\
  =& \int^\infty_{-by_2/c} \exp(-\frac{c}{2}z^2) \dz
     + \gamma \int^{-by_2/c}_{-\infty} \exp(-\frac{c}{2}z^2) \dz \\
  \notag
  =& \int^\infty_0 \exp(-\frac{c}{2}z^2) \dz
     - \int^{-by_2/c}_0 \exp(-\frac{c}{2}z^2) \dz \\
     &+ \gamma \int^0_{-\infty} \exp(-\frac{c}{2}z^2) \dz
     - \gamma \int^0_{-by_2/c} \exp(-\frac{c}{2}z^2) \dz \\
  =& (1 + \gamma) \int^\infty_0 \exp(-\frac{c}{2}z^2) \dz
     - (1 - \gamma) \int^{-by_2/c}_0 \exp(-\frac{c}{2}z^2) \dz \\
  =& (1 + \gamma) \sqrt{\frac{\pi}{2c}}
     + (1 - \gamma) \int^{by_2/c}_0 \exp(-\frac{c}{2}z^2) \dz.
\end{align}
Using the above equation,
\begin{align}
  \notag
  & \int^\infty_{-\infty} \int^\infty_{-\infty} \phi'(y_1) \phi'(y_2)
  \exp(-\frac{1}{2} \y^\top \tilde{\bSigma}^{-1} \y) \dy \\
  \notag
  =& (1 + \gamma) \sqrt{\frac{\pi}{2c}} \int^\infty_{-\infty} \phi'(y_2)
     \exp( -\frac{ac-b^2}{2c}y^2_2 ) \dy_2 \\
     & + (1 - \gamma) \int^\infty_{-\infty}
       \phi'(y_2) \int^{by_2/c}_0 \exp(-\frac{c}{2}z^2) \dz
       \exp( -\frac{ac-b^2}{2c}y^2_2 ) \dy_2 \\
  =& \frac{ \pi (1 + \gamma)^2 }{ 2 \sqrt{ac - b^2} }
     + (1 + \gamma) (1 - \gamma) \int^\infty_0
       \int^{by_2/c}_0 \exp(-\frac{c}{2}z^2) \dz
       \exp( -\frac{ac-b^2}{2c}y^2_2 ) \dy_2.
\end{align}
Thus,
\begin{align}
  \notag
  & \int^\infty_{-\infty} \int^\infty_{-\infty}
  \frac{ \phi'(x_1) \phi'(x_2) }{ 2\pi \sqrt{|\bSigma|} }
  \exp(-\frac{1}{2} \x^\top \bSigma^{-1} \x) \dx \\
  \notag
  =& \frac{(1+\gamma)^2}{4} \\
     &+ \frac{ (1 + \gamma) (1 - \gamma) \sqrt{ac-b^2} }{2\pi}
        \int^\infty_0 \int^{by_2/c}_0 \exp(-\frac{c}{2}z^2) \dz
        \exp( -\frac{ac-b^2}{2c}y^2_2 ) \dy_2.
\end{align}
Using $t := \sgn(b) z \sqrt{c/2}$,
\begin{align}
  \int^{by_2/c}_0 \exp(-\frac{c}{2}z^2) \dz
  =& \sgn(b) \sqrt{\frac{2}{c}}
     \int^{\sqrt{\frac{b^2}{2c}}y_2}_0 \exp(-t^2) \dt \\
  =& \sgn(b) \sqrt{\frac{\pi}{2c}}
     \qty( 1 - \erfc\qty(\sqrt{\frac{b^2}{2c}}y_2) ).
\end{align}
From \cite{chang2011chernoff}, with $\alpha = \sqrt{e/(2\pi)}$,
\begin{align}
  \alpha \exp(-\frac{b^2}{c}y^2_2)
  \leq & \erfc\qty(\sqrt{\frac{b^2}{2c}}y_2), \\
  \abs{ \int^{by_2/c}_0 \exp(-\frac{c}{2}z^2) \dz }
  \leq & \sqrt{\frac{\pi}{2c}} \qty(1 - \alpha\exp(-\frac{b^2}{c}y^2_2)).
\end{align}
Thus,
\begin{align}
  \notag
  & \abs{ \int^\infty_0 \int^{by_2/c}_0 \exp(-\frac{c}{2}z^2) \dz
  \exp( -\frac{ac-b^2}{2c}y^2_2 ) \dy_2 } \\
  \leq &
  \sqrt{\frac{\pi}{2c}} \qty( \int^\infty_0 \exp( -\frac{ac-b^2}{2c}y^2_2 ) \dy_2
  - \alpha \int^\infty_0 \exp( -\frac{ac+b^2}{2c}y^2_2 ) \dy_2) \\
  =& \sqrt{\frac{\pi}{2c}} \qty( \frac{1}{2} \sqrt{\frac{2\pi c}{ac-b^2}}
     - \frac{\alpha}{2} \sqrt{\frac{2\pi c}{ac+b^2}}) \\
  =& \frac{\pi}{2} \qty( \frac{1}{\sqrt{ac-b^2}}
     - \frac{\alpha}{\sqrt{ac+b^2}} ).
\end{align}
Finally,
\begin{align}
  \notag
  & \abs{ \int^\infty_{-\infty} \int^\infty_{-\infty}
  \frac{ \phi'(x_1) \phi'(x_2) }{ 2\pi \sqrt{|\bSigma|} }
  \exp(-\frac{1}{2} \x^\top \bSigma^{-1} \x) \dx - \frac{(1+\gamma)^2}{4} } \\
  \leq & \frac{ (1 + \gamma) (1 - \gamma) \sqrt{ac-b^2} }{2\pi}
         \qty( \frac{\pi}{2} \qty( \frac{1}{\sqrt{ac-b^2}}
         - \frac{\alpha}{\sqrt{ac+b^2}} ) ) \\
  =& \frac{(1 + \gamma)(1 - \gamma)}{4} \qty( 1 - \alpha 
     \sqrt{ \frac{ac-b^2}{ac+b^2} } ).
\end{align}
\end{proof}

\begin{tcolorbox}
\begin{lemma}[Expectation of product of derivatives of activation functions, part 2]
\label{le:expected-derivatives-2}
For any $\z_1 \ne \z_2 \in \bbR^d$,
\begin{align}
  \abs{ \Phi(\z_1, \z_2) - \frac{(1+\gamma)^2}{4} }
  \leq \frac{(1+\gamma)(1-\gamma)}{4} \lambda(\z_1, \z_2).
\end{align}
\end{lemma}
\end{tcolorbox}

\begin{proof}
By the reproductive property of Gaussian distributions, $\ip{\v}{\z_1} + a$ follows $\calN(0,\|\z_1\|^2/d+1)$. Since any linear combination of $\ip{\v}{\z_1} + a$ and $\ip{\v}{\z_2} + a$ has a univariate Gaussian distribution, $\ip{\v}{\z_1} + a$ and $\ip{\v}{\z_2} + a$ follow a multivariate Gaussian distribution. The covariance matrix $\Sigma$ can be computed as
\begin{align}
  \Sigma =
  \begin{bmatrix}
    \|\z_1\|^2/d + 1 & \ip{\z_1}{\z_2}/d + 1 \\
    \ip{\z_1}{\z_2}/d + 1 & \|\z_2\|^2/d + 1
  \end{bmatrix}.
\end{align}
Thus, by \cref{le:expected-derivatives-1}, the claim is established.
\end{proof}

\section{Main Proof}
\label{sec:main-proof}
For notational simplicity, we use the following abbreviation for $i \in [m]$ and $n \in [N]$:
\begin{align}
  h_{f,i,t}(\z) :=& \ip{\v_i(t)}{\z} + a_i(t), &
  h_{g,i,t}(\z) :=& \ip{\w_i(t)}{\z} + b_i(t), \\
  \psi_{f,i,t}(\z) :=& \phi(h_{f,i,t}(\z)), &
  \psi_{g,i,t}(\z) :=& \phi(h_{g,i,t}(\z)), \\
  \psi'_{f,i,t}(\z) :=& \phi'(h_{f,i,t}(\z)), &
  \psi'_{g,i,t}(\z) :=& \phi'(h_{g,i,t}(\z)), \\
  \ell_{f,n,t} :=& \ell(-y_nf(\x_n;\bthetaVa(t))), &
  \ell_{g,n,t} :=& \ell(-\yadv_ng(\xadv_n;\bthetaWb(t))), \\
  \ell'_{f,n,t} :=& \ell'(-y_nf(\x_n;\bthetaVa(t))), &
  \ell'_{g,n,t} :=& \ell'(-\yadv_ng(\xadv_n;\bthetaWb(t))), \\
  \barell'_f(t) :=& \frac{1}{N} \sum^N_n \ell'_{f,n,t}, &
  \barell'_g(t) :=& \frac{1}{N} \sum^N_n \ell'_{g,n,t}.
\end{align}
Moreover, denote the subset of $[m]$ consisting of the smallest $\eta\sqrt{m} \in [m-1]$ elements in terms of $|h_{f,i,0}(\z)|$ by $\calS_f(\z)$. Similarly, we define $\calS_g(\z)$ based on $|h_{g,i,0}(\z)|$. The function $\kappa_{f,i}(\z)$ returns one if $i \in \calS_f(\z)$ and zero otherwise. Similarly, we define $\kappa_{g,i}(\z)$. For $\exp(-2(\eta\sqrt{m}+1)^2/m) < \delta < 1$, let
\begin{align}
  \Cthr(\z,\delta) := \sqrt{
    - \pi \qty(\frac{\|\z\|^2}{d}+1) \ln(
    1 - \qty( \frac{\eta\sqrt{m}+1}{m}
    - \sqrt{-\frac{\ln\delta}{2m}} )^2 )
  } = \tildeTheta(1).
\end{align}

\subsection{Assumptions on Properties of Neural Networks and Their Justifications}
For reference, we state the following assumption on the network $f$:

\begin{tcolorbox}
\begin{assumption}[Properties of neural network $f$]
\label{asm:properties-network-f}
Let $\z \in \bbR^d$ be a real-valued vector and $\exp(-2(\eta\sqrt{m}+1)^2/m) < \delta < 1$ be a positive value.
\begin{enumerate}[label=(\alph*)]

  \item \label{item:properties-network-v}
  For any $i \in [m]$ and $j \in [d]$, $|v_{i,j}(0)| < \sqrt{ (2/d) \ln(2dm/\delta) }$.
  
  \item \label{item:properties-network-alpha}
  For any $i \in [m]$, $|\alpha_i| < \sqrt{ (2/m) \ln(2m/\delta)}$.
  
  \item \label{item:properties-network-vz}
  For any $i \in [m]$, $|\ip{\v_i(0)}{\z}| < \sqrt{ (2/d) \ln(2m/\delta) } \|\z\|$.
  
  \item \label{item:properties-network-h}
  For any $i \in [m]$, $|h_{f,i,0}(\z)| < \sqrt{2 (\|\z\|^2/d + 1) \ln (2m/\delta)}$.
  
  \item \label{item:properties-network-f}
  $|f(\z;0)| < \sqrt{ 2 (\|\z\|^2/d + 1) \ln(2/\delta) }$.
  
  \item \label{item:properties-network-sum-v}
  For any $j \in [d]$,
  \begin{align}
    \abs{ \sum^m_i \alpha_i \psi'_{f,i,0}(\x_n) v_{i,j}(0) }
    < \sqrt{ \frac{2}{m} \ln(\frac{2}{\delta})
             \sum^m_i \psi'_{f,i,0}(\x_n)^2 v_{i,j}(0)^2 }.
  \end{align}
  
  \item \label{item:properties-network-sum-vz}
  \begin{align}
    \abs{ \sum^m_i \alpha_i \psi'_{f,i,0}(\x_n) \ip{\v_i(0)}{\z} }
    <& \sqrt{ \frac{2}{m} \ln(\frac{2}{\delta})
              \sum^m_i \psi'_{f,i,0}(\x_n)^2 \ip{\v_i(0)}{\z}^2 }.
  \end{align}
  
  \item \label{item:properties-network-NTK}
  \begin{align}
    \abs{
      \sum^m_i \alpha^2_i \psi'_{f,i,0}(\x_n) \psi'_{f,i,0}(\z) - \Phi(\x_n, \z)
    } < \frac{16}{\sqrt{m}} \ln(\frac{2}{\delta}).
  \end{align}
  
  \item \label{item:properties-network-number}
  There are at most $\eta\sqrt{m}$ instances such that $|h_{f,i,0}(\z)| < \Cthr(\z,\delta) / \sqrt{2}$.
  
\end{enumerate}
\end{assumption}
\end{tcolorbox}

\cref{asm:properties-network-f} is justified as follows:

\begin{tcolorbox}
\begin{lemma}[Justification of \cref{asm:properties-network-f}]
\label{le:properties-network-f}
~
\begin{enumerate}[label=(\alph*)]
  \item With probability at least $1 - \delta$, \cref{asm:properties-network-f}\cref{item:properties-network-v} holds.
  \item With probability at least $1 - \delta$, \cref{asm:properties-network-f}\cref{item:properties-network-alpha} holds.
  \item With probability at least $1 - \delta$, \cref{asm:properties-network-f}\cref{item:properties-network-vz} holds.
  \item With probability at least $1 - \delta$, \cref{asm:properties-network-f}\cref{item:properties-network-h} holds.
  \item With probability at least $1 - \delta$, \cref{asm:properties-network-f}\cref{item:properties-network-f} holds.
  \item With probability at least $1 - \delta$,
  \cref{asm:properties-network-f}\cref{item:properties-network-sum-v} holds.
  \item With probability at least $1 - \delta$,
  \cref{asm:properties-network-f}\cref{item:properties-network-sum-vz} holds.
  \item With probability at least $1 - \delta$,
  \cref{asm:properties-network-f}\cref{item:properties-network-NTK} holds.
  \item With probability at least $1 - \delta$,
  \cref{asm:properties-network-f}\cref{item:properties-network-number} holds.
  \item \label{item:properties-network-all}
  With probability at least $1 - 9 \delta$,
  \cref{asm:properties-network-f} holds.
\end{enumerate}
\end{lemma}
\end{tcolorbox}

\begin{proof}
~

\cref*{item:properties-network-v,item:properties-network-alpha,item:properties-network-NTK}
See \cref{le:properties-Gaussian}.

\cref*{item:properties-network-vz,item:properties-network-h,item:properties-network-f,item:properties-network-sum-v,item:properties-network-sum-vz,item:properties-network-number}
By the reproductive property of Gaussian random variables, $\ip{\v_i(0)}{\z}$, $h_{f,i,0}(\z)$, $f(\z;0)$, $\sum^m_i \alpha_i \psi'_{f,i,0}(\x_n) v_{i,j}(0)$, and $\sum^m_i \alpha_i \psi'_{f,i,0}(\x_n) \ip{\v_i(0)}{\z}$ follow the Gaussian $\calN(0, \|\z\|^2/d)$, $\calN(0, \|\z\|^2/d+1)$, $\calN(0, \|\z\|^2/d+1)$, $\calN(0, (1/m) \sum^m_i \psi'_{f,i,0}(\x_n)^2 v_{i,j}(0)^2)$, and $\calN(0, (1/m) \sum^m_i \psi'_{f,i,0}(\x_n)^2 \ip{\v_i(0)}{\z}^2)$, respectively. By \cref{le:properties-Gaussian}, the claim is established.

\cref*{item:properties-network-all}
By Bonferroni's inequality, the claim is established.

\end{proof}

Similarly, we consider the following assumption on the network $g$ and its justification:

\begin{tcolorbox}
\begin{assumption}[Properties of neural network $g$]
\label{asm:properties-network-g}
Let $\z \in \bbR^d$ be a real-valued vector and $\exp(-2(\eta\sqrt{m}+1)^2/m) < \delta < 1$ be a positive value.
\begin{enumerate}[label=(\alph*)]

  \item For any $i \in [m]$ and $j \in [d]$, $|w_{i,j}(0)| < \sqrt{(2/d) \ln(2dm/\delta)}$.
  
  \item For any $i \in [m]$, $|\beta_i| < \sqrt{(2/m) \ln(2m/\delta)}$.
  
  \item For any $i \in [m]$, $|\ip{\w_i(0)}{\z}| < \sqrt{(2/d) \ln(2m/\delta)} \|\z\|$.
  
  \item For any $i \in [m]$, $|h_{g,i,0}(\z)| < \sqrt{2 (\|\z\|^2/d + 1) \ln (2m/\delta)}$.
  
  \item $|g(\z;0)| < \sqrt{ 2 (\|\z\|^2/d + 1) \ln(2/\delta) }$.
  
  \item For any $j \in [d]$,
  \begin{align}
    \abs{ \sum^m_i \beta_i \psi'_{g,i,0}(\xadv_n) w_{i,j}(0) }
    < \sqrt{ \frac{2}{m} \ln(\frac{2}{\delta})
             \sum^m_i \psi'_{g,i,0}(\xadv_n)^2 w_{i,j}(0)^2 }.
  \end{align}
  
  \item
  \begin{align}
    \abs{ \sum^m_i \beta_i \psi'_{g,i,0}(\xadv_n) \ip{\w_i(0)}{\z} }
    <& \sqrt{ \frac{2}{m} \ln(\frac{2}{\delta})
              \sum^m_i \psi'_{g,i,0}(\xadv_n)^2 \ip{\w_i(0)}{\z}^2 }.
  \end{align}
  
  \item
  \begin{align}
    \abs{
      \sum^m_i \beta^2_i \psi'_{g,i,0}(\xadv_n) \psi'_{g,i,0}(\z)
      - \Phi(\xadv_n, \z)
    } < \frac{16}{\sqrt{m}} \ln(\frac{2}{\delta}).
  \end{align}
  
  \item There are at most $\eta\sqrt{m}$ instances such that $|h_{g,i,0}(\z)| < \Cthr(\z,\delta) / \sqrt{2}$.
  
\end{enumerate}
\end{assumption}
\end{tcolorbox}

\begin{tcolorbox}
\begin{lemma}[Justification of \cref{asm:properties-network-g}]
\label{le:properties-network-g}
With probability at least $1 - 9 \delta$, \cref{asm:properties-network-g} holds.
\end{lemma}
\end{tcolorbox}

\subsection{Wide Width Assumptions}
Then, we consider the condition of network width for lazy training.

\begin{tcolorbox}
\begin{assumption}[Wide width for neural network $f$]
\label{asm:width-ap-f}
Let $\z \in \bbR^d$ be a real-valued vector and $\exp(-2(\eta\sqrt{m}+1)^2/m) < \delta < 1$ be a positive value. Network width $m$ satisfies the following inequalities:
\begin{align}
  \label[ineq]{ineq:width}
  m >& \frac
  { 4 \ln(2m/\delta) ( \sum^N_n (|\ip{\x_n}{\z}| + 1)
    \int^{T_f}_0 \ell'_{f,n,t} \dt )^2 }
  { N^2 \Cthr(\z,\delta)^2 }
  = \tildecalO\qty( d^2 \qty(\int^{T_f}_0 \barell'_{f,n,t} \dt)^2 ), \\
  \label[ineq]{ineq:width-extend}
  m >& \frac
  { 4 \sum^N_{n,k} |\ip{\x_n}{\x_k}|
    \int^{T_f}_0 \ell'_{f,n,t} \dt \int^{T_f}_0 \ell'_{f,k,t} \dt }
  { N^2 }
  = \tildecalO\qty( d^2 \qty(\int^{T_f}_0 \barell'_{f,n,t} \dt)^2 ).
\end{align}
\end{assumption}
\end{tcolorbox}

Note that only \cref{ineq:width} is required to satisfy lazy training. We impose \cref{ineq:width-extend} to simplify some rearrangements of equations. This assumption restricts the transitions of the derivatives of hidden outputs during training.

\begin{tcolorbox}
\begin{lemma}[Lazy training in network $f$]
\label{le:kernel-regime-f}
If \cref{asm:properties-network-f,asm:width-ap-f} hold, then $\psi'_{f,i,t}(\z) = \psi'_{f,i,0}(\z)$ for any $i \in [m] \setminus \calS_f(\z)$ and $0 \leq t \leq T_f$.
\end{lemma}
\end{tcolorbox}

\begin{proof}
By \cref{asm:properties-network-f}, the time evolution of $h_{f,i,t}(\z)$ from $t = 0$ to $t = T_f$ can be computed as
\begin{align}
  | \Delta h_{f,i,T_f}(\z) |
  :=& \abs{ \ip*{ - \int^{T_f}_0 \bnabla_{\v_i}
      \calL(\btheta_{\V,\a}(t);\calD) \dt }{ \z }
      - \int^{T_f}_0 \nabla_{a_i} \calL(\btheta_{\V,\a}(t);\calD) \dt } \\
  =& \abs{ \int^{T_f}_0 \frac{\alpha_i}{N}
     \sum^N_n y_n \ell'_{f,n,t}
     \psi'_{f,i,t}(\x_n) (\ip{\x_n}{\z} + 1) \dt } \\
  \leq &
  \frac{|\alpha_i|}{N} \sum^N_n |\ip{\x_n}{\z} + 1|
  \int^{T_f}_0 \ell'_{f,n,t} \dt \\
  \label[ineq]{ineq:tmp-nbvr}
  <& \frac{1}{N}
     \sqrt{ \frac{2}{m} \ln(\frac{2m}{\delta}) }
     \sum^N_n |\ip{\x_n}{\z} + 1|
     \int^{T_f}_0 \ell'_{f,n,t} \dt.
\end{align}
By \cref{asm:properties-network-f}, if the right term of \cref{ineq:tmp-nbvr} is smaller than $\Cthr(\z,\delta) / \sqrt{2}$, then the largest $m - \eta \sqrt{m}$ instances in terms of $|h_{f,i,0}(\z)|$ satisfy $\sgn(h_{f,i,t}(\z)) = \sgn(h_{f,i,0}(\z))$ for $0 \leq t \leq T_f$. This condition can be represented as \cref{ineq:width}.
\end{proof}

The same discussion can be applied to the network $g$.

\begin{tcolorbox}
\begin{assumption}[Wide width for neural network $g$]
\label{asm:width-ap-g}
Let $\z \in \bbR^d$ be a real-valued vector and $\exp(-2(\eta\sqrt{m}+1)^2/m) < \delta < 1$ be a positive value. Network width $m$ satisfies the following inequalities:
\begin{align}
  m >& \frac
  { 4 \ln(2m/\delta) ( \sum^N_n (|\ip{\xadv_n}{\z}| + 1)
    \int^{T_g}_0 \ell'_{g,n,t} \dt )^2 }
  { N^2 \Cthr(\z,\delta)^2 }
  = \tildecalO\qty( d^2 \qty(\int^{T_g}_0 \barell'_{g,n,t} \dt)^2 ), \\
  m >& \frac
  { 4 \sum^N_{n,k} |\ip{\xadv_n}{\xadv_k}|
    \int^{T_g}_0 \ell'_{g,n,t} \dt
    \int^{T_g}_0 \ell'_{g,k,t} \dt }
  { N^2 }
  = \tildecalO\qty( d^2 \qty(\int^{T_g}_0 \barell'_{g,n,t} \dt)^2 ).
\end{align}
\end{assumption}
\end{tcolorbox}

\begin{tcolorbox}
\begin{lemma}[Lazy training in network $g$]
\label{le:kernel-regime-g}
If \cref{asm:properties-network-g,asm:width-ap-g} hold, then $\psi'_{g,i,t}(\z) = \psi'_{g,i,0}(\z)$ for any $i \in [m] \setminus \calS_g(\z)$ and $0 \leq t \leq T_g$.
\end{lemma}
\end{tcolorbox}

We can integrate \cref{asm:properties-network-f,asm:properties-network-g} into \cref{asm:width}.

\subsection{Main Part}

\begin{tcolorbox}
\begin{lemma}[Representation of network $f$]
\label{le:representation-network-f}
If \cref{asm:properties-network-f,asm:width-ap-f} hold, then the network output at the training time $T_f$ can be represented as
\begin{align}
  \notag
  f(\z;T_f)
  =& f(\z;0)
  + \sum^m_i \alpha_i \kappa_{f,i}(\z)
    (\psi'_{f,i,T_f}(\z) - \psi'_{f,i,0}(\z)) h_{f,i,0}(\z) \\
  \notag
  &+ \frac{1}{N} \sum^N_n y_n (\ip{\x_n}{\z} + 1)
     \int^{T_f}_0 \ell'_{f,n,t} \dt \qty(
       \sum^m_i \alpha^2_i \psi'_{f,i,0}(\x_n) \psi'_{f,i,0}(\z)
       - \Phi(\x_n, \z)
     ) \\
  \notag
  &+ \frac{1}{N} \sum^N_n y_n \Phi(\x_n, \z) (\ip{\x_n}{\z} + 1)
     \int^{T_f}_0 \ell'_{f,n,t} \dt \\
  \notag
  &+ \frac{1}{N} \sum^N_n y_n (\ip{\x_n}{\z} + 1)
     \sum^m_i \alpha^2_i \kappa_{f,i}(\x_n) \\
     & \quad \times
       \int^{T_f}_0 \ell'_{f,n,t} ( \psi'_{f,i,t}(\x_n) \psi'_{f,i,T_f}(\z)
       - \psi'_{f,i,0}(\x_n) \psi'_{f,i,0}(\z) ) \dt.
\end{align}
\end{lemma}
\end{tcolorbox}

\begin{proof}
First, see \cref{le:kernel-regime-f}. The time evolution of $\v_i(t)$ from $t = 0$ to $t = T_f$ can be computed as
\begin{align}
  \Delta \v_i(T_f)
  :=& - \int^{T_f}_0 \bnabla_{\v_i}
      \calL(\btheta_{\V,\a}(t);\calD) \dt \\
  =& \int^{T_f}_0 \frac{\alpha_i}{N} \sum^N_n y_n
     \ell'_{f,n,t} \psi'_{f,i,t}(\x_n) \x_n \dt \\
  =& \frac{\alpha_i}{N} \sum^N_n y_n \x_n
     \int^{T_f}_0 \ell'_{f,n,t}
     \psi'_{f,i,t}(\x_n) \dt \\
  \notag
  =& \frac{\alpha_i}{N} \sum^N_n
     (1-\kappa_{f,i}(\x_n)) y_n \x_n \psi'_{f,i,0}(\x_n)
     \int^{T_f}_0 \ell'_{f,n,t} \dt \\
  &+ \frac{\alpha_i}{N} \sum^N_n
     \kappa_{f,i}(\x_n) y_n \x_n
     \int^{T_f}_0 \ell'_{f,n,t} 
     \psi'_{f,i,t}(\x_n) \dt \\
  \notag
  =& \frac{\alpha_i}{N} \sum^N_n y_n \x_n
     \psi'_{f,i,0}(\x_n)
     \int^{T_f}_0 \ell'_{f,n,t} \dt \\
  \notag
  &- \frac{\alpha_i}{N} \sum^N_n
     \kappa_{f,i}(\x_n) y_n \x_n
     \psi'_{f,i,0}(\x_n)
     \int^{T_f}_0 \ell'_{f,n,t} \dt \\
  &+ \frac{\alpha_i}{N} \sum^N_n
     \kappa_{f,i}(\x_n) y_n \x_n
     \int^{T_f}_0 \ell'_{f,n,t} 
     \psi'_{f,i,t}(\x_n) \dt \\
  \notag
  =& \frac{\alpha_i}{N} \sum^N_n y_n \x_n
     \psi'_{f,i,0}(\x_n)
     \int^{T_f}_0 \ell'_{f,n,t} \dt \\
  &+ \frac{\alpha_i}{N} \sum^N_n
     \kappa_{f,i}(\x_n) y_n \x_n
     \int^{T_f}_0 \ell'_{f,n,t}
     ( \psi'_{f,i,t}(\x_n)
     - \psi'_{f,i,0}(\x_n) ) \dt.
\end{align}
Similarly, the time evolution of $a_i(t)$ from $t = 0$ to $t = T_f$ can be computed as
\begin{align}
  \Delta a_i(T_f)
  =& \frac{\alpha_i}{N} \sum^N_n y_n
     \int^{T_f}_0 \ell'_{f,n,t}
     \psi'_{f,i,t}(\x_n) \dt \\
  \notag
  =& \frac{\alpha_i}{N} \sum^N_n y_n
     \psi'_{f,i,0}(\x_n)
     \int^{T_f}_0 \ell'_{f,n,t} \dt \\
  &+ \frac{\alpha_i}{N} \sum^N_n
     \kappa_{f,i}(\x_n) y_n
     \int^{T_f}_0 \ell'_{f,n,t}
     ( \psi'_{f,i,t}(\x_n)
     - \psi'_{f,i,0}(\x_n) ) \dt.
\end{align}
Thus,
\begin{align}
  \notag
  & \Delta h_{f,i,T_f}(\z) \\
  =& \frac{\alpha_i}{N} \sum^N_n y_n
     (\ip{\x_n}{\z} + 1)
     \int^{T_f}_0 \ell'_{f,n,t}
     \psi'_{f,i,t}(\x_n) \dt \\
  \notag
  =& \frac{\alpha_i}{N} \sum^N_n y_n
     (\ip{\x_n}{\z} + 1)
     \psi'_{f,i,0}(\x_n)
     \int^{T_f}_0 \ell'_{f,n,t} \dt \\
  &+ \frac{\alpha_i}{N} \sum^N_n
     \kappa_{f,i}(\x_n) y_n
     (\ip{\x_n}{\z} + 1)
     \int^{T_f}_0 \ell'_{f,n,t}
     ( \psi'_{f,i,t}(\x_n)
     - \psi'_{f,i,0}(\x_n) ) \dt.
\end{align}
The original network at the training time $T_f$ can be computed as
\begin{align}
  f(\z;T_f)
  :=& \sum^m_i \alpha_i \psi_{f,i,T_f}(\z) \\
  \label{eq:tmp-dhgo}
  =& \sum^m_i \alpha_i \psi'_{f,i,T_f}(\z) h_{f,i,0}(\z)
     + \sum^m_i \alpha_i \psi'_{f,i,T_f}(\z) \Delta h_{f,i,T_f}(\z)).
\end{align}
The first term of \cref{eq:tmp-dhgo} can be rearranged as
\begin{align}
  \notag
  & \sum^m_i \alpha_i \psi'_{f,i,T_f}(\z) h_{f,i,0}(\z) \\
  =& \sum^m_i \alpha_i (1 - \kappa_{f,i}(\z)) \psi'_{f,i,0}(\z) h_{f,i,0}(\z)
     + \sum^m_i \alpha_i \kappa_{f,i}(\z) \psi'_{f,i,T_f}(\z) h_{f,i,0}(\z) \\
  =& f(\z;0)
     + \sum^m_i \alpha_i \kappa_{f,i}(\z)
       (\psi'_{f,i,T_f}(\z) - \psi'_{f,i,0}(\z)) h_{f,i,0}(\z).
\end{align}
The second term of \cref{eq:tmp-dhgo} can be rearranged as
\begin{align}
  \notag
  & \sum^m_i \alpha_i \psi'_{f,i,T_f}(\z) \Delta h_{f,i,T_f}(\z)) \\
  =& \sum^m_i \alpha_i (1 - \kappa_{f,i}(\z))
    \psi'_{f,i,0}(\z) \Delta h_{f,i,T_f}(\z)
    + \sum^m_i \alpha_i \kappa_{f,i}(\z)
      \psi'_{f,i,T_f}(\z) \Delta h_{f,i,T_f}(\z) \\
  \label{eq:tmp-qtiu}
  =& \sum^m_i \alpha_i \psi'_{f,i,0}(\z)
     \Delta h_{f,i,T_f}(\z)
  + \sum^m_i \alpha_i \kappa_{f,i}(\z)
    ( \psi'_{f,i,T_f}(\z) - \psi'_{f,i,0}(\z) )
    \Delta h_{f,i,T_f}(\z).
\end{align}
The first term of \cref{eq:tmp-qtiu} can be rearranged as
\begin{align}
  \notag
  & \sum^m_i \alpha_i \psi'_{f,i,0}(\z)
    \Delta h_{f,i,T_f}(\z) \\
  \notag
  =& \sum^m_i \alpha_i \psi'_{f,i,0}(\z) \Bigg[
     \frac{\alpha_i}{N} \sum^N_n y_n
     (\ip{\x_n}{\z} + 1) \psi'_{f,i,0}(\x_n)
     \int^{T_f}_0 \ell'_{f,n,t} \dt \\
     &+ \frac{\alpha_i}{N} \sum^N_n
        \kappa_{f,i}(\x_n) y_n
        (\ip{\x_n}{\z} + 1)
        \int^{T_f}_0 \ell'_{f,n,t}
          ( \psi'_{f,i,t}(\x_n)
          - \psi'_{f,i,0}(\x_n) ) \dt \Bigg] \\
  \notag
  =& \frac{1}{N} \sum^N_n y_n (\ip{\x_n}{\z} + 1)
     \int^{T_f}_0 \ell'_{f,n,t} \dt \sum^m_i
     \alpha^2_i \psi'_{f,i,0}(\x_n) \psi'_{f,i,0}(\z) \\
     \notag
     &+ \frac{1}{N} \sum^N_n
        y_n (\ip{\x_n}{\z} + 1) \\
        \label{eq:tmp-qggi}
        & \quad \times
        \sum^m_i \alpha^2_i \kappa_{f,i}(\x_n)
        \psi'_{f,i,0}(\z) \int^{T_f}_0 \ell'_{f,n,t}
        ( \psi'_{f,i,t}(\x_n)
        - \psi'_{f,i,0}(\x_n) ) \dt.
\end{align}
The first term of \cref{eq:tmp-qggi} can be rearranged as
\begin{align}
  & \frac{1}{N} \sum^N_n y_n (\ip{\x_n}{\z} + 1)
    \int^{T_f}_0 \ell'_{f,n,t} \dt
    \sum^m_i \alpha^2_i \psi'_{f,i,0}(\x_n) \psi'_{f,i,0}(\z) \\
  =& \frac{1}{N} \sum^N_n y_n (\ip{\x_n}{\z} + 1)
     \int^{T_f}_0 \ell'_{f,n,t} \dt
     \qty( \sum^m_i \alpha^2_i \psi'_{f,i,0}(\x_n) \psi'_{f,i,0}(\z)
     - \Phi(\x_n, \z) ) \\
  &+ \frac{1}{N} \sum^N_n y_n \Phi(\x_n, \z) (\ip{\x_n}{\z} + 1)
     \int^{T_f}_0 \ell'_{f,n,t} \dt.
\end{align}
The second term of \cref{eq:tmp-qtiu} can be rearranged as
\begin{align}
  \notag
  & \sum^m_i \alpha_i \kappa_{f,i}(\z)
    ( \psi'_{f,i,T_f}(\z) - \psi'_{f,i,0}(\z) )
    \Delta h_{f,i,T_f}(\z) \\
  \notag
  =& \sum^m_i \alpha_i \kappa_{f,i}(\z)
     ( \psi'_{f,i,T_f}(\z) - \psi'_{f,i,0}(\z) ) \\
     & \times
     \frac{\alpha_i}{N} \sum^N_n y_n
     (\ip{\x_n}{\z} + 1) \int^{T_f}_0 \ell'_{f,n,t}
     \psi'_{f,i,t}(\x_n) \dt \\
  \notag
  =& \frac{1}{N} \sum^N_n y_n (\ip{\x_n}{\z} + 1) \\
     \label{eq:tmp-nbrh}
     & \times
     \sum^m_i \alpha^2_i \kappa_{f,i}(\z)
     ( \psi'_{f,i,T_f}(\z) - \psi'_{f,i,0}(\z) )
     \int^{T_f}_0 \ell'_{f,n,t} \psi'_{f,i,t}(\x_n) \dt.
\end{align}
The sum of the second term of \cref{eq:tmp-qggi} and \cref{eq:tmp-nbrh} can be rearranged as
\begin{align}
  \notag
  & \frac{1}{N} \sum^N_n y_n (\ip{\x_n}{\z} + 1)
  \sum^m_i \alpha^2_i \kappa_{f,i}(\x_n)
  \psi'_{f,i,0}(\z) \int^{T_f}_0 \ell'_{f,n,t}
  ( \psi'_{f,i,t}(\x_n) - \psi'_{f,i,0}(\x_n) ) \dt \\
  \notag
  &+ \frac{1}{N} \sum^N_n y_n (\ip{\x_n}{\z} + 1)
     \sum^m_i \alpha^2_i \kappa_{f,i}(\z)
     ( \psi'_{f,i,T_f}(\z) - \psi'_{f,i,0}(\z) )
     \int^{T_f}_0 \ell'_{f,n,t} \psi'_{f,i,t}(\x_n) \dt \\
  \notag
  =& \frac{1}{N} \sum^N_n y_n (\ip{\x_n}{\z} + 1)
     \sum^m_i \alpha^2_i \kappa_{f,i}(\x_n) \\
     & \times
     \int^{T_f}_0 \ell'_{f,n,t} ( \psi'_{f,i,t}(\x_n) \psi'_{f,i,T_f}(\z)
     - \psi'_{f,i,0}(\x_n) \psi'_{f,i,0}(\z) ) \dt.
\end{align}
\end{proof}

\begin{tcolorbox}
\begin{lemma}[Upper bounds of terms in \cref{le:representation-network-f}]
\label{le:bounds-network-output}
Assume \cref{asm:properties-network-f,asm:width-ap-f}.
\begin{enumerate}[label=(\alph*)]

  \item \label{item:bounds-kdbg}
  \begin{align}
    \notag
    & \abs{
      \sum^m_i \alpha_i \kappa_{f,i}(\z)
      (\psi'_{f,i,T_f}(\z) - \psi'_{f,i,0}(\z))
      h_{f,i,0}(\z)
    } \\
    < & 2 \eta (1-\gamma) \ln(2m/\delta) \sqrt{\|\z\|^2/d + 1}
    = \tildecalO(1).
  \end{align}
  
  \item \label{item:bounds-ytgb}
  \begin{align}
    \notag
    & \abs{ \frac{1}{N} \sum^N_n y_n
    (\ip{\x_n}{\z} + 1) \int^{T_f}_0 \ell'_{f,n,t} \dt 
    \qty( \sum^m_i \alpha^2_i \psi'_{f,i,0}(\x_n)
    \psi'_{f,i,0}(\z) - \Phi(\x_n, \z) ) } \\
    < &
    \frac{ 8 \Cthr(\z,\delta) \ln(2/\delta)  }{ \sqrt{ \ln(2m/\delta) } }
    = \tildecalO(1).
  \end{align}
  
  \item \label{item:bounds-jhre}
  \begin{align}
    \notag
    & \Bigg | \frac{1}{N} \sum^N_n y_n (\ip{\x_n}{\z} + 1)
     \sum^m_i \alpha^2_i \kappa_{f,i}(\x_n) \\
     \notag
     & \times
     \int^{T_f}_0 \ell'_{f,n,t} ( \psi'_{f,i,t}(\x_n) \psi'_{f,i,T_f}(\z)
     - \psi'_{f,i,0}(\x_n) \psi'_{f,i,0}(\z) ) \dt \Bigg | \\
    < &
    \eta (1-\gamma^2) \Cthr(\z,\delta) \sqrt{\ln(2m/\delta)}
    = \tildecalO(1).
  \end{align}
  
\end{enumerate}
\end{lemma}
\end{tcolorbox}

\begin{proof}
~

\cref*{item:bounds-kdbg}
By \cref{asm:properties-network-f},
\begin{align}
  \abs{
    \sum^m_i \alpha_i \kappa_{f,i}(\z)
    (\psi'_{f,i,T_f}(\z) - \psi'_{f,i,0}(\z)) h_{f,i,0}(\z)
  }
  \leq &
  (1-\gamma) \sum^m_i |\alpha_i \kappa_{f,i}(\z) h_{f,i,0}(\z)| \\
  < &
  2 \eta (1-\gamma) \ln(2m/\delta) \sqrt{\|\z\|^2/d+1}.
\end{align}

\cref*{item:bounds-ytgb}
By \cref{asm:properties-network-f},
\begin{align}
  \notag
  & \abs{ \frac{1}{N} \sum^N_n y_n
  (\ip{\x_n}{\z} + 1) \int^{T_f}_0 \ell'_{f,n,t} \dt 
  \qty( \sum^m_i \alpha^2_i \psi'_{f,i,0}(\x_n) \psi'_{f,i,0}(\z)
    - \Phi(\x_n, \z) ) } \\
  < &
  \frac{16}{N\sqrt{m}} \ln(\frac{2}{\delta})
  \sum^N_n |\ip{\x_n}{\z} + 1| \int^{T_f}_0 \ell'_{f,n,t} \dt.
\end{align}
By \cref{asm:width-ap-f},
\begin{align}
  \notag
  & \frac{16}{N\sqrt{m}} \ln(\frac{2}{\delta})
  \sum^N_n |\ip{\x_n}{\z} + 1| \int^{T_f}_0 \ell'_{f,n,t} \dt \\
  \notag
  < &
  \frac
  { N \Cthr(\z,\delta) }
  { 2 \sqrt{ \ln(2m/\delta) } \sum^N_n
    (|\ip{\x_n}{\z}| + 1) \int^{T_f}_0 \ell'_{f,n,t} \dt } \\
  & \times \frac{16}{N} \ln(\frac{2}{\delta})
  \sum^N_n |\ip{\x_n}{\z} + 1| \int^{T_f}_0 \ell'_{f,n,t} \dt \\
  \leq &
  \frac{ 8 \Cthr(\z,\delta) \ln(2/\delta) }{ \sqrt{ \ln(2m/\delta) } }.
\end{align}

\cref*{item:bounds-jhre}
By \cref{asm:properties-network-f},
\begin{align}
  \notag
  & \abs{ \sum^m_i \alpha^2_i \kappa_{f,i}(\x_n)
    \int^{T_f}_0 \ell'_{f,n,t}
    ( \psi'_{f,i,t}(\x_n) \psi'_{f,i,T_f}(\z)
      - \psi'_{f,i,0}(\x_n) \psi'_{f,i,0}(\z) ) \dt } \\
  \leq &
  (1-\gamma^2) \int^{T_f}_0 \ell'_{f,n,t} \dt
  \sum^m_i \alpha^2_i \kappa_{f,i}(\x_n) \\
  < &
  \frac{ 2 \eta (1-\gamma^2) }{ \sqrt{m} } \ln\qty(\frac{2m}{\delta})
  \int^{T_f}_0 \ell'_{f,n,t} \dt.
\end{align}
By \cref{asm:width-ap-f},
\begin{align}
  \notag
  &
  \frac{ 2 \eta (1-\gamma^2) }{ N \sqrt{m} }
  \ln\qty(\frac{2m}{\delta})
  \sum^N_n |\ip{\x_n}{\z} + 1| \int^{T_f}_0 \ell'_{f,n,t} \dt \\
  \notag
  < &
  \frac
  { N \Cthr(\z,\delta) }
  { 2 \sqrt{ \ln(2m/\delta) } \sum^N_n
    (|\ip{\x_n}{\z}| + 1) \int^{T_f}_0 \ell'_{f,n,t} \dt } \\
  & \times 
  \frac{ 2 \eta (1-\gamma^2) }{ N } \ln\qty(\frac{2m}{\delta})
  \sum^N_n |\ip{\x_n}{\z} + 1| \int^{T_f}_0 \ell'_{f,n,t} \dt \\
  \leq &
  \eta (1-\gamma^2) \Cthr(\z,\delta) \sqrt{\ln(2m/\delta)}.
\end{align}

\end{proof}

\begin{tcolorbox}
\begin{lemma}[Network prediction of $f$]
\label{le:prediction-f}
Assume \cref{asm:properties-network-f,asm:width-ap-f}. If
\begin{align}
  \abs{
    \frac{1}{N} \sum^N_n y_n \Phi(\x_n, \z) \ip{\x_n}{\z}
    \int^{T_f}_0 \ell'_{f,n,t} \dt
  } \geq \tildecalO\qty(1 + \int^{T_f}_0 \barell'_{f,n,t} \dt ),
\end{align}
then
\begin{align}
  \sgn(f(\z;T_f))
  = \sgn\qty( \frac{1}{N} \sum^N_n y_n \Phi(\x_n, \z) \ip{\x_n}{\z}
    \int^{T_f}_0 \ell'_{f,n,t} \dt ).
\end{align}
\end{lemma}
\end{tcolorbox}

\begin{proof}
By \cref{asm:properties-network-f,le:representation-network-f,le:bounds-network-output}, if
\begin{align}
  \notag
  & \abs{ \frac{1}{N} \sum^N_n y_n \Phi(\x_n,\z) \ip{\x_n}{\z}
    \int^{T_f}_0 \ell'_{f,n,t} \dt } \\
  \notag
  \geq &
  |f(\z;0)|
  + \abs{ \sum^m_i \alpha_i \kappa_{f,i}(\z)
          (\psi'_{f,i,T_f}(\z) - \psi'_{f,i,0}(\z)) h_{f,i,0}(\z) } \\
  \notag
  &+ \abs{ \frac{1}{N} \sum^N_n y_n (\ip{\x_n}{\z} + 1)
     \int^{T_f}_0 \ell'_{f,n,t} \dt \qty(
     \sum^m_i \alpha^2_i \psi'_{f,i,0}(\x_n) \psi'_{f,i,0}(\z)
     - \Phi(\x_n, \z) ) } \\
  \notag
  &+ \Bigg| \frac{1}{N} \sum^N_n y_n (\ip{\x_n}{\z} + 1)
     \sum^m_i \alpha^2_i \kappa_{f,i}(\x_n) \\
     \notag
     & \quad \times
     \int^{T_f}_0 \ell'_{f,n,t} ( \psi'_{f,i,t}(\x_n) \psi'_{f,i,T_f}(\z)
     - \psi'_{f,i,0}(\x_n) \psi'_{f,i,0}(\z) ) \dt \Bigg| \\
  &+ \abs{ \frac{1}{N} \sum^N_n y_n \Phi(\x_n,\z) \int^{T_f}_0 \ell'_{f,n,t} \dt } \\
  =& \tildecalO\qty(1 + \int^{T_f}_0 \barell'_{f,n,t} \dt ),
\end{align}
then
\begin{align}
  \sgn(f(\z;T_f)) = 
  \sgn\qty( \frac{1}{N} \sum^N_n y_n \Phi(\x_n,\z) \ip{\x_n}{\z}
  \int^{T_f}_0 \ell'_{f,n,t} \dt ).
\end{align}
\end{proof}

\begin{tcolorbox}
\begin{lemma}[Adversarial perturbation]
\label{le:AP}
If \cref{asm:properties-network-f,asm:width-ap-f} hold, then the adversarial perturbation defined as \cref{eq:AP} can be represented as
\begin{align}
  & \r_n
  = \epsilon \yadv_n \frac{ \sum^m_i \alpha_i \psi'_{f,i,T_f}(\x_n) \v_i(T_f) }
    {\| \sum^m_i \alpha_i \psi'_{f,i,T_f}(\x_n) \v_i(T_f) \|},
  \qquad\mathrm{and} \\
  \notag
  & \sum^m_i \alpha_i \psi'_{f,i,T_f}(\x_n) \v_i(T_f) \\
  \notag
  =& \sum^m_i \alpha_i \psi'_{f,i,0}(\x_n) \v_i(0)
     + \sum^m_i \alpha_i \kappa_{f,i}(\x_n)
       (\psi'_{f,i,T_f}(\x_n) - \psi'_{f,i,0}(\x_n)) \v_i(0) \\
  \notag
  &+ \frac{1}{N} \sum^N_k y_k \x_k \int^{T_f}_0 \ell'_{f,k,t} \dt
     \qty( \sum^m_i \alpha^2_i \psi'_{f,i,0}(\x_n) \psi'_{f,i,0}(\x_k)
     - \Phi(\x_n, \x_k) ) \\
  \notag
  &+ \frac{1}{N} \sum^N_k y_k \Phi(\x_n, \x_k) \x_k \int^{T_f}_0 \ell'_{f,k,t} \dt \\
  \notag
  &+ \frac{1}{N} \sum^N_k y_k \x_k \sum^m_i \alpha^2_i \kappa_{f,i}(\x_k) \\
     & \quad \times
     \int^{T_f}_0 \ell'_{f,k,t} ( \psi'_{f,i,T_f}(\x_n) \psi'_{f,i,t}(\x_k)
     - \psi'_{f,i,0}(\x_n) \psi'_{f,i,0}(\x_k) ) \dt.
\end{align}
\end{lemma}
\end{tcolorbox}

\begin{proof}
The main term of the adversarial perturbation can be computed as
\begin{align}
  \frac{ \bnabla_{\x_n} \ell_{f,n,T_f} }
  {\| \bnabla_{\x_n} \ell_{f,n,T_f} \|}
  =& \frac{ - \yadv_n \ell'_{f,n,T_f}
            \bnabla_{\x_n} f(\x_n;\bthetaVa(T_f)) }
     {\| \ell'_{f,n,T_f} \bnabla_{\x_n}
         f(\x_n;\bthetaVa(T_f)) \|} \\
  =& - \yadv_n
     \frac{ \sum^m_i \alpha_i
            \psi'_{f,i,T_f}(\x_n) \v_i(T_f) }
     { \| \sum^m_i \alpha_i
          \psi'_{f,i,T_f}(\x_n) \v_i(T_f) \| }.
\end{align}
The leading term can be rearranged as
\begin{align}
  \label{eq:tmp-bnsw}
  \sum^m_i \alpha_i \psi'_{f,i,T_f}(\x_n) \v_i(T_f)
  = \sum^m_i \alpha_i \psi'_{f,i,T_f}(\x_n) \v_i(0)
  + \sum^m_i \alpha_i \psi'_{f,i,T_f}(\x_n) \Delta \v_i(T_f).
\end{align}
The first term of \cref{eq:tmp-bnsw} can be rearranged as
\begin{align}
  \notag
  & \sum^m_i \alpha_i \psi'_{f,i,T_f}(\x_n) \v_i(0) \\
  =& \sum^m_i \alpha_i \psi'_{f,i,0}(\x_n) \v_i(0)
     + \sum^m_i \alpha_i \kappa_{f,i}(\x_n)
       (\psi'_{f,i,T_f}(\x_n) - \psi'_{f,i,0}(\x_n)) \v_i(0).
\end{align}
The second term of \cref{eq:tmp-bnsw} can be rearranged as
\begin{align}
  \notag
  & \sum^m_i \alpha_i \psi'_{f,i,T_f}(\x_n) \Delta \v_i(T_f) \\
  \label{eq:tmp-nbaa}
  =& \sum^m_i \alpha_i \psi'_{f,i,0}(\x_n) \Delta \v_i(T_f)
     + \sum^m_i \alpha_i \kappa_{f,i}(\x_n)
       (\psi'_{f,i,T_f}(\x_n) - \psi'_{f,i,0}(\x_n) \Delta \v_i(T_f).
\end{align}
The first term of \cref{eq:tmp-nbaa} can be rearranged as
\begin{align}
  \notag
  & \sum^m_i \alpha_i \psi'_{f,i,0}(\x_n)
    \Delta \v_i(T_f) \\
  \notag
  =& \sum^m_i \alpha_i \psi'_{f,i,0}(\x_n)
     \Bigg[
     \frac{\alpha_i}{N} \sum^N_k y_k \x_k
     \psi'_{f,i,0}(\x_k)
     \int^{T_f}_0 \ell'_{f,k,t} \dt \\
     &+ \frac{\alpha_i}{N} \sum^N_k
        \kappa_{f,i}(\x_k) y_k \x_k
        \int^{T_f}_0 \ell'_{f,k,t}
        ( \psi'_{f,i,t}(\x_k)
        - \psi'_{f,i,0}(\x_k) ) \dt
     \Bigg] \\
  \notag
  =& \frac{1}{N} \sum^N_k y_k \x_k
     \int^{T_f}_0 \ell'_{f,k,t} \dt
     \sum^m_i \alpha^2_i \psi'_{f,i,0}(\x_n)
     \psi'_{f,i,0}(\x_k) \\
     \label{eq:tmp-aart}
     &+ \frac{1}{N} \sum^N_k y_k \x_k
        \sum^m_i \alpha^2_i \kappa_{f,i}(\x_k)
        \psi'_{f,i,0}(\x_n) \int^{T_f}_0
        \ell'_{f,k,t} ( \psi'_{f,i,t}(\x_k)
        - \psi'_{f,i,0}(\x_k) ) \dt.
\end{align}
The second term of \cref{eq:tmp-nbaa} can be rearranged as
\begin{align}
  \notag
  & \sum^m_i \alpha_i \kappa_{f,i}(\x_n)
    (\psi'_{f,i,T_f}(\x_n) - \psi'_{f,i,0}(\x_n))
    \Delta \v_i(T_f) \\
  =& \sum^m_i \alpha_i \kappa_{f,i}(\x_n)
     (\psi'_{f,i,T_f}(\x_n) - \psi'_{f,i,0}(\x_n))
     \frac{\alpha_i}{N} \sum^N_k y_k \x_k
     \int^{T_f}_0 \ell'_{f,k,t}
     \psi'_{f,i,t}(\x_k) \dt \\
  \label{eq:tmp-baaa}
  =& \frac{1}{N} \sum^N_k y_k \x_k
     \sum^m_i \alpha^2_i \kappa_{f,i}(\x_n)
     (\psi'_{f,i,T_f}(\x_n) - \psi'_{f,i,0}(\x_n))
     \int^{T_f}_0 \ell'_{f,k,t}
     \psi'_{f,i,t}(\x_k) \dt.
\end{align}
The sum of the second term of \cref{eq:tmp-aart} and \cref{eq:tmp-baaa} can be rearranged as
\begin{align}
  \notag
  & \frac{1}{N} \sum^N_k y_k \x_k \sum^m_i \alpha^2_i \kappa_{f,i}(\x_k)
    \psi'_{f,i,0}(\x_n) \int^{T_f}_0 \ell'_{f,k,t} ( \psi'_{f,i,t}(\x_k)
    - \psi'_{f,i,0}(\x_k) ) \dt \\
  \notag
  &+ \frac{1}{N} \sum^N_k y_k \x_k \sum^m_i \alpha^2_i \kappa_{f,i}(\x_n)
     (\psi'_{f,i,T_f}(\x_n) - \psi'_{f,i,0}(\x_n))
     \int^{T_f}_0 \ell'_{f,k,t} \psi'_{f,i,t}(\x_k) \dt \\
  \notag
  =& \frac{1}{N} \sum^N_k y_k \x_k \sum^m_i \alpha^2_i \kappa_{f,i}(\x_k) \\
     & \times
     \int^{T_f}_0 \ell'_{f,k,t} ( \psi'_{f,i,T_f}(\x_n) \psi'_{f,i,t}(\x_k)
     - \psi'_{f,i,0}(\x_n) \psi'_{f,i,0}(\x_k) ) \dt.
\end{align}
\end{proof}

\begin{tcolorbox}
\begin{lemma}[Upper bound of norm of adversarial perturbation]
\label{le:norm-AP}
Assume \cref{asm:properties-network-f,asm:width-ap-f}.
\begin{enumerate}[label=(\alph*)]

  \item \label{item:norm-AP-jhrt}
  \begin{align}
    \norm{ \sum^m_i \alpha_i \psi'_{f,i,0}(\x_n) \v_i(0) }^2
    < 4 \ln(2dm/\delta) \ln(2/\delta) = \tildecalO(1).
  \end{align}
  
  \item \label{item:norm-AP-bgrt}
  \begin{align}
    \notag
    & \norm{ \sum^m_i \alpha_i \kappa_{f,i}(\x_n)
    (\psi'_{f,i,T_f}(\x_n) - \psi'_{f,i,0}(\x_n)) \v_i(0) }^2 \\
    <& 4 \eta^2 (1 - \gamma)^2 \ln(2m/\delta) \ln(2dm/\delta)
    = \tildecalO(1).
  \end{align}
  
  \item \label{item:norm-AP-loji}
  \begin{align}
    \notag
    & \norm{ \frac{1}{N} \sum^N_k y_k \x_k \int^{T_f}_0 \ell'_{f,k,t} \dt
    \qty( \sum^m_i \alpha^2_i \psi'_{f,i,0}(\x_n)
    \psi'_{f,i,0}(\x_k) - \Phi(\x_n, \x_k) ) }^2 \\
    <& 64 \ln^2\qty(2/\delta)
    = \tildecalO(1).
  \end{align}
  
  \item \label{item:norm-AP-zxer}
  \begin{align}
    \notag
    & \Bigg \| \frac{1}{N} \sum^N_k y_k \x_k \sum^m_i \alpha^2_i \kappa_{f,i}(\x_k) \\
    \notag
    & \times
    \int^{T_f}_0 \ell'_{f,k,t} ( \psi'_{f,i,T_f}(\x_n) \psi'_{f,i,t}(\x_k)
    - \psi'_{f,i,0}(\x_n) \psi'_{f,i,0}(\x_k) ) \dt \Bigg \|^2 \\
    < & \eta^2 (1 - \gamma^2)^2 \ln^2(2m/\delta)
    = \tildecalO(1).
  \end{align}
  
  \item \label{item:norm-AP-aqgg}
  \begin{align}
    \norm{ \sum^m_i \alpha_i \psi'_{f,i,T_f}(\x_n) \v_i(T_f) }
    < \tildecalO\qty(\sqrt{d} \int^{T_f}_0 \barell'_{f,k,t} \dt).
  \end{align}
  
  \item \label{item:norm-AP-mmnl}
  Let
  \begin{align}
    \notag
    \r'_n :=& \sum^m_i \alpha_i \psi'_{f,i,0}(\x_n) \v_i(0)
    + \sum^m_i \alpha_i \kappa_{f,i}(\x_n)
      (\psi'_{f,i,T_f}(\x_n) - \psi'_{f,i,0}(\x_n)) \v_i(0) \\
    \notag
    &+ \frac{1}{N} \sum^N_k y_k \x_k \int^{T_f}_0 \ell'_{f,k,t} \dt
       \qty( \sum^m_i \alpha^2_i \psi'_{f,i,0}(\x_n) \psi'_{f,i,0}(\x_k)
       - \Phi(\x_n, \x_k) ) \\
    \notag
    &+ \frac{1}{N} \sum^N_k y_k \x_k \sum^m_i \alpha^2_i \kappa_{f,i}(\x_k) \\
       & \quad \times
       \int^{T_f}_0 \ell'_{f,k,t} ( \psi'_{f,i,T_f}(\x_n) \psi'_{f,i,t}(\x_k)
       - \psi'_{f,i,0}(\x_n) \psi'_{f,i,0}(\x_k) ) \dt.
  \end{align}
  Then, $\|\r'_n\| < \tildecalO(1)$.
  
\end{enumerate}
\end{lemma}
\end{tcolorbox}

\begin{proof}
~

\cref*{item:norm-AP-jhrt}
The given left term can be rearranged as
\begin{align}
  \norm{ \sum^m_i \alpha_i \psi'_{f,i,0}(\x_n) \v_i(0) }^2
  = \sum^d_j \qty( \sum^m_i \alpha_i \psi'_{f,i,0}(\x_n) v_{i,j}(0) )^2.
\end{align}
By \cref{asm:properties-network-f},
\begin{align}
  \qty( \sum^m_i \alpha_i \psi'_{f,i,0}(\x_n) v_{i,j}(0) )^2
  <& (2/m) \ln(2/\delta)
     \sum^m_i \psi'_{f,i,0}(\x_n)^2 v_{i,j}(0)^2 \\
  <& (4/d) \ln(2dm/\delta) \ln(2/\delta).
\end{align}

\cref*{item:norm-AP-bgrt}
The given left term can be rearranged as
\begin{align}
  \notag
  & \norm{ \sum^m_i \alpha_i \kappa_{f,i}(\x_n)
  (\psi'_{f,i,T_f}(\x_n) - \psi'_{f,i,0}(\x_n))
  \v_i(0) }^2 \\
  =&
  \sum^d_j \qty(
    \sum^m_i \alpha_i \kappa_{f,i}(\x_n)
    (\psi'_{f,i,T_f}(\x_n) - \psi'_{f,i,0}(\x_n))
    v_{i,j}(0)
  )^2 \\
  \leq &
  (1 - \gamma)^2 \sum^d_j \qty(
    \sum^m_i |\alpha_i \kappa_{f,i}(\x_n) v_{i,j}(0)|
  )^2.
\end{align}
By \cref{asm:properties-network-f},
\begin{align}
  \qty( \sum^m_i |\alpha_i \kappa_{f,i}(\x_n) v_{i,j}(0)| )^2
  < (4\eta^2/d) \ln(2m/\delta) \ln(2dm/\delta).
\end{align}

\cref*{item:norm-AP-loji}
By \cref{asm:properties-network-f},
\begin{align}
  \notag
  & \norm{ \frac{1}{N} \sum^N_k y_k \x_k \int^{T_f}_0 \ell'_{f,k,t} \dt
  \qty( \sum^m_i \alpha^2_i \psi'_{f,i,0}(\x_n)
  \psi'_{f,i,0}(\x_k) - \Phi(\x_n, \x_k) ) }^2 \\
  < & \frac{256}{mN^2} \ln^2\qty(\frac{2}{\delta})
      \sum^N_{n,k} |\ip{\x_n}{\x_k}|
      \int^{T_f}_0 \ell'_{f,n,t} \dt \int^{T_f}_0 \ell'_{f,k,t} \dt.
\end{align}
By \cref{asm:width-ap-f},
\begin{align}
  \notag
  & \frac{256}{N^2} \ln^2\qty(\frac{2}{\delta}) \sum^N_{n,k} |\ip{\x_n}{\x_k}|
    \int^{T_f}_0 \ell'_{f,n,t} \dt \int^{T_f}_0 \ell'_{f,k,t} \dt \\
  \notag
  & \times
  \frac
  { N^2 }
  { 4 \sum^N_{n,k} |\ip{\x_n}{\x_k}|
    \int^{T_f}_0 \ell'_{f,n,t} \dt \int^{T_f}_0 \ell'_{f,k,t} \dt } \\
  < & 64 \ln^2(2/\delta).
\end{align}

\cref*{item:norm-AP-zxer}
The given left term can be rearranged as
\begin{align}
  \notag
  & \abs{ \sum^m_i \alpha^2_i \kappa_{f,i}(\x_k)
  \int^{T_f}_0 \ell'_{f,k,t} ( \psi'_{f,i,T_f}(\x_n) \psi'_{f,i,t}(\x_k)
  - \psi'_{f,i,0}(\x_n) \psi'_{f,i,0}(\x_k) ) \dt } \\
  \leq &
  (1 - \gamma^2) \sum^m_i \alpha^2_i \kappa_{f,i}(\x_k)
  \int^{T_f}_0 \ell'_{f,k,t} \dt.
\end{align}
By \cref{asm:properties-network-f},
\begin{align}
  \sum^m_i \alpha^2_i \kappa_{f,i}(\x_k)
  < \frac{ 2 \eta }{ \sqrt{m} } \ln(\frac{2m}{\delta}).
\end{align}
Thus,
\begin{align}
  \notag
  &  \norm{ \frac{1}{N} \sum^N_k y_k \x_k \sum^m_i \alpha^2_i \kappa_{f,i}(\x_k)
  \int^{T_f}_0 \ell'_{f,k,t} ( \psi'_{f,i,T_f}(\x_n) \psi'_{f,i,t}(\x_k)
  - \psi'_{f,i,0}(\x_n) \psi'_{f,i,0}(\x_k) ) \dt }^2 \\
  < &
  \frac{ 4 \eta^2 (1 - \gamma^2)^2 }{ m N^2 } \ln^2\qty(\frac{2m}{\delta})
  \sum^N_{n,k} |\ip{\x_n}{\x_k}|
  \int^{T_f}_0 \ell'_{f,n,t} \dt \int^{T_f}_0 \ell'_{f,k,t} \dt.
\end{align}
By \cref{asm:width-ap-f},
\begin{align}
  \notag
  & \frac{ 4 \eta^2 (1 - \gamma^2)^2 }{ m N^2 } \ln^2\qty(\frac{2m}{\delta})
  \sum^N_{n,k} |\ip{\x_n}{\x_k}|
  \int^{T_f}_0 \ell'_{f,n,t} \dt \int^{T_f}_0 \ell'_{f,k,t} \dt \\
  \notag
  < &
  \frac{ 4 \eta^2 (1 - \gamma^2)^2 }{ N^2 } \ln^2\qty(\frac{2m}{\delta})
  \sum^N_{n,k} |\ip{\x_n}{\x_k}|
  \int^{T_f}_0 \ell'_{f,n,t} \dt \int^{T_f}_0 \ell'_{f,k,t} \dt \\
  & \times
  \frac
  { N^2 }
  { 4 \sum^N_{n,k} |\ip{\x_n}{\x_k}|
    \int^{T_f}_0 \ell'_{f,n,t} \dt \int^{T_f}_0 \ell'_{f,k,t} \dt } \\
  =& \eta^2 (1 - \gamma^2)^2 \ln^2(2m/\delta).
\end{align}

\cref*{item:norm-AP-aqgg}
As $\|\s_1+\s_2+\s_3+\s_4+s_5\|^2 \leq 25 \max(\|\s_1\|^2, \|\s_2\|^2, \|\s_3\|^2, \|\s_4\|^2, \|\s_5\|^2)$ for any $\s_1,\s_2,\s_3,\s_4,\s_5 \in \bbR^d$,
\begin{align}
  \norm{ \sum^m_i \alpha_i \psi'_{f,i,T_f}(\x_n) \v_i(T_f) }^2
  < & 25 \max\qty(
    \begin{aligned}
      4 \ln(2dm/\delta) \ln(2/\delta), \\
      4 \eta^2 (1 - \gamma)^2 \ln(2m/\delta) \ln(2dm/\delta), \\
      64 \ln^2\qty(2/\delta), \\
      \eta^2 (1 - \gamma^2)^2 \ln^2(2m/\delta), \\
      \norm{ \frac{1}{N} \sum^N_k y_k \Phi(\x_n, \x_k) 
        \x_k \int^{T_f}_0 \ell'_{f,k,t} \dt }^2 \\
    \end{aligned}
  ).
\end{align}

\cref*{item:norm-AP-mmnl}
Similarly to \cref*{item:norm-AP-aqgg},
\begin{align}
  \|\r'_n\|^2 <
  16 \max\qty(
    \begin{aligned}
      4 \ln(2dm/\delta) \ln(2/\delta), \\
      4 \eta^2 (1 - \gamma)^2 \ln(2m/\delta) \ln(2dm/\delta), \\
      64 \ln^2\qty(2/\delta), \\
      \eta^2 (1 - \gamma^2)^2 \ln^2(2m/\delta)
    \end{aligned}
  ).
\end{align}

\end{proof}

\begin{tcolorbox}
\begin{lemma}[Upper bounds of inner products with adversarial perturbation]
\label{le:ip-AP}
Assume \cref{asm:properties-network-f,asm:width-ap-f}.
\begin{enumerate}[label=(\alph*)]

  \item \label{item:ip-AP-nbqq}
  \begin{align}
    \abs{ \sum^m_i \alpha_i \psi'_{f,i,0}(\x_n) \ip{\v_i(0)}{\z} }
    < 2 \sqrt{ (1/d) \ln(2/\delta) \ln(2m/\delta) } \|\z\|
    = \tildecalO(1).
  \end{align}
  
  \item \label{item:ip-AP-oooi}
  \begin{align}
    \notag
    & \abs{ \sum^m_i \alpha_i \kappa_{f,i}(\x_n)
      (\psi'_{f,i,T_f}(\x_n) - \psi'_{f,i,0}(\x_n)) \ip{\v_i(0)}{\z} } \\
    <& 2 \eta (1 - \gamma) \ln(2m/\delta) \|\z\| / \sqrt{d}
    = \tildecalO(1).
  \end{align}
  
  \item \label{item:ip-AP-lkjr}
  \begin{align}
    \notag
    & \abs{ \frac{1}{N} \sum^N_k y_k \ip{\x_k}{\z} \int^{T_f}_0 \ell'_{f,k,t} \dt
    \qty( \sum^m_i \alpha^2_i \psi'_{f,i,0}(\x_n) \psi'_{f,i,0}(\x_k)
          - \Phi(\x_n,\x_k) ) } \\
    <& \frac{ 8 \Cthr(\z,\delta) \ln(2/\delta) }{ \sqrt{ \ln(2m/\delta) } }
    = \tildecalO(1).
  \end{align}
  
  \item \label{item:ip-AP-bvuu}
  \begin{align}
    \notag
    & \Bigg| \frac{1}{N} \sum^N_k y_k \ip{\x_k}{\z}
    \sum^m_i \alpha^2_i \kappa_{f,i}(\x_k) \\
    \notag
    & \quad \times
    \int^{T_f}_0 \ell'_{f,k,t}
    ( \psi'_{f,i,T_f}(\x_n) \psi'_{f,i,t}(\x_k)
    - \psi'_{f,i,0}(\x_n) \psi'_{f,i,0}(\x_k) ) \dt \Bigg| \\
    < & \eta (1-\gamma^2) \Cthr(\z,\delta) \sqrt{\ln(2m/\delta)}
    = \tildecalO(1).
  \end{align}

\end{enumerate}
\end{lemma}
\end{tcolorbox}

\begin{proof}
~

\cref*{item:ip-AP-nbqq}
By \cref{asm:properties-network-f},
\begin{align}
  \abs{ \sum^m_i \alpha_i \psi'_{f,i,0}(\x_n) \ip{\v_i(0)}{\z} }
  <& \sqrt{ \frac{2}{m} \ln(\frac{2}{\delta})
            \sum^m_i \psi'_{f,i,0}(\x_n)^2 \ip{\v_i(0)}{\z}^2 } \\
  \leq & \sqrt{ \frac{2}{m} \ln(\frac{2}{\delta})
                \sum^m_i \ip{\v_i(0)}{\z}^2 } \\
  <& 2 \sqrt{ (1/d) \ln(2/\delta) \ln(2m/\delta) } \|\z\|.
\end{align}

\cref*{item:ip-AP-oooi}
By \cref{asm:properties-network-f},
\begin{align}
  \notag
  & \abs{ \sum^m_i \alpha_i \kappa_{f,i}(\x_n)
          (\psi'_{f,i,T_f}(\x_n) - \psi'_{f,i,0}(\x_n)) \ip{\v_i(0)}{\z} } \\
  \leq & (1 - \gamma) \sum^m_i \abs{ \alpha_i \kappa_{f,i}(\x_n) \ip{\v_i(0)}{\z} } \\
  <& 2 \eta (1 - \gamma) \ln(2m/\delta) \|\z\| / \sqrt{d}.
\end{align}

\cref*{item:ip-AP-lkjr,item:ip-AP-bvuu}
Similarly to \cref{le:bounds-network-output}.

\end{proof}

\begin{tcolorbox}
\begin{lemma}[Representation of network $g$]
\label{le:representation-network-g}
Suppose that \cref{asm:properties-network-f,asm:width-ap-f,asm:properties-network-g,asm:width-ap-g}.
\begin{enumerate}[label=(\alph*)]
  
  \item \label{item:prediction-f-g-a}
  In Scenario~(a), i.e., $\xadv_n := \r_n$,
  \begin{align}
    \notag
    & g(\z;T_g) \\
    \notag
    =& g(\z;0)
       + \sum^m_i \beta_i \kappa_{i,g}(\z)
         (\psi'_{g,i,T_g}(\z) - \psi'_{g,i,0}(\z)) h_{g,i,0}(\z) \\
       \notag
       &+ \frac{1}{N} \sum^N_n \yadv_n (\ip{\r_n}{\z} + 1)
          \int^{T_g}_0 \ell'_{g,n,t} \dt \\
          \notag
          & \quad \times
          \qty(
            \sum^m_i \beta^2_i \psi'_{g,i,0}(\r_n) \psi'_{g,i,0}(\z)
            - \Phi(\r_n, \z)
          ) \\
       \notag
       &+ \frac{1}{N} \sum^N_n \yadv_n \Phi(\r_n, \z)
          \int^{T_g}_0 \ell'_{g,n,t} \dt \\
       \notag
       &+ \frac{1}{N} \sum^N_n \yadv_n (\ip{\r_n}{\z} + 1)
          \sum^m_i \beta^2_i \kappa_{i,g}(\r_n) \\
          \notag
          & \quad \times
            \int^{T_g}_0 \ell'_{g,n,t} ( \psi'_{g,i,t}(\r_n) \psi'_{g,i,T_g}(\z)
            - \psi'_{g,i,0}(\r_n) \psi'_{g,i,0}(\z) ) \dt \\
       \notag
       &+ \frac{ \epsilon }
          {N \| \sum^m_i \alpha_i \psi'_{f,i,T_f}(\x_n) \v_i(T_f) \|}
       \Bigg[
         \sum^N_n \Phi(\r_n, \z) \int^{T_g}_0 \ell'_{g,n,t} \dt
         \Bigg\{ \\
           \notag
           & \quad
            \sum^m_i \alpha_i \psi'_{f,i,0}(\x_n) \ip{\v_i(0)}{\z} \\
           \notag
           & \quad
            + \sum^m_i \alpha_i \kappa_{f,i}(\x_n)
              (\psi'_{f,i,T_f}(\x_n) - \psi'_{f,i,0}(\x_n)) \ip{\v_i(0)}{\z} \\
           \notag
           & \quad
            + \frac{1}{N} \sum^N_k y_k \ip{\x_k}{\z} \int^{T_f}_0 \ell'_{f,k,t} \dt
              \qty( \sum^m_i \alpha^2_i \psi'_{f,i,0}(\x_n) \psi'_{f,i,0}(\x_k)
              - \Phi(\x_n, \x_k) ) \\
           \notag
           & \quad
            + \frac{1}{N} \sum^N_k y_k \ip{\x_k}{\z}
              \sum^m_i \alpha^2_i \kappa_{f,i}(\x_k) \\
              \notag
              & \qquad \times
              \int^{T_f}_0 \ell'_{f,k,t} ( \psi'_{f,i,T_f}(\x_n) \psi'_{f,i,t}(\x_k)
              - \psi'_{f,i,0}(\x_n) \psi'_{f,i,0}(\x_k) ) \dt
         \Bigg\} \\
         & \quad
           + \frac{1}{N} \sum^N_n \Phi(\r_n, \z) \int^{T_g}_0 \ell'_{g,n,t} \dt
             \sum^N_k y_k \Phi(\x_n, \x_k)
             \ip{\x_k}{\z} \int^{T_f}_0 \ell'_{f,k,t} \dt
       \Bigg].
  \end{align}

  \item \label{item:prediction-f-g-b}
  In Scenario~(b), i.e., $\xadv_n := \x_n + \r_n$,
  \begin{align}
    \notag
    & g(\z;T_g) \\
    \notag
    =& g(\z;0)
       + \sum^m_i \beta_i \kappa_{i,g}(\z)
         (\psi'_{g,i,T_g}(\z) - \psi'_{g,i,0}(\z)) h_{g,i,0}(\z) \\
       \notag
       &+ \frac{1}{N} \sum^N_n \yadv_n (\ip{\xadv_n}{\z} + 1)
          \int^{T_g}_0 \ell'_{g,n,t} \dt \\
          \notag
          & \quad \times
          \qty(
            \sum^m_i \beta^2_i \psi'_{g,i,0}(\xadv_n) \psi'_{g,i,0}(\z)
            - \Phi(\xadv_n, \z)
          ) \\
       \notag
       &+ \frac{1}{N} \sum^N_n \yadv_n \Phi(\xadv_n, \z) (\ip{\x_n}{\z} + 1)
          \int^{T_g}_0 \ell'_{g,n,t} \dt \\
       \notag
       &+ \frac{1}{N} \sum^N_n \yadv_n (\ip{\xadv_n}{\z} + 1)
          \sum^m_i \beta^2_i \kappa_{i,g}(\xadv_n) \\
       \notag
       & \quad \times
         \int^{T_g}_0 \ell'_{g,n,t} ( \psi'_{g,i,t}(\xadv_n) \psi'_{g,i,T_g}(\z)
         - \psi'_{g,i,0}(\xadv_n) \psi'_{g,i,0}(\z) ) \dt \\
       \notag
       &+ \frac{ \epsilon }
          {N \| \sum^m_i \alpha_i \psi'_{f,i,T_f}(\x_n) \v_i(T_f) \|}
       \Bigg[
         \sum^N_n \Phi(\xadv_n, \z) \int^{T_g}_0 \ell'_{g,n,t} \dt
         \Bigg\{ \\
         \notag
         & \quad
           \sum^m_i \alpha_i \psi'_{f,i,0}(\x_n) \ip{\v_i(0)}{\z} \\
         \notag
         & \quad
           + \sum^m_i \alpha_i \kappa_{i,g}(\x_n)
             (\psi'_{f,i,T_f}(\x_n) - \psi'_{f,i,0}(\x_n)) \ip{\v_i(0)}{\z} \\
         \notag
         & \quad
           + \frac{1}{N} \sum^N_k y_k \ip{\x_k}{\z} \int^{T_f}_0 \ell'_{f,k,t} \dt
             \qty( \sum^m_i \alpha^2_i \psi'_{f,i,0}(\x_n) \psi'_{f,i,0}(\x_k)
             - \Phi(\x_n, \x_k) ) \\
         \notag
         & \quad
           + \frac{1}{N} \sum^N_k y_k \ip{\x_k}{\z}
             \sum^m_i \alpha^2_i \kappa_{i,g}(\x_k) \\
           \notag
           & \qquad \times
           \int^{T_f}_0 \ell'_{f,k,t} ( \psi'_{f,i,T_f}(\x_n) \psi'_{f,i,t}(\x_k)
           - \psi'_{f,i,0}(\x_n) \psi'_{f,i,0}(\x_k) ) \dt
         \Bigg\} \\
         \notag
         & \quad
           + \frac{1}{N} \sum^N_n \Phi(\xadv_n, \z)
             \int^{T_g}_0 \ell'_{g,n,t} \dt \\
             & \qquad \times
             \sum^N_k y_k \Phi(\x_n, \x_k)
             \ip{\x_k}{\z} \int^{T_f}_0 \ell'_{f,k,t} \dt
       \Bigg].
  \end{align}
  
\end{enumerate}
\end{lemma}
\end{tcolorbox}

\begin{proof}
Similarly to \cref{le:representation-network-f},
\begin{align}
  \notag
  g(\z;T_g)
  =& g(\z;0)
     + \sum^m_i \beta_i \kappa_{i,g}(\z)
       (\psi'_{g,i,T_g}(\z) - \psi'_{g,i,0}(\z)) h_{g,i,0}(\z) \\
     \notag
     &+ \frac{1}{N} \sum^N_n \yadv_n (\ip{\xadv_n}{\z} + 1)
        \int^{T_g}_0 \ell'_{g,n,t} \dt \\
        \notag
        & \quad \times
        \qty(
          \sum^m_i \beta^2_i \psi'_{g,i,0}(\xadv_n) \psi'_{g,i,0}(\z)
          - \Phi(\xadv_n, \z)
        ) \\
     \notag
     &+ \frac{1}{N} \sum^N_n \yadv_n \Phi(\xadv_n, \z)
        \ip{\xadv_n}{\z} \int^{T_g}_0 \ell'_{g,n,t} \dt \\
     \notag
     &+ \frac{1}{N} \sum^N_n \yadv_n \Phi(\xadv_n, \z)
        \int^{T_g}_0 \ell'_{g,n,t} \dt \\
     \notag
     &+ \frac{1}{N} \sum^N_n \yadv_n (\ip{\xadv_n}{\z} + 1)
        \sum^m_i \beta^2_i \kappa_{i,g}(\xadv_n) \\
     & \quad \times
       \int^{T_g}_0 \ell'_{g,n,t} ( \psi'_{g,i,t}(\xadv_n) \psi'_{g,i,T_g}(\z)
       - \psi'_{g,i,0}(\xadv_n) \psi'_{g,i,0}(\z) ) \dt.
\end{align}
By \cref{le:AP},
\begin{align}
  \notag
  & \frac{1}{N} \sum^N_n \yadv_n \Phi(\xadv_n, \z)
  \ip{\r_n}{\z} \int^{T_g}_0 \ell'_{g,n,t} \dt \\
  =& \frac{
       \epsilon \sum^N_n \Phi(\xadv_n, \z) \int^{T_g}_0 \ell'_{g,n,t} \dt
       \sum^m_i \alpha_i \psi'_{f,i,T_f}(\x_n) \ip{\v_i(T_f)}{\z}
     }{N \| \sum^m_i \alpha_i \psi'_{f,i,T_f}(\x_n) \v_i(T_f) \|}.
\end{align}
The numerator can be also rearranged as
\begin{align}
  \notag
  & \sum^N_n \Phi(\xadv_n, \z) \int^{T_g}_0 \ell'_{g,n,t} \dt
  \sum^m_i \alpha_i \psi'_{f,i,T_f}(\x_n) \ip{\v_i(T_f)}{\z} \\
  \notag
  =& \sum^N_n \Phi(\xadv_n, \z) \int^{T_g}_0 \ell'_{g,n,t} \dt
     \Bigg(
       \sum^m_i \alpha_i \psi'_{f,i,0}(\x_n) \ip{\v_i(0)}{\z} \\
       \notag
       &+ \sum^m_i \alpha_i \kappa_{f,i}(\x_n)
         (\psi'_{f,i,T_f}(\x_n) - \psi'_{f,i,0}(\x_n)) \ip{\v_i(0)}{\z} \\
       \notag
       &+ \frac{1}{N} \sum^N_k y_k \ip{\x_k}{\z} \int^{T_f}_0 \ell'_{f,k,t} \dt
         \qty( \sum^m_i \alpha^2_i \psi'_{f,i,0}(\x_n) \psi'_{f,i,0}(\x_k)
         - \Phi(\x_n, \x_k) ) \\
       \notag
       &+ \frac{1}{N} \sum^N_k y_k \ip{\x_k}{\z} \sum^m_i \alpha^2_i \kappa_{f,i}(\x_k) \\
         \notag
         & \quad \times
         \int^{T_f}_0 \ell'_{f,k,t} ( \psi'_{f,i,T_f}(\x_n) \psi'_{f,i,t}(\x_k)
         - \psi'_{f,i,0}(\x_n) \psi'_{f,i,0}(\x_k) ) \dt
     \Bigg) \\
     &+ \frac{1}{N} \sum^N_n \Phi(\xadv_n, \z) \int^{T_g}_0 \ell'_{g,n,t} \dt
        \sum^N_k y_k \Phi(\x_n, \x_k)
        \ip{\x_k}{\z} \int^{T_f}_0 \ell'_{f,k,t} \dt.
\end{align}
\end{proof}

\begin{tcolorbox}
\begin{lemma}[Network prediction of $g$]
\label{le:prediction-g}
Suppose that \cref{asm:properties-network-f,asm:width-ap-f,asm:properties-network-g,asm:width-ap-g}.
\begin{enumerate}[label=(\alph*)]

  \item \label{item:prediction-g-a}
  In Scenario~(a), if
  \begin{align}
    \notag
    & \abs{ \frac{1}{N^2} \sum^N_n \Phi(\r_n, \z)
      \int^{T_g}_0 \ell'_{g,n,t} \dt
      \sum^N_k y_k \Phi(\x_n, \x_k) \ip{\x_k}{\z}
      \int^{T_f}_0 \ell'_{f,k,t} \dt } \\
    \notag
    >& \tildecalO\Bigg(
      \frac{ \sqrt{d} \int^{T_f}_0 \barell'_{f,k,t} \dt }{ \epsilon }
      \qty( 1 + \abs{ \frac{1}{N} \sum^N_n \yadv_n
      \Phi(\r_n, \z) \int^{T_g}_0 \ell'_{g,n,t} \dt } ) \\
      &+ \int^{T_g}_0 \barell'_{g,n,t} \dt
    \Bigg),
  \end{align}
  then
  \begin{align}
    \notag
    \sgn( g(\z;T_g) )
    =& \sgn\Bigg(
         \frac{1}{N^2} \sum^N_n \Phi(\r_n, \z)
         \int^{T_g}_0 \ell'_{g,n,t} \dt \\
         & \times \sum^N_k y_k \Phi(\x_n, \x_k) \ip{\x_k}{\z}
                  \int^{T_f}_0 \ell'_{f,k,t} \dt
       \Bigg).
  \end{align}

  \item \label{item:prediction-g-b}
  In Scenario~(b), if
  \begin{align}
    \notag
    & \abs{ \frac{1}{N^2} \sum^N_n \Phi(\xadv_n, \z)
      \int^{T_g}_0 \ell'_{g,n,t} \dt
      \sum^N_k y_k \Phi(\x_n, \x_k) \ip{\x_k}{\z}
      \int^{T_f}_0 \ell'_{f,k,t} \dt } \\
    \notag
    >& \tildecalO\Bigg(
      \frac{ \sqrt{d} \int^{T_f}_0 \barell'_{f,k,t} \dt }{ \epsilon }
      \Bigg( 1 + \abs{ \frac{1}{N} \sum^N_n \yadv_n \Phi(\xadv_n, \z)
             (\ip{\x_n}{\z} + 1) \int^{T_g}_0 \ell'_{g,n,t} \dt } \Bigg) \\
      &+ \int^{T_g}_0 \barell'_{g,n,t} \dt
    \Bigg),
  \end{align}
  then
  \begin{align}
    \notag
    \sgn( g(\z;T_g) )
    =& \sgn\Bigg(
         \frac{1}{N^2} \sum^N_n \Phi(\xadv_n, \z)
         \int^{T_g}_0 \ell'_{g,n,t} \dt \\
         & \times \sum^N_k y_k \Phi(\x_n, \x_k) \ip{\x_k}{\z}
                  \int^{T_f}_0 \ell'_{f,k,t} \dt
       \Bigg).
  \end{align}
  
\end{enumerate}
\end{lemma}
\end{tcolorbox}

\begin{proof}
We prove \cref*{item:prediction-f-g-a}. Similarly, \cref*{item:prediction-f-g-b} can be established. By \cref{le:bounds-network-output,le:norm-AP,le:ip-AP,le:representation-network-g}, if
\begin{align}
  \notag
  & \abs{ \frac{1}{N^2} \sum^N_n \Phi(\r_n, \z) \int^{T_g}_0 \ell'_{g,n,t} \dt
  \sum^N_k y_k \Phi(\x_n, \x_k) \ip{\x_k}{\z} \int^{T_f}_0 \ell'_{f,k,t} \dt } \\
  \notag
  >&
  \frac{\| \sum^m_i \alpha_i \psi'_{f,i,T_f}(\x_n) \v_i(T_f) \|}{ \epsilon }
  \Bigg\{
  |g(\z;0)|
  +
  \abs{ \sum^m_i \beta_i \kappa_{i,g}(\z)
  (\psi'_{g,i,T_g}(\z) - \psi'_{g,i,0}(\z)) h_{g,i,0}(\z) } \\
  \notag
  &+
  \Bigg| \frac{1}{N} \sum^N_n \yadv_n (\ip{\r_n}{\z} + 1)
  \int^{T_g}_0 \ell'_{g,n,t} \dt
  \qty( \sum^m_i \beta^2_i \psi'_{g,i,0}(\r_n) \psi'_{g,i,0}(\z)
  - \Phi(\r_n, \z)) \Bigg| \\
  \notag
  &+
  \abs{ \frac{1}{N} \sum^N_n \yadv_n
  \Phi(\r_n, \z) \int^{T_g}_0 \ell'_{g,n,t} \dt } \\
  \notag
  &+
  \Bigg| \frac{1}{N} \sum^N_n \yadv_n (\ip{\r_n}{\z} + 1)
  \sum^m_i \beta^2_i \kappa_{i,g}(\r_n) \\
  \notag
  & \quad \times
  \int^{T_g}_0 \ell'_{g,n,t} ( \psi'_{g,i,t}(\r_n) \psi'_{g,i,T_g}(\z)
  - \psi'_{g,i,0}(\r_n) \psi'_{g,i,0}(\z) ) \dt \Bigg|
  \Bigg\} \\
  \notag
  &+
  \frac{1}{N} \sum^N_n \Phi(\r_n, \z) \int^{T_g}_0 \ell'_{g,n,t} \dt
  \Bigg\{
  \abs{ \sum^m_i \alpha_i \psi'_{f,i,0}(\x_n) \ip{\v_i(0)}{\z} } \\
  \notag
  &+
  \abs{ \sum^m_i \alpha_i \kappa_{f,i}(\x_n)
  (\psi'_{f,i,T_f}(\x_n) - \psi'_{f,i,0}(\x_n)) \ip{\v_i(0)}{\z} } \\
  \notag
  &+
  \abs{ \frac{1}{N} \sum^N_k y_k \ip{\x_k}{\z} \int^{T_f}_0 \ell'_{f,k,t} \dt
  \qty( \sum^m_i \alpha^2_i \psi'_{f,i,0}(\x_n) \psi'_{f,i,0}(\x_k)
  - \Phi(\x_n, \x_k) ) } \\
  \notag
  &+
  \Bigg| \frac{1}{N} \sum^N_k y_k \ip{\x_k}{\z}
  \sum^m_i \alpha^2_i \kappa_{f,i}(\x_k) \\
  & \quad \times
  \int^{T_f}_0 \ell'_{f,k,t} ( \psi'_{f,i,T_f}(\x_n) \psi'_{f,i,t}(\x_k)
  - \psi'_{f,i,0}(\x_n) \psi'_{f,i,0}(\x_k) ) \dt \Bigg|
  \Bigg\} \\
  \notag
  =&
  \frac{\tildecalO\qty(\sqrt{d} \int^{T_f}_0 \barell'_{f,k,t} \dt)}{\epsilon}
  \qty( \tildecalO(1) + \abs{ \frac{1}{N} \sum^N_n \yadv_n
  \Phi(\r_n, \z) \int^{T_g}_0 \ell'_{g,n,t} \dt } ) \\
  &+ \tildecalO\qty( \int^{T_g}_0 \barell'_{g,n,t} \dt ),
\end{align}
then
\begin{align}
  \notag
  & \sgn(g(\z;T_g)) \\
  =& \sgn\qty(
    \frac{1}{N^2} \sum^N_n \Phi(\r_n, \z) \int^{T_g}_0 \ell'_{g,n,t} \dt
    \sum^N_k y_k \Phi(\x_n, \x_k) \ip{\x_k}{\z} \int^{T_f}_0 \ell'_{f,k,t} \dt
  ).
\end{align}
\end{proof}

\begin{tcolorbox}
\begin{theorem}[Perturbation learning, Scenario (a), general case]
\label{th:PL-a-general}
Consider Scenario~(a) in \cref{setting}. Let $\delta = \Theta(1)$ be a small positive number and
\begin{align}
  \hatf^\gen(\z)
  :=& \frac{1}{N} \sum^N_n y_n \Phi(\x_n, \z) \ip{\x_n}{\z}
      \int^{T_f}_0 \ell'_{f,n,t} \dt, \\
  \hatg^\gen_a(\z)
  :=& \frac{1}{N^2} \sum^N_n \Phi(\r_n, \z) \int^{T_g}_0 \ell'_{g,n,t} \dt
      \sum^N_k y_k \Phi(\x_n, \x_k) \ip{\x_k}{\z} \int^{T_f}_0 \ell'_{f,k,t} \dt.
\end{align}
Under \cref{asm:width}, for any $\z \in \bbR^d$, if
\begin{align}
  & |\hatf^\gen(\z)| > \tildecalO\qty(1 + \int^{T_f}_0 \barell_{f,n,t} \dt), \\
  \notag
  & |\hatg^\gen_a(\z)| \\
  >& \tildecalO\qty(
    \frac{ \sqrt{d} \int^{T_f}_0 \barell'_{f,k,t} \dt }{ \epsilon }
    \qty( 1 + \abs{ \frac{1}{N} \sum^N_n \yadv_n
    \Phi(\r_n, \z) \int^{T_g}_0 \ell'_{g,n,t} \dt } )
    + \int^{T_g}_0 \barell'_{g,n,t} \dt
  ), \\
  & \sgn(\hatf^\gen(\z)) = \sgn(\hatg^\gen_a(\z)),
\end{align}
then, with probability at least $1 - \delta$, $\sgn(f(\z;T_f)) = \sgn(g(\z;T_g))$ holds.
\end{theorem}
\end{tcolorbox}

\begin{proof}
By Bonferroni's inequality and \cref{asm:width-ap-f,asm:width-ap-g,le:properties-network-f,le:properties-network-g,le:prediction-f,le:prediction-g}, the claim is established.
\end{proof}

\begin{tcolorbox}
\begin{theorem}[Perturbation learning, Scenario (b), general case]
\label{th:PL-b-general}
Consider Scenario~(b) in \cref{setting}. Let $\delta = \Theta(1)$ be a small positive number and
\begin{align}
  \hatf^\gen(\z)
  :=& \frac{1}{N} \sum^N_n y_n \Phi(\x_n, \z) \ip{\x_n}{\z}
      \int^{T_f}_0 \ell'_{f,n,t} \dt, \\
  \hatg^\gen_b(\z)
  :=& \frac{1}{N^2} \sum^N_n \Phi(\xadv_n, \z) \int^{T_g}_0 \ell'_{g,n,t} \dt
      \sum^N_k y_k \Phi(\x_n, \x_k) \ip{\x_k}{\z} \int^{T_f}_0 \ell'_{f,k,t} \dt.
\end{align}
Under \cref{asm:width}, for any $\z \in \bbR^d$, if
\begin{align}
  & |\hatf^\gen(\z)| > \tildecalO\qty(1 + \int^{T_f}_0 \barell_{f,n,t} \dt), \\
  \notag
  & |\hatg^\gen_b(\z)| \\
  \notag
  >& \tildecalO\Bigg(
    \frac{ \sqrt{d} \int^{T_f}_0 \barell'_{f,k,t} \dt }{ \epsilon }
    \qty( 1 + \abs{ \frac{1}{N} \sum^N_n \yadv_n \Phi(\xadv_n, \z)
    (\ip{\x_n}{\z} + 1) \int^{T_g}_0 \ell'_{g,n,t} \dt } ) \\
    &+ \int^{T_g}_0 \barell'_{g,n,t} \dt
  \Bigg), \\
  & \sgn(\hatf^\gen(\z)) = \sgn(\hatg^\gen_b(\z)),
\end{align}
then, with probability at least $1 - \delta$, $\sgn(f(\z;T_f)) = \sgn(g(\z;T_g))$ holds.
\end{theorem}
\end{tcolorbox}

\begin{proof}
By Bonferroni's inequality and \cref{asm:width-ap-f,asm:width-ap-g,le:properties-network-f,le:properties-network-g,le:prediction-f,le:prediction-g}, the claim is established.
\end{proof}

\begin{tcolorbox}
\AP*
\end{tcolorbox}

\begin{proof}
By Bonferroni's inequality and \cref{asm:width-ap-f,asm:width-ap-g,le:properties-network-f,le:properties-network-g,le:AP,le:norm-AP}, the claim is established.
\end{proof}

\begin{tcolorbox}
\PLa*
\end{tcolorbox}

\begin{proof}
By Bonferroni's inequality and \cref{le:Hoeffding,th:PL-a-general}, the claim is established.
\end{proof}

\begin{tcolorbox}
\PLb*
\end{tcolorbox}

\begin{proof}
By Bonferroni's inequality and \cref{le:Hoeffding,th:PL-b-general}, the claim is established.
\end{proof}

\begin{tcolorbox}
\begin{lemma}[Sufficient condition of agreement condition]
\label{le:sufficient-agree-cond}
If
\begin{align}
  \frac{ | \sum^N_n y_n \ip{\x_n}{\z} \int^{T_f}_0 \ell'_{f,n,t} \dt | }
       { \max_{ \x \in \{\x_1,\ldots,\x_N,\z\} }
         \sum^N_n \lambda(\x_n, \x) |\ip{\x_n}{\z}|
         \int^{T_f}_0 \ell'_{f,n,t} \dt }
  > \frac{1 - \gamma}{1 + \gamma},
\end{align}
then $\sgn(\hatf^\gen(\z)) = \sgn(\hatg^\gen_a(\z)) = \sgn(\hatg^\gen_b(\z))$ holds.
\end{lemma}
\end{tcolorbox}

\begin{proof}
By \cref{le:expected-derivatives-2},
\begin{align}
  \notag
  & \abs{ \sum^N_n y_n \ip{\x_n}{\z} \int^{T_f}_0 \ell'_{f,n,t} \dt
    \qty(\Phi(\x_n, \z) - \frac{(1+\gamma)^2}{4} ) } \\
  \leq &
  \frac{(1+\gamma)(1-\gamma)}{4}
  \sum^N_n \lambda(\x_n, \z) |\ip{\x_n}{\z}| \int^{T_f}_0 \ell'_{f,n,t} \dt.
\end{align}
In addition,
\begin{align}
  \notag
  & \frac{(1+\gamma)^2}{4} \abs{ \sum^N_n y_n
  \ip{\x_n}{\z} \int^{T_f}_0 \ell'_{f,n,t} \dt } \\
  &> \frac{(1+\gamma)(1-\gamma)}{4} \sum^N_n \lambda(\x_n, \z)
     |\ip{\x_n}{\z}| \int^{T_f}_0 \ell'_{f,n,t} \dt \\
  \label[ineq]{ineq:tmp-luna}
  & \Longleftarrow
  \frac{ \abs{ \sum^N_n y_n \ip{\x_n}{\z} \int^{T_f}_0 \ell'_{f,n,t} \dt } }
  { \sum^N_n \lambda(\x_n, \z) |\ip{\x_n}{\z}| \int^{T_f}_0 \ell'_{f,n,t} \dt }
  > \frac{1 - \gamma}{1 + \gamma}.
\end{align}
Thus, if \cref{ineq:tmp-luna} holds, then
\begin{align}
  \sgn\qty( \sum^N_n y_n \Phi(\x_n, \z)
  \ip{\x_n}{\z} \int^{T_f}_0 \ell'_{f,n,t} \dt )
  = \sgn\qty( \sum^N_n y_n \ip{\x_n}{\z} \int^{T_f}_0 \ell'_{f,n,t} \dt ).
\end{align}
Similarly, for any $k \in [N]$, if
\begin{align}
  \label[ineq]{ineq:tmp-luby}
  \frac{ \abs{ \sum^N_n y_n \ip{\x_n}{\z} \int^{T_f}_0 \ell'_{f,n,t} \dt } }
  { \sum^N_n \lambda(\x_n, \x_k) |\ip{\x_n}{\z}| \int^{T_f}_0 \ell'_{f,n,t} \dt }
  > \frac{1 - \gamma}{1 + \gamma},
\end{align}
then
\begin{align}
  \sgn\qty( \sum^N_n y_n \Phi(\x_n, \x_k)
  \ip{\x_n}{\z} \int^{T_f}_0 \ell'_{f,n,t} \dt )
  = \sgn\qty( \sum^N_n y_n \ip{\x_n}{\z} \int^{T_f}_0 \ell'_{f,n,t} \dt ).
\end{align}
When \cref{ineq:tmp-luby} holds for every $k \in [N]$,
\begin{align}
  \notag
  &\sgn\qty(
    \sum^N_n \Phi(\r_n, \z) \int^{T_g}_0 \ell'_{g,n,t} \dt
    \sum^N_l y_l \Phi(\x_n, \x_l) \ip{\x_l}{\z} \int^{T_f}_0 \ell'_{f,l,t} \dt
  ) \\
  =& \sgn\qty(
    \sum^N_l y_l \Phi(\x_n, \x_l) \ip{\x_l}{\z} \int^{T_f}_0 \ell'_{f,l,t} \dt
  ) \\
  =& \sgn\qty( \sum^N_n y_n \ip{\x_n}{\z} \int^{T_f}_0 \ell'_{f,n,t} \dt ).
\end{align}
By integrating \cref{ineq:tmp-luna,ineq:tmp-luby}, the claim is established.

\end{proof}

\clearpage

\section*{NeurIPS Paper Checklist}

\begin{enumerate}

\item {\bf Claims}
    \item[] Question: Do the main claims made in the abstract and introduction accurately reflect the paper's contributions and scope?
    \item[] Answer: \answerYes{} 
    \item[] Justification: We have accurately described the contributions and limitations in Abstract and Introduction.

\item {\bf Limitations}
    \item[] Question: Does the paper discuss the limitations of the work performed by the authors?
    \item[] Answer: \answerYes{} 
    \item[] Justification: The limitations are described in \cref{sec:comparison}.

\item {\bf Theory Assumptions and Proofs}
    \item[] Question: For each theoretical result, does the paper provide the full set of assumptions and a complete (and correct) proof?
    \item[] Answer: \answerYes{} 
    \item[] Justification: The assumption is described in \cref{asm:width} and a brief proof is provided in \cref{sec:sketch}. The complete proof can be found in \cref{sec:main-proof}.

    \item {\bf Experimental Result Reproducibility}
    \item[] Question: Does the paper fully disclose all the information needed to reproduce the main experimental results of the paper to the extent that it affects the main claims and/or conclusions of the paper (regardless of whether the code and data are provided or not)?
    \item[] Answer: \answerYes{} 
    \item[] Justification: The experimental setup is described in detail in \cref{sec:experiments-ap}.

\item {\bf Open access to data and code}
    \item[] Question: Does the paper provide open access to the data and code, with sufficient instructions to faithfully reproduce the main experimental results, as described in supplemental material?
    \item[] Answer: \answerYes{} 
    \item[] Justification: All datasets we used are either openly accessible or can be artificially generated. The code is provided as supplementary material.

\item {\bf Experimental Setting/Details}
    \item[] Question: Does the paper specify all the training and test details (e.g., data splits, hyperparameters, how they were chosen, type of optimizer, etc.) necessary to understand the results?
    \item[] Answer: \answerYes{} 
    \item[] Justification: The experimental setup is described in detail in \cref{sec:experiments-ap}.

\item {\bf Experiment Statistical Significance}
    \item[] Question: Does the paper report error bars suitably and correctly defined or other appropriate information about the statistical significance of the experiments?
    \item[] Answer: \answerNo{} 
    \item[] Justification: Error bars were not measured due to computational costs. However, we provide extensive experimental results to support our theoretical findings in \cref{sec:experiments-ap}.

\item {\bf Experiments Compute Resources}
    \item[] Question: For each experiment, does the paper provide sufficient information on the computer resources (type of compute workers, memory, time of execution) needed to reproduce the experiments?
    \item[] Answer: \answerYes{} 
    \item[] Justification: Our experiments were conducted on an NVIDIA A100.
    
\item {\bf Code Of Ethics}
    \item[] Question: Does the research conducted in the paper conform, in every respect, with the NeurIPS Code of Ethics \url{https://neurips.cc/public/EthicsGuidelines}?
    \item[] Answer: \answerYes{} 
    \item[] Justification: Our paper strictly adheres to the NeurIPS Code of Ethics.

\item {\bf Broader Impacts}
    \item[] Question: Does the paper discuss both potential positive societal impacts and negative societal impacts of the work performed?
    \item[] Answer: \answerNA{} 
    \item[] Justification: This does not apply as it is a theoretical study.
    
\item {\bf Safeguards}
    \item[] Question: Does the paper describe safeguards that have been put in place for responsible release of data or models that have a high risk for misuse (e.g., pretrained language models, image generators, or scraped datasets)?
    \item[] Answer: \answerNA{} 
    \item[] Justification: Our theoretical research does not involve such releases.

\item {\bf Licenses for existing assets}
    \item[] Question: Are the creators or original owners of assets (e.g., code, data, models), used in the paper, properly credited and are the license and terms of use explicitly mentioned and properly respected?
    \item[] Answer: \answerYes{} 
    \item[] Justification: We have accurately cited credits.

\item {\bf New Assets}
    \item[] Question: Are new assets introduced in the paper well documented and is the documentation provided alongside the assets?
    \item[] Answer: \answerNA{} 
    \item[] Justification: We do not provide such assets.

\item {\bf Crowdsourcing and Research with Human Subjects}
    \item[] Question: For crowdsourcing experiments and research with human subjects, does the paper include the full text of instructions given to participants and screenshots, if applicable, as well as details about compensation (if any)? 
    \item[] Answer: \answerNA{} 
    \item[] Justification: We did not conduct such experiments.

\item {\bf Institutional Review Board (IRB) Approvals or Equivalent for Research with Human Subjects}
    \item[] Question: Does the paper describe potential risks incurred by study participants, whether such risks were disclosed to the subjects, and whether Institutional Review Board (IRB) approvals (or an equivalent approval/review based on the requirements of your country or institution) were obtained?
    \item[] Answer: \answerNA{} 
    \item[] Justification: We did not conduct experiments that require this.

\end{enumerate}

\end{document}